\documentclass[12pt]{article}
\usepackage[utf8]{inputenc}
\usepackage[a4paper, total={6in, 8in}]{geometry}

\usepackage{verbatim}
\usepackage{color}
\usepackage[table]{xcolor}

\usepackage{graphicx}
\graphicspath{{./Figures/}}
\usepackage{epstopdf}
\usepackage{amsmath,amsmath,amssymb}
\usepackage{url}
\usepackage{footmisc}
\usepackage{multirow}

\usepackage[caption=false,font=footnotesize]{subfig}
\usepackage{algorithm}
\usepackage{algpseudocode}

\title{Decomposition in Decision and Objective Space for Multi-Modal Multi-Objective Optimization}
\author{Monalisa Pal and Sanghamitra Bandyopadhyay\thanks{M. Pal and S. Bandyopadhyay are with Machine Intelligence Unit, Indian Statistical Institute, 203 Barrackpore Trunk Road, Kolkata - 700108, West Bengal, India. E-mail: M. Pal (monalisap90@gmail.com) and S. Bandyopadhyay (sanghami@isical.ac.in). This work has been partially supported by Indian Statistical Institute, Kolkata. It has also been supported by J. C. Bose Fellowship (SB/SJ/JCB-033/2016) from Department of Science and Technology, Government of India. This version corresponds to the accepted manuscript (DOI: 10.1016/j.swevo.2021.100842) at https://www.sciencedirect.com/science/article/abs/pii/S2210650221000031.}}
\date{}

\providecommand{\keywords}[1]{\textbf{\textit{Keywords:}} #1}

\begin{document}
\maketitle
\begin{abstract}
    Multi-modal multi-objective optimization problems (MMMOPs) have multiple subsets within the Pareto-optimal Set, each independently mapping to the same Pareto-Front. Prevalent multi-objective evolutionary algorithms are not purely designed to search for multiple solution subsets, whereas, algorithms designed for MMMOPs demonstrate degraded performance in the objective space. This motivates the design of better algorithms for addressing MMMOPs. The present work identifies the crowding illusion problem originating from using crowding distance globally over the entire decision space. Subsequently, an evolutionary framework, called graph Laplacian based Optimization using Reference vector assisted Decomposition (LORD), is proposed, which uses decomposition in both objective and decision space for dealing with MMMOPs. Its filtering step is further extended to present LORD-II algorithm, which demonstrates its dynamics on multi-modal many-objective problems. The efficacies of the frameworks are established by comparing their performance on test instances from the CEC 2019 multi-modal multi-objective test suite and polygon problems with the state-of-the-art algorithms for MMMOPs and other multi- and many-objective evolutionary algorithms. The manuscript is concluded by mentioning the limitations of the proposed frameworks and future directions to design still better algorithms for MMMOPs. The source code is available at \url{https://worksupplements.droppages.com/lord}.
\end{abstract}

\section{Introduction}
\label{sec1}
Multi-objective optimization deals with problems having two or more conflicting objectives (optimization criteria) \cite{SWEVO_2011,UM_2014TEVCReview_PartI}. The mathematical formulation of a box-constrained multi-objective minimization problem \eqref{eq:MOOdef1} presents the mapping from a $N$-dimensional vector ($\mathbf{X} = \left[x_1, \cdots,x_N\right]$) in the decision space ($\mathcal{D}$) to a $M$-dimensional vector ($\mathbf{F}(\mathbf{X})$) in the objective space \cite{SWEVO_2011,UM_2014TEVCReview_PartI}.
\begin{equation}
	\begin{gathered}
		\text{Minimize }\mathbf{F}(\mathbf{X}) = \left[f_1(\mathbf{X}), f_2(\mathbf{X}), \cdots, f_M(\mathbf{X})\right]\\
		\text{where, } \mathbf{X}\in \mathcal{D}\left(\subseteq\mathbb{R}^N\right)\text{, }\mathbf{F}(\mathbf{X}):\mathcal{D}\mapsto\mathbb{R}^M\\
		\text{and } \mathcal{D}:x_j^L\leq x_j\leq x_j^U, \forall j = 1, 2, \cdots, N
	\end{gathered}
	\label{eq:MOOdef1}
\end{equation}

Pareto-dominance relation is used for comparison of two vectors, i.e., $\mathbf{X}$ Pareto-dominates $\mathbf{Y}$, as defined below.
\begin{equation}
	\begin{gathered}
		\forall i \in \{1, 2, \cdots, M\},\text{ and }\exists j \in \{1,2,\cdots,M\},\\
		\mathbf{X} \prec \mathbf{Y} \iff \left(f_i(\mathbf{X})\leq f_i(\mathbf{Y}) \land f_j(\mathbf{X})<f_j(\mathbf{Y})\right)
	\end{gathered}
	\label{eq:paretodominance}
\end{equation}

A Pareto-optimal solution $\mathbf{X}^\star \in \mathcal{D}$ is attained, if $\nexists \mathbf{X} \in \mathcal{D}$ that dominates $\mathbf{X}^\star$. A set of all such Pareto-optimal solutions form the Pareto-optimal Set (PS) and their images in the objective space yield the Pareto-Front (PF) \cite{SWEVO_2011,UM_2014TEVCReview_PartI}.

\begin{figure}[!t]
	\centering
	\subfloat[PS of MMF4]{\includegraphics[width=0.5\linewidth]{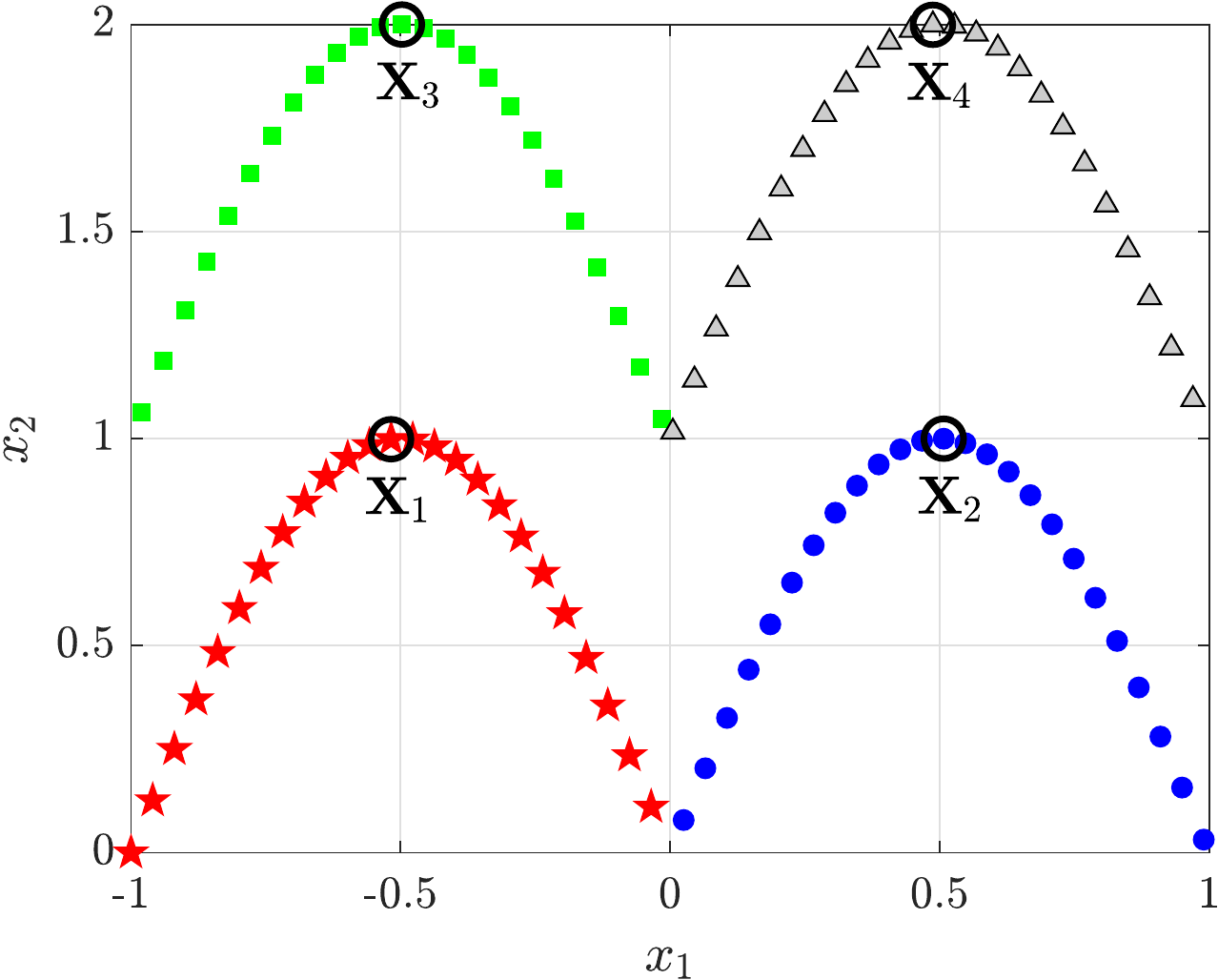}
		\label{fig:MMMOP_PSs}}
	\subfloat[PF of MMF4]{\includegraphics[width=0.5\linewidth]{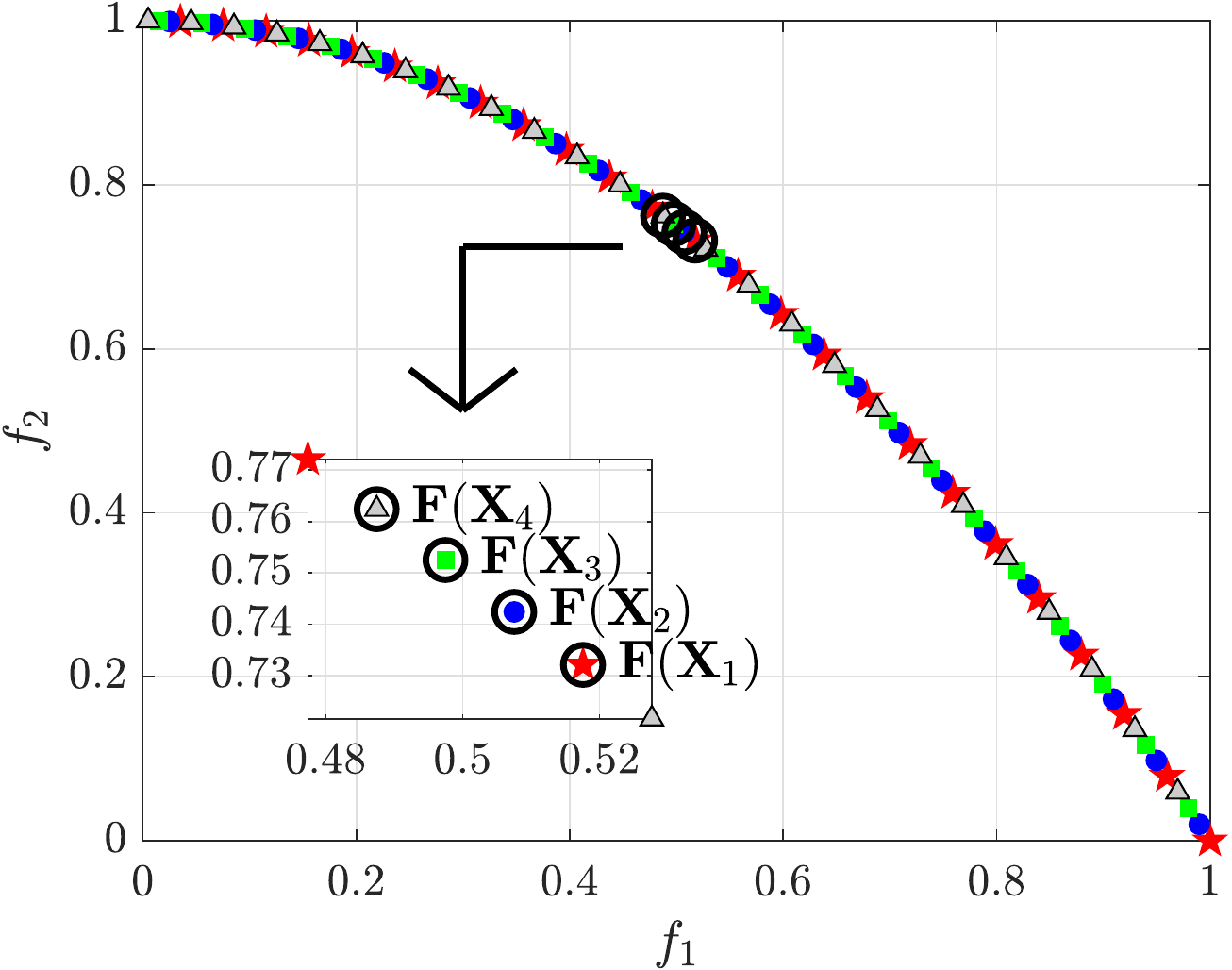}
		\label{fig:MMMOP_PF}}
	\caption{Four solution vectors ($\mathbf{X}_1, \mathbf{X}_2, \mathbf{X}_3$ and $\mathbf{X}_4$) mapping to \emph{almost same} objective vectors ($\mathbf{F}(\mathbf{X}_1), \mathbf{F}(\mathbf{X}_2), \mathbf{F}(\mathbf{X}_3)$ and $\mathbf{F}(\mathbf{X}_4)$) for a benchmark test problem (MMF4 \cite{CEC19_TR}).}
	\label{fig:MMMOP}
\end{figure}

The notion of Multi-Modal Multi-Objective Problem (MMMOP) \cite{Review_MMMOP} arises when a set of $k_{PS}$ $(\geq 2)$ distinct decision vectors ($\mathcal{A}_M = \left\{\mathbf{X}_1\right.$, $\mathbf{X}_2$, $\cdots$, $\left.\mathbf{X}_{k_{PS}}\right\}$) maps to \emph{almost same} objective vectors, i.e., $\forall \left(\mathbf{X}_i, \mathbf{X}_j\right)\in \mathcal{A}_M \times \mathcal{A}_M$, $\left\Vert \mathbf{F}\left(\mathbf{X}_i\right) - \mathbf{F}\left(\mathbf{X}_j\right) \right\Vert < \epsilon$ (a small number) as illustrated in Fig. \ref{fig:MMMOP} for a benchmark test problem (MMF4 \cite{CEC19_TR}). Thus, the PS can consist of multiple subsets of non-dominated solutions, where each subset can independently generate the entire PF. 

Research on MMMOPs is motivated to discover those $k_{PS}$ alternative solutions for nearly the same objective values to facilitate the comparison of non-numeric, domain-specific attributes of these equivalent solutions during decision-making. Moreover, when the practical implementation of a solution is hindered, a nearly equivalent alternative can be beneficial. Such MMMOPs are seen in rocket engine design \cite{Appl2_MMMOP}, feature selection \cite{Appl1_MMMOP} and path-planning problems \cite{Appl3_MMMOP}.

To optimize such MMMOPs, an Evolutionary Algorithm (EA) faces the following challenges:
\begin{enumerate}
	\item Maintaining diversity in the decision space, i.e., representing and maintaining diversity within each of the multiple solution subsets which independently maps to a diverse approximation of the PF.
	\item Necessity of a large population to efficiently represent an MMMOP. For example, if $k_{PF}$ points (e.g., $100$) represent a 2-objective PF and $k_{PS}$ decision vectors (e.g., $4$ for MMF4 problem \cite{CEC19_TR,DETriM}) map to each point of the PF, then the final population size required is $k_{PF}\times k_{PS}$ (e.g., $100\times 4 = 400$).
\end{enumerate}

Such MMMOPs cannot be tackled by any of the existing Multi-Objective Evolutionary Algorithms (MOEAs) \cite{UM_2014TEVCReview_PartI,SWEVO_2011}: Pareto-dominance based EAs (such as NSGA-II \cite{NSGA2} and $\theta$-DEA \cite{thDEA}), indicator-based EAs (such as HypE \cite{HypE} and GDE-MOEA \cite{GDE-MOEA}) and decomposition-based EAs (such as MOEA/D \cite{MOEA/D}, NSGA-III \cite{NSGA3} and MOEA/DD \cite{MOEADD}), as these MOEAs focus primarily on the objective space to regulate the population, overlooking the solution distribution in the decision space. 

For decomposition-based EAs, Das and Dennis's approach \cite{NBI} is used which involves a two-layered recursive procedure to define reference vectors from the origin to a set of uniformly distributed points on a unit hyperplane in the objective space. These reference vectors partition the objective space into multiple sub-spaces and thereby, reduces the problem complexity by restricting one or more steps of the EA within these subspaces \cite{Sys_Decomp}. As these algorithms neither suffer from the reduced selection pressure in high dimensional objective space like Pareto-dominance based EAs \cite{SWEVO_2011,UM_2014TEVCReview_PartI} nor require the extreme computational effort for hypervolume-like indicator evaluation \cite{HypE}, decomposition based EAs are extensively used for $M$-objective problems.

Omni-optimizer \cite{Omni} is the earliest work to consider the solution diversity in the decision space\footnote{In this article, decision space, variable space and solution space are considered as synonymous.}. It uses of crowding distance in the decision space (CDX) after the non-dominated sorting \cite{Omni} but hampers the solution diversity in the objective space. The work in \cite{Lit21} uses neighborhood count and Lebesgue contribution to promote solution diversity in the decision and objective spaces, respectively. However, its evaluation involves large computational cost. The work in \cite{IGDX} considers CDX and a probabilistic model to estimate PS and PF but performs poorly when PS is a linear manifold.

Extensive research on EAs for MMMOPs (MMMOEAs) started with decision-niched NSGA-II (DN-NSGA-II) \cite{DN-NSGA2} which replaces the crowding distance in the objective space (CDF) with CDX in NSGA-II. Thus, the diversity of solutions in the objective space receives a significant setback. Another study combines NSGA-II with Weighted Sum Crowding Distance and Neighborhood Based Mutation (NSGA-II-WSCD-NBM) \cite{GECCO19_poster} for addressing MMMOPs. Unlike these preliminary MMMOEAs, MO\_Ring\_PSO\_SCD \cite{MO_Ring_PSO_SCD} establishes that diversity preservation and niching methods (like ring topology) play vital roles for MMMOPs. Although computationally expensive (exponential with the number of decision variables), Zoning search (ZS) \cite{ZoneS} enhance diversity in the decision space. MOEA/D with addition and deletion operators (MOEA/D-AD) \cite{MOEA/D-AD} introduces the notion of \emph{almost same} Pareto-optimal solutions. Its recent extension (ADA) \cite{ADA_2020} proposes adaptive tuning of $K$ (number of equivalent solutions per subspace) for its operation. Multi-Modal Multi-objective Evolutionary Algorithm with Two Archive and Recombination (TriMOEA\_TA\&R) strategy \cite{TriMOEA_TAR} benefits those MMMOPs where a subspace can be extracted from the convergence-related decision variables \cite{TriMOEA_TAR}. It also proposed the notion of differential treatment for diversity in decision and in objective space. Two recent studies: Differential Evolution based algorithm for MMMOPs (DE-TriM) \cite{DETriM} and Multi-Modal Neighborhood sensitive Archived Evolutionary Multi-Objective optimizer (MM-NAEMO) \cite{MMNAEMO_CEC19}, use reference vector assisted decomposition of objective space and adaptive reproduction strategies. However, these MMOEAs have inferior performance in the objective space as compared to the standard MOEAs. Earlier in 2019, a Niching Indicator based Multi-Modal Many-Objective optimizer (NIMMO) \cite{NIMMO_SWEVO19} demonstrated its performance on a few Multi-Modal Many-Objective Problems (MMMaOPs). However, NIMMO \cite{NIMMO_SWEVO19} investigated its performance only on MMMaOPs \cite{POLY_PROBS} with $2$-dimensional decision space.

Thus, several MMMOEAs \cite{MO_Ring_PSO_SCD,MMNAEMO_CEC19,DN-NSGA2,ZoneS} exhibit poor convergence and diversity in the objective space and have been tested only on non-scalable problems (with small values of $N$, $M$ or $k_{PS}$), which motivate further design of better MMMOEAs for problems with high number of variables ($N$), objectives ($M$) and subsets of PS ($k_{PS}$).

The framework proposed in this article is called graph \underline{L}aplacian based \underline{O}ptimization using \underline{R}eference vector assisted \underline{D}ecomposition (LORD) and is used for MMMOPs. It is further extended to LORD-II for MMMaOPs. This paper identifies the crowding illusion problem originating from using crowding distance globally over the entire decision space. To reduce its adverse effects, graph Laplacian based clustering (spectral clustering) is used to decompose the decision space while reference vector based approach is used to decompose the objective space. Diversity preservation is conducted in each decomposed sub-region in a collaborative manner. This divide-and-conquer approach uses adaptive hyper-parameters which make the proposed frameworks adaptive to problem characteristics. Performance analysis with the proposed frameworks using the test functions from CEC 2019 competition MMMOPs \cite{CEC19_TR} and polygon problems \cite{POLY_PROBS} establish their efficacy. 

In the rest of the paper, Section \ref{sec2} mentions the motivation behind the proposed work, Section \ref{sec3} outlines the proposed evolutionary frameworks, Section \ref{sec:Exp} presents the experiments to establish their efficacy and Section \ref{sec5} concludes the article with scope of future work in this direction. 

\section{Motivation for the proposed approach}
 \label{sec2}
This section illustrates the crowding illusion problem and discusses a few observations which motivates the design of the proposed approach.
\subsection{Crowding illusion problem}
 \label{sec_illusion}
\begin{figure}[!t]
	\centering
	\subfloat[Decision Space (MMF3)]{\includegraphics[width=0.49\linewidth]{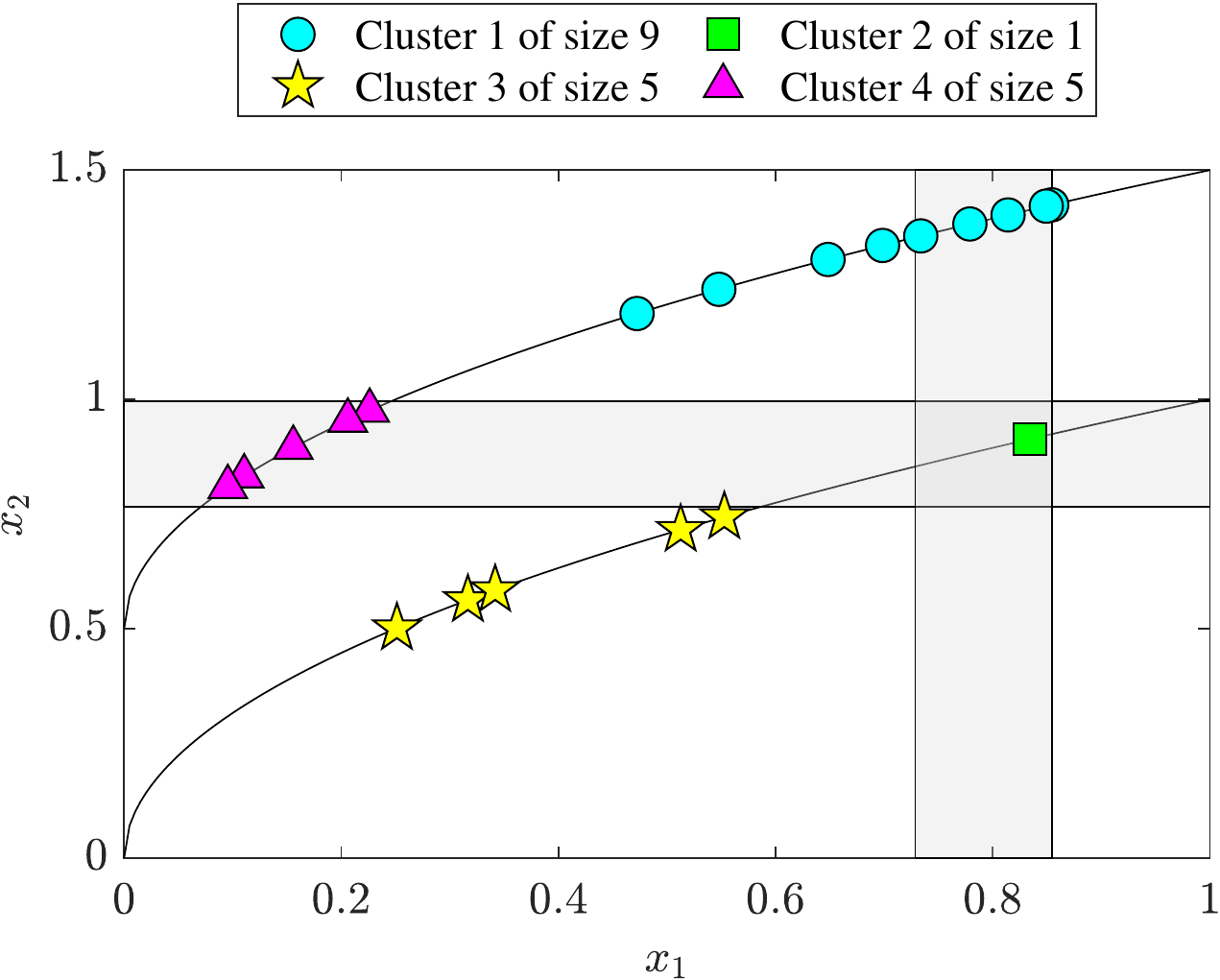}%
		\label{fig:PS_CDX}}
	\hfil
	\subfloat[Objective Space (MMF3)]{\includegraphics[width=0.49\linewidth]{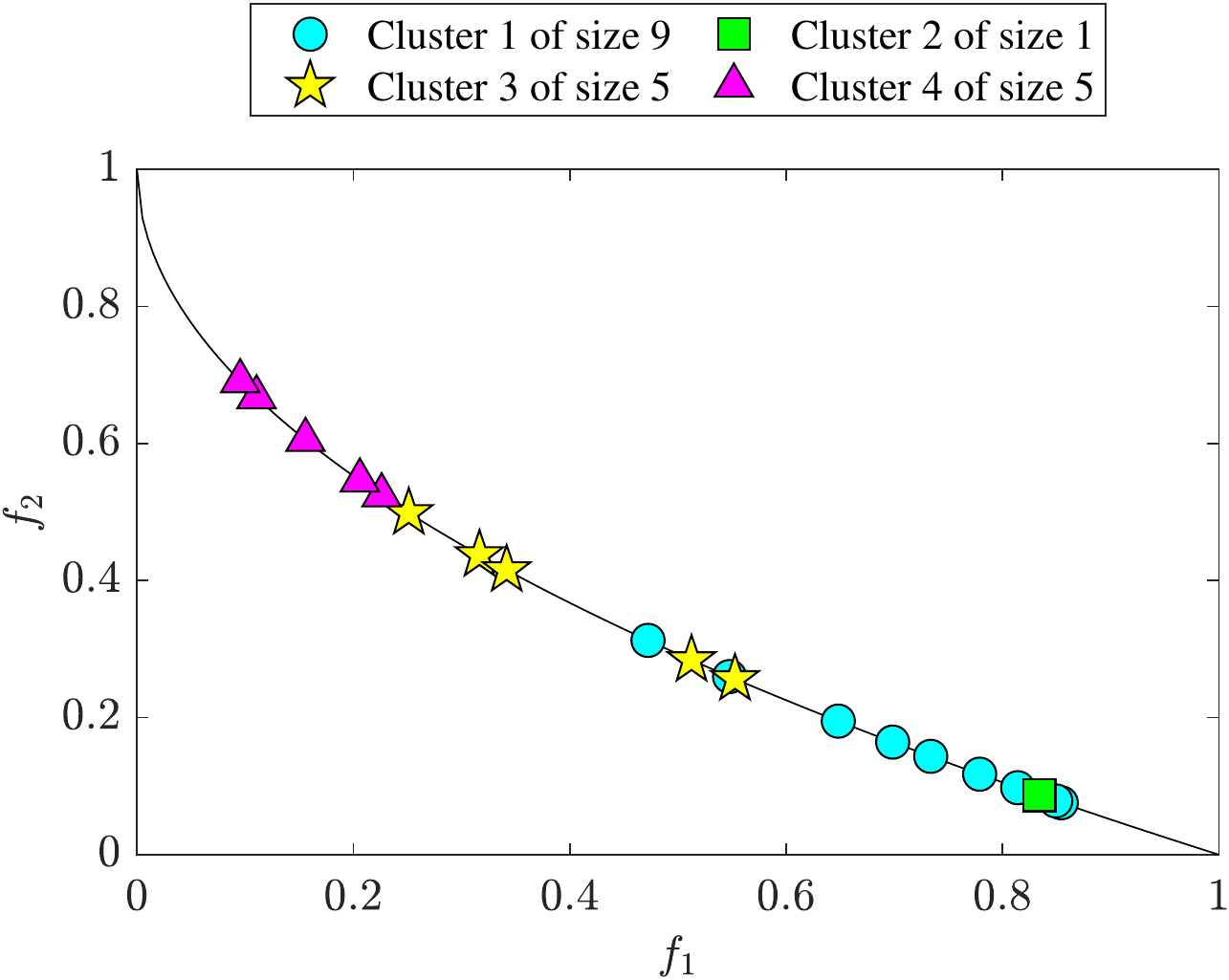}%
		\label{fig:PF_CDX}}
	\caption{Crowding illusion problem in a benchmark test problem (MMF3 \cite{CEC19_TR}) arises due to overlap along different dimensions of the decision space which gives the illusion that \textcolor{green}{$\blacksquare$} is crowded. Usual sorting of solutions from least crowded to most crowded generates   \protect\includegraphics[width=0.45\linewidth]{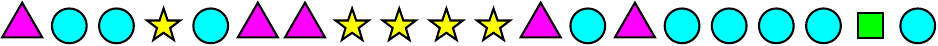} (with \textcolor{green}{$\blacksquare$} solution in $19^\text{th}$ position) whereas LORD\textquoteright s sorting generates \protect\includegraphics[width=0.45\linewidth]{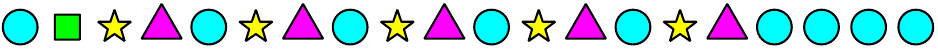} (with \textcolor{green}{$\blacksquare$} solution in $2^\text{nd}$ position).}
	\label{fig:drawback_CDX}
\end{figure}

Most MMMOEAs \cite{Omni,DN-NSGA2,MO_Ring_PSO_SCD,DETriM,GECCO19_poster,IGDX} use CDX to assess the solution distribution. However, using CDX over the entire decision space can be illusional. To describe this issue, let the example in Fig. \ref{fig:drawback_CDX} be considered. It has an isolated \textcolor{green}{$\blacksquare$} solution in the estimated PS. However, due to overlap along different dimensions of the decision space, \textcolor{green}{$\blacksquare$} has nearby neighbors in both objective and decision space impacting the evaluation of CDX and CDF (perimeter of hyper-rectangle bounded by neighbors \cite{MO_Ring_PSO_SCD,DETriM,Omni,DN-NSGA2}). Thus, by the crowding distance-based sorting approach of \cite{MO_Ring_PSO_SCD,DETriM,Omni,DN-NSGA2}, this \textcolor{green}{$\blacksquare$} solution appears towards at the end of the sorted list as a more crowded solution. This ambiguity arising due to the use of CDX globally over the entire decision space is being termed as the crowding illusion problem, henceforth.

\subsection{Observations which motivate the design of the proposed approach}
The following aspects motivate the design of LORD and LORD-II:
\begin{enumerate}
	\item As there is no established formulation for the solution diversity in the decision space, it is either denoted by the solution distribution \cite{MO_Ring_PSO_SCD,DETriM} or by the number of optimal solutions \cite{TriMOEA_TAR}. Thus, LORD and LORD-II characterize the solution diversity in the decision space using both the number and distribution of solutions.
	
	\item To reduce the effect of the crowding illusion problem (Section \ref{sec_illusion}), LORD clusters a set of non-dominated solutions and, then, computes crowding distance within each cluster. For the example in Fig. \ref{fig:drawback_CDX}, the \textcolor{green}{$\blacksquare$} solution appears as a much less crowded solution by using this sorting approach.
	
	\item To yield competitive performance in objective space as the standard MOEAs, unlike other MMMOEAs, LORD demonstrates the synergism of diversity preservation, adaptation of hyper-parameters, reference-vector based decomposition of the objective space, and utilization of the neighborhood property \cite{NAEMO} during mating pool formation and candidate selection.
\end{enumerate}

To the best of the authors' knowledge, the proposed work is the first of its kind to use graph Laplacian (a decomposition strategy) in the decision space along with reference vector assisted decomposition of objective space for handling MMMOPs. Moreover, the proposed framework is also the first to study its scalability on problems in terms of $N$, $M$ and $k_{PS}$. Herein, lies the novelty of the proposed work. 

\section{Proposed Algorithmic Framework}
 \label{sec3}
This section outlines the overall framework of LORD while highlighting the major contributions in the population filtering step (line \ref{s:filter} of Algorithm \ref{alg:framework}).

\subsection{General Framework}
\begin{algorithm}[!t]
	\begin{algorithmic}[1]
		\Require $prob(N, M)$: An MMMOP having $N$-dimensional decision space (lower-bounded by $\mathbf{X}^L$ and upper-bounded by $\mathbf{X}^U$) and $M$-dimensional objective space; $n_{pop}$: Population size; $MaxFES$: Maximal of fitness evaluations; $\mathcal{W}$: Set of $n_{dir}$ reference vectors (as in \cite{NBI,MOEADD})
		\Ensure $\mathcal{A}_{G_{max}}$: Final population estimating PS; $\mathcal{A}_{\mathbf{F},G_{max}}$: Objective vectors $\forall\mathbf{X}\in\mathcal{A}_{G_{max}}$ estimating PF
		\Procedure {LORD}{$prob$, $n_{pop}$, $MaxFES$, $\mathcal{W}$}
		\State \parbox[t]{\dimexpr0.85\columnwidth-\algorithmicindent}{Set $\mathcal{A}_{G}$, $\mathcal{N}$, $F^{DE}_{m,G}$, $CR_{m,G}$, $\eta_{cm,G}$, for $G=1$ \strut} \label{s:Init}
		\For{$G = 1$ to $G_{max}$} \label{s:gen_start}
		\State $\mathbf{S}_{F^{DE}}\gets\emptyset$, $\mathbf{S}_{CR}\gets\emptyset$, $\mathbf{S}_{\eta_c}\gets\emptyset$
		\For{$k = 1$ to $n_{dir}$ (for each direction)}
		\State \parbox[t]{\dimexpr0.8\columnwidth-\algorithmicindent}{$\left[\mathbf{X}_{child}\right.$, $\mathbf{X}_1$, $F^{DE}$, $CR$, $\left.\eta_c\right]\gets$ PERTURB($\mathcal{A}_G$, $\mathbf{N}_k$,$F^{DE}_{m,G}$,$CR_{m,G}$,$\eta_{cm,G}$,$\mathbf{W}_k$,$P_{mut}$)\strut} \label{s:perturb}
		\If{$\mathbf{X}_{1}\nprec \mathbf{X}_{child}$}
		\State $\mathcal{A}_{G}\gets$ FILTER($\mathcal{A}_G$, $\mathbf{X}_{child}$) \label{s:filter}
		\If{$\mathbf{X}_{child}\in\mathcal{A}_{G}$} \label{s:append_starts}
		\State \parbox[t]{\dimexpr0.7\columnwidth-\algorithmicindent}{$\mathbf{S}_{F^{DE}}\gets \mathbf{S}_{F^{DE}}\cup F^{DE}$, $\mathbf{S}_{CR}\gets \mathbf{S}_{CR}\cup CR$, $\mathbf{S}_{\eta_c}\gets \mathbf{S}_{\eta_c} \cup \eta_c$\strut} \label{s:append}
		\EndIf \label{s:append_ends}
		\EndIf
		\EndFor \label{s:gen_ends}
		\State \parbox[t]{\dimexpr0.7\columnwidth-\algorithmicindent}{$F^{DE}_{m,G+1}\gets mean(\mathbf{S}_{F^{DE}})$,\\ $CR_{m,G+1}\gets mean(\mathbf{S}_{CR})$,\\ $\eta_{cm,G+1}\gets mean(\mathbf{S}_{\eta_c})$\strut} \label{s:parameter_update}
		\EndFor
		\State \Return $\mathcal{A}_{G_{max}}$ and $\mathcal{A}_{\mathbf{F},G_{max}}=\{\mathbf{F}(\mathbf{X})|\mathbf{X}\in\mathcal{A}_{G_{max}}\}$ \label{s:return}
		\EndProcedure
	\end{algorithmic}
	\caption{General Framework}
	\label{alg:framework}
\end{algorithm}

The overall framework is described in terms of the following features:

\textbf{Input and Output:}

LORD (Algorithm \ref{alg:framework}) considers the problem description ($prob(N, M)$), population size ($n_{pop}$), maximum function evaluations ($MaxFES$) and the set of reference-vectors ($\mathcal{W}$ satisfying Eq. \eqref{eq:ref_vec} to decompose the objective space) as input. It estimates PS and PF as the output. For reference vector assisted decomposition of the objective space, Das and Dennis\textquoteright s approach \cite{NBI,NAEMO} is used. Its major blocks are outlined next.
\begin{equation}
\begin{gathered}
\mathcal{W}=[\mathbf{W}_1, \mathbf{W}_2, \cdots, \mathbf{W}_{n_{dir}}]^T, \\
\text{where }\mathbf{W}_i=[w_{i1}, w_{i2}, \cdots, w_{iM}] \text{ and } \sum_{j=1}^{M}w_{ij}=1, \text{ for } i = 1 \text{ to }n_{dir}
\end{gathered}
\label{eq:ref_vec}
\end{equation}

\textbf{Initialization:}

During initialization (line \ref{s:Init}), the population ($\mathcal{A}_{G=1}$) is formed with $n_{pop}$ candidates randomly scattered over $\mathcal{D}$ using Eq. \eqref{eq11:Init}.
\begin{equation}
\begin{gathered}
\mathcal{A}_{G=1} = \left[\mathbf{X}_1,\cdots,\mathbf{X}_{n_{pop}}\right]^T \text{, where } \mathbf{X}_{i} = \left[x_{i1}, \cdots, x_{iN}\right]\\ 
\text{and } x_{ij}=x_j^L+rand(0,1)\times\left(x_j^U-x_j^L\right)\text{ for }j =1,\cdots,N
\end{gathered}
\label{eq11:Init}
\end{equation}

The mean values of reproduction parameters ($F^{DE}_{m,G=1}$, $CR_{m,G=1}$ and $\eta_{cm,G=1}$) are initialized. For the $k^\text{th}$ reference vector ($\mathbf{W}_k$), the indices of other reference vectors are stored in the $k^\text{th}$ row of the neighborhood lookup matrix, $\mathbf{N}_k\in\mathcal{N}$, sorted by distance from $\mathbf{W}_k$. As $\mathbf{W}_k$ defines a subspace and $\mathbf{N}_k$ points to its neighbors, $\mathcal{N}$ plays a vital role in mating pool formation to promote diversity.

\textbf{Generations of LORD:}

The for-loop (lines \ref{s:gen_start} to \ref{s:gen_ends}) executes different generations of LORD until $G_{max}=\lfloor MaxFES/n_{dir} \rfloor$. Each generation $G$ iterates over all $n_{dir}$ sub-spaces. Within one iteration, solution perturbation (line \ref{s:perturb}) and population filtering (line \ref{s:filter}) are performed, as described in the next paragraphs. If the child candidate $\mathbf{X}_{child}$ survives the filtering step, the reproduction parameters involved in its creation are appended to respective success vectors ($\mathbf{S}_{F^{DE}}$, $\mathbf{S}_{CR}$ and $\mathbf{S}_{\eta_c}$) in steps \ref{s:append_starts} to \ref{s:append_ends}. When $G$ ends, the mean of reproduction parameters are updated in line \ref{s:parameter_update} using respective success vectors. The population ($\mathcal{A}_{G_{max}}$) at the end of $G_{max}$ generations estimates the PS and the set $\mathcal{A}_{\mathbf{F}, G_{max}}$ of corresponding objective vectors represents the estimated PF.

\textbf{Solution Perturbation:}
\begin{algorithm}[!t]
	\begin{algorithmic}[1]
		\Require $\mathcal{A}_G$: Population; $\mathbf{N}_k$: Mating pool; \{$F^{DE}_{m,G}$, $CR_{m,G}$, $\eta_{cm,G}$\}: Reproduction parameters; $\mathbf{W}_k$: $k^\text{th}$ reference vector; $P_{mut}$: Probability of mutation switching
		\Ensure $\mathbf{X}_{child}$: Child; \{$F^{DE}$, $CR$, $\eta_c$\}: Reproduction parameters used
		\Procedure {PERTURB}{$\mathcal{A}_G$, $\mathbf{N}_k$, $F^{DE}_{m,G}$, $CR_{m,G}$, $\eta_{cm,G}$, $\mathbf{W}_k$, $P_{mut}$}
		\If{no candidate is associated with $\mathbf{W}_k$} 
		\State $\mathbf{N}' \gets$ First $k_{nbr}$ non-empty vectors from $\mathbf{N}_k$ \label{s:line_3}
		\State $\mathbf{W}_{r} \gets$ Reference vector for random index $r\in\mathbf{N}'$
		\State $\mathbf{X}_1 \gets$ Random candidate associated with $\mathbf{W}_{r}$ \label{s:line_5}
		\Else
		\State $\mathbf{X}_1 \gets$ Random candidate associated with $\mathbf{W}_k$ \label{s:assoc}
		\EndIf
		\If{$rand(0,1) < P_{mut}$} \label{s:line_9}
		\State $\mathcal{A}^{mat}_{k,G}\gets$ MATING\_POOL($\mathbf{N}_k$, $\mathcal{A}_{G}$, 1) \label{s:line_10}
		\State $\eta_c \gets N\left(\eta_{cm,G}, 5\right)$ \label{s:line_11}
		\State $\mathbf{X}_2 \gets$ Randomly from $\mathcal{A}^{mat}_{k,G}$
		\State $\mathbf{X}_{child}' \gets$ SBX-crossover \cite{NSGA3} with $\mathbf{X}_1$, $\mathbf{X}_2$, $\eta_c$ 
		\State $\mathbf{X}_{child} \gets$ Polynomial mutation \cite{polymut_recentref} on $\mathbf{X}_{child}'$ \label{s:line_14}
		\State $F^{DE}\gets\emptyset$, $CR\gets\emptyset$
		\Else \label{s:line_16}
		\State $\mathcal{A}^{mat}_{k,G}\gets$ MATING\_POOL($\mathbf{N}_k$, $\mathcal{A}_{G}$, 3) \label{s:line_17}
		\State $F^{DE} \gets N\left(F^{DE}_{m,G}, 0.1\right)$, $CR \gets N\left(CR_{m,G}, 0.1\right)$ \label{s:line_18}
		\State $\left[\mathbf{X}_2, \mathbf{X}_3, \mathbf{X}_4\right] \gets$ Randomly from $\mathcal{A}^{mat}_{k,G}$
		\State \parbox[t]{\dimexpr0.8\columnwidth-\algorithmicindent}{$\mathbf{X}_{child}' \gets$ DE/rand/1/bin \cite{Sys_DE} with $\mathbf{X}_1$ to $\mathbf{X}_4$, $F^{DE}$, $CR$ \strut}
		\State $\mathbf{X}_{child} \gets$ Polynomial mutation \cite{polymut_recentref} on $\mathbf{X}_{child}'$ \label{s:line_21}
		\State $\eta_c\gets\emptyset$
		\EndIf
		\State \Return $\mathbf{X}_{child}$, $\mathbf{X}_{1}$, $F^{DE}$, $CR$, $\eta_c$ \label{s:line_24}
		\EndProcedure
	\end{algorithmic}
	\caption{Reproduction of Child Candidate}
	\label{alg:reproduction}
\end{algorithm}

To make LORD adaptive to the local properties of the fitness landscape, probabilistic mutation switching \cite{NAEMO,RSA} is used. The creation of $\mathbf{X}^{child}$ in line \ref{s:perturb} of Algorithm \ref{alg:framework} uses Algorithm \ref{alg:reproduction}. The first parent $\mathbf{X}_1$ is randomly chosen from the candidates associated with $\mathbf{W}_k$ (line \ref{s:assoc}). This association is dictated by Eq. \eqref{eq:sub-space} where $d2$ is the minimum perpendicular distance from $\mathbf{F}(\mathbf{X})$ to $\mathbf{W}_k$. 
\begin{equation}
\begin{gathered}
\text{Sub-space associated to }\mathbf{W}_k: \{\mathbf{F}(\mathbf{X})\in\mathbb{R}^M | d2\left(\mathbf{X}|\mathbf{W}_k\right) \leq d2\left(\mathbf{X}|\mathbf{W}_j\right)\},\\
\text{for }j=\{1, 2,\cdots,n_{dir}\}\text{, } k\neq j \text{ and } \mathbf{X}\in \mathcal{D}.
\end{gathered}
\label{eq:sub-space}
\end{equation}

The other remaining parents (in line \ref{s:line_10} or \ref{s:line_17}) and also $\mathbf{X}_1$ (if the $k^\text{th}$ sub-space is empty in lines \ref{s:line_3} to \ref{s:line_5}) are randomly chosen using the mating pool ($\mathcal{A}^{mat}_{k,G}$) formation principle (elaborated in next paragraph). The parameter $P_{mut}$ decides between DE/rand/1/bin \cite{DEMO,Sys_DE} or SBX crossover \cite{NSGA3,DN-NSGA2} (in line \ref{s:line_9} or \ref{s:line_16}). The reproduction parameters ($\eta_c$, $F^{DE}$ and $CR$) are sampled from Gaussian distributions \cite{SaDE,NAEMO} with mean values provided by $\eta_{cm,G}$, $F^{DE}_{m,G}$ and $CR_{m,G}$. The standard deviations of the distributions are varied in the range $\left[0.05, 0.2\right]$ for $F^{DE}$ and $CR$ and $\left[1, 6\right]$ for $\eta_c$ and the best-performing values are specified in lines \ref{s:line_11} and \ref{s:line_18}. For SBX crossover, one of two children is randomly considered (in line 13). Both SBX crossover and DE/rand/1/bin are followed by polynomial mutation \cite{polymut_recentref} in lines \ref{s:line_14} and \ref{s:line_21}, respectively, as it helps to avoid local optima \cite{polymut_recentref}. The child candidate ($\mathbf{X}_{child}$), the parent candidate ($\mathbf{X}_1$) and the sampled values of reproduction parameters are returned in line \ref{s:line_24}. Depending on the choice of if-condition in line \ref{s:line_9} of Algorithm \ref{alg:reproduction}, either $\eta_c$ or $F^{DE}$ and $CR$ are empty so only used values of reproduction parameters are appended to the success vectors in line \ref{s:append} of Algorithm \ref{alg:framework}. 

\textbf{Mating Pool Formation:}
\begin{algorithm}[!t]
	\begin{algorithmic}[1]
		\Require $\mathbf{N}_k$: Sorted array of nearest neighboring directions of $\mathbf{W}_k$; $\mathcal{A}_G$: Population in decision space; $n_{\mathcal{S}}$: Number of sub-spaces to be chosen
		\Ensure $\mathcal{A}^{mat}_{k,G}$: Sub-population selected for mating
		\Procedure {MATING\_POOL}{$\mathbf{N}_k$, $\mathcal{A}_G$, $n_{\mathcal{S}}$}
		\State $\mathbf{N}' \gets$ First $k_{nbr}$ non-empty vectors from $\mathbf{N}_k$ \label{s:k_nbr}
		\State \parbox[t]{\dimexpr0.85\columnwidth-\algorithmicindent}{$\{\mathbf{W}_{r_1}, \cdots, \mathbf{W}_{r_{n_{\mathcal{S}}}}\} \gets$ Reference vectors for random indices $\{r_1,\cdots,r_{n_{\mathcal{S}}}\}\in\mathbf{N}'$\strut} \label{s:nbr_subspace}
		\State \parbox[t]{\dimexpr0.85\columnwidth-\algorithmicindent}{$\mathcal{A}^{mat}_{k,G} \gets$ Candidates of $\mathcal{A}_G$ associated with $\{\mathbf{W}_{r_1}, \cdots, \mathbf{W}_{r_{n_{\mathcal{S}}}}\}$\strut} \label{s:pool_form}
		\State \Return $\mathcal{A}^{mat}_{k,G}$ \label{s:pool_return}
		\EndProcedure
	\end{algorithmic}
	\caption{Mating Pool Formation}
	\label{alg:mating_pool}
\end{algorithm}

The idea is to leverage the neighborhood property \cite{NAEMO} for a uniform exploration, i.e., higher chances of $\mathbf{X}^{child}$ creation in an unexplored subspace occurs if candidates from its neighboring regions participate in mating \cite{NAEMO,MOEADD}. The mating pool ($\mathcal{A}^{mat}_{k,G}$) formation in line \ref{s:line_10} or \ref{s:line_17} of Algorithm \ref{alg:reproduction} uses Algorithm \ref{alg:mating_pool}. It considers $k_{nbr}$ nearest non-empty reference vectors of $\mathbf{W}_k$ (line \ref{s:k_nbr}), from which $n_\mathcal{S}$ random reference vectors $\{\mathbf{W}_{r_1}, \cdots, \mathbf{W}_{r_{n_{\mathcal{S}}}}\}$ are selected in line \ref{s:nbr_subspace}. The parameter $n_\mathcal{S}$ is the minimum number of additional parents required as per a reproduction strategy (3 for DE/rand/1/bin and 1 for SBX crossover). All candidates associated with $\{\mathbf{W}_{r_1}, \cdots, \mathbf{W}_{r_{n_{\mathcal{S}}}}\}$ form $\mathcal{A}^{mat}_{k,G}$ in line \ref{s:pool_form} and returned from line \ref{s:pool_return}. For association, Eq. \eqref{eq:sub-space} is considered as done in \cite{MOEADD,NSGA3}.

\textbf{Population Filtering:}

If $\mathbf{X}_{child}$ is better than its parent ($\mathbf{X}_1$), to maintain a constant $n_{pop}$, one of the candidates from $\mathcal{A}_G\cup\mathbf{X}^{child}$ is removed in line \ref{s:filter} of Algorithm \ref{alg:framework} by calling the filtering operation, which is described after explaining the approach to decompose the population in the decision space.

\subsection{Decomposition of the Decision Space}
\label{ss:decompose_DS}
The filtering step in line \ref{s:filter} of Algorithm \ref{alg:framework} involves graph Laplacian based partitioning \cite{SpecClus,Aparajita_SpecClus} of a set of solutions ($\mathcal{A}^{nd}$) in the decision space. This graph partitioning helps in picking up solutions from each cluster to yield a balanced solution set and to reduce the effect of crowding illusion during diversity estimation. Spectral graph theory is associated with studying properties of a graph using eigen decomposition of its Laplacian matrix representation \cite{SpecClus,Aparajita_SpecClus}. These properties assist in the spectral clustering of $\mathcal{A}^{nd}$ through the following steps:

1) \emph{Create nearest neighbor graph ($\mathcal{G}$):} All candidates of $\mathcal{A}^{nd}$ are used as the nodes of graph $\mathcal{G}$. Euclidean distances between all pairs of candidates in $\mathcal{A}^{nd}$ are evaluated. Edges are placed between pairs of candidates (nodes) where distance is less than a threshold of $\varepsilon_L$. Specifically, $\mathcal{G}$ (binary symmetric matrix) is the adjacency matrix representation.

2) \emph{Obtain symmetric normalized graph Laplacian ($\mathcal{L}_{sym}$):} A diagonal matrix $\mathcal{G}_d$ is created using the degree of each node (row sum) of $\mathcal{G}$. Using the identity matrix $I$ of the same order as $\mathcal{G}$ and $\mathcal{G}_d$, $\mathcal{L}_{sym}$ \cite{SpecClus} is obtained by Eq. \eqref{eq:Lsym}.
\begin{equation}
\mathcal{L}_{sym} = I - \mathcal{G}_d^{-1/2}\mathcal{G}\mathcal{G}_d^{-1/2}
\label{eq:Lsym}
\end{equation}

3) \emph{Obtain number of connected components ($k_\mathcal{CC}$):} The algebraic multiplicity of $0$ eigen value of $\mathcal{L}_{sym}$ \cite{SpecClus,Aparajita_SpecClus} gives the number of connected components ($k_\mathcal{CC}$) of $\mathcal{G}$.

4) \emph{Assign candidates (nodes) to $k_\mathcal{CC}$ clusters:} By Cheeger's inequality \cite{Cheeger1,Cheeger2}, the sparsest cut of a graph is approximated by the second smallest eigenvalue of $\mathcal{L}_{sym}$. Thus, all the eigen vectors from the second smallest to the $k_\mathcal{CC}^\text{th}$ eigenvalues are clustered ($\mathcal{C}_1, \cdots, \mathcal{C}_{k_\mathcal{CC}}$) using k-means \cite{kmeans} for assigning the candidates of $\mathcal{A}^{nd}$ to the clusters (partitions) in the decision space. An example of spectral clustering a non-dominated set of solutions is illustrated in Fig. \ref{fig:PS_CDX} for a benchmark test problem (MMF3) \cite{CEC19_TR}. 

For reducing crowding illusion (Section \ref{sec_illusion}), spectral clustering of $\mathcal{A}^{nd}$ is chosen over k-means clustering due to the following reasons: (1) k-means is effective only for globular structures whereas spectral clustering is effective for non-globular structure as well \cite{spec_versus_kmeans} (a comparison is provided in Section 5 of the supplementary material at \url{https://worksupplements.droppages.com/lord}), (2) $k_{\mathcal{CC}}$ for k-means is not known apriori whereas $k_{\mathcal{CC}}$ for spectral clustering can be obtained mathematically and (3) k-means (performed in the step 4 of spectral clustering of $\mathcal{A}^{nd}$) becomes independent of the number of decision variables ($N$).

\subsection{Filtering Scheme of LORD framework}
\begin{algorithm}[!t]
	\begin{algorithmic}[1]
		\Require $\mathcal{A}_G$: Current population; $\mathbf{X}_{child}$: Child candidate
		\Ensure $\mathcal{A}_{G}$: Filtered population of size $n_{pop}$; 
		\Procedure {FILTER}{$\mathcal{A}_G$, $\mathbf{X}_{child}$}\label{a4l1}
		\State $\mathcal{A}^{all}_{\mathbf{F},G}=\{\mathbf{F}(\mathbf{X})|\mathbf{X}\in(\mathcal{A}_{G}\cup\mathbf{X}_{child})\}$\label{a4l2}
		\State $\mathcal{A}^{nd}_{\mathbf{F}} \gets$ Last non-dominated rank of $\mathcal{A}^{all}_{\mathbf{F},G}$\label{a4l3}
		\State $\mathcal{A}^{nd}=\{\mathbf{X}|\mathbf{F}(\mathbf{X})\in\mathcal{A}^{nd}_{\mathbf{F}}\}$\label{a4l4} 
		\State $\{\mathcal{C}_1, \cdots, \mathcal{C}_{k_\mathcal{CC}}\} \gets$ Spectral clustering of $\mathcal{A}^{nd}$\label{a4l5}
		\State Evaluate SCD cluster-wise \label{a4l6}
		\State $\mathcal{A}^{nd}_{s} \gets $ Select one-by-one from $\mathcal{C}_1$ to $\mathcal{C}_{k_\mathcal{CC}}$ w.r.t. SCD \label{a4l7}
		\For{$j=\left\vert \mathcal{A}^{nd}_s\right\vert$ to $1$ (starting from most-crowded)} \label{a4l8}
		\State $\mathbf{W}_k \gets$ Direction where $\mathbf{X}_j\in\mathcal{A}^{nd}_s$ is associated\label{a4l9}
		\If{\#candidates associated with $\mathbf{W}_k > 1$}\label{a4l10}
		\State $\mathbf{X}_{del}\gets$ Assign $\mathbf{X}_j$ for deletion\label{a4l11}
		\State Break loop\label{a4l12}
		\EndIf\label{a4l13}
		\EndFor\label{a4l14}
		\If{no $\mathbf{X}_{del}$ is chosen}\label{a4l15}
		\State $\mathbf{X}_{del}\gets$ Last candidate of $\mathcal{A}^{nd}_s$\label{a4l16}
		\EndIf\label{a4l17}
		\State $\mathcal{A}_{G} \gets (\mathcal{A}_{G} \cup \mathbf{X}_{child})-\mathbf{X}_{del}$\label{a4l18}
		\State \Return $\mathcal{A}_{G}$\label{a4l19}
		\EndProcedure
	\end{algorithmic}
	\caption{Filter for constant Population Size (LORD)}
	\label{alg:filter1}
\end{algorithm}
\begin{figure*}[!t]
	\centering
	\includegraphics[width=\linewidth]{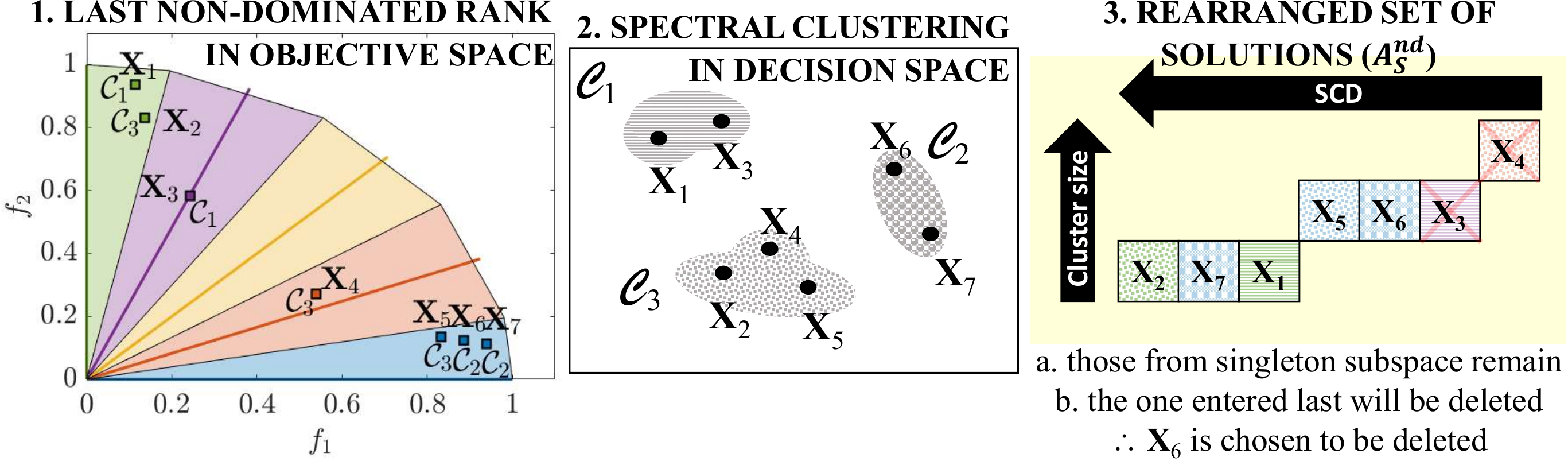}
	\caption{Filtering scheme of LORD on the last non-dominated rank ($\mathcal{A}^{nd}$).}
	\label{fig:LORD}
\end{figure*}
The filtering steps of the LORD framework (Algorithm \ref{alg:filter1}) consists of three major steps as follows:
\begin{enumerate}
	\item \emph{Obtain last non-dominated rank (maintaining convergence in objective space):} Using non-dominated sorting on $\left(\mathcal{A}_G\cup\mathbf{X}_{child}\right)$, the solutions ($\mathcal{A}^{nd}$) in the last non-dominated rank \cite{MOEADD,SS_ndSort} are obtained in lines \ref{a4l3} to \ref{a4l4}. If $\left\vert\mathcal{A}^{nd}\right\vert=1$, lines 5 to 17 yield the only $\mathbf{X}_{del}\in\mathcal{A}^{nd}$ for elimination. Otherwise, some $\mathbf{X}_{del}\in\mathcal{A}^{nd}$ (i.e., from the least converged set of mutually non-dominated candidates) is eliminated by the next steps.
	
	\item \emph{Spectral clustering of candidates from $\mathcal{A}^{nd}$ (maintaining diversity in decision space):} The candidates in $\mathcal{A}^{nd}$ is partitioned in line \ref{a4l5} using the steps mentioned in Section \ref{ss:decompose_DS} as shown in Fig. \ref{fig:LORD}. Evaluating the crowding in the respective spaces, Special Crowding Distance (SCD) \cite{MO_Ring_PSO_SCD,DETriM} combines CDF and CDX. SCD is evaluated per cluster in line \ref{a4l6}. A sorted set ($\mathcal{A}^{nd}_s$) of candidates is formed by rearranging $\mathcal{A}^{nd}$ in line \ref{a4l7} where at first the candidates with the highest SCD is selected from each cluster (e.g., $\mathbf{X}_2\in\mathcal{C}_3$, $\mathbf{X}_7\in\mathcal{C}_2$ and $\mathbf{X}_1\in\mathcal{C}_1$ in Fig. \ref{fig:LORD}), then candidates with the second-highest SCD is selected from each cluster (e.g., $\mathbf{X}_5\in\mathcal{C}_3$, $\mathbf{X}_6\in\mathcal{C}_2$ and $\mathbf{X}_3\in\mathcal{C}_1$ in Fig. \ref{fig:LORD}) and so on (e.g., finally $\mathbf{X}_4\in\mathcal{C}_3$).
	
	\item \emph{Association based elimination of candidate from $\mathcal{A}^{nd}_s$ (maintaining diversity in objective space):} Starting from the worst candidate (e.g., $\mathbf{X}_4$) in $\mathcal{A}^{nd}_s$, the reference vector $\mathbf{W}_k$ is obtained in line \ref{a4l9} with which $\mathbf{X}_j\in\mathcal{A}^{nd}_s$ is associated. If multiple candidates of ($\mathcal{A}_{G}\cup\mathbf{X}_{child}$) are associated with $\mathbf{W}_k$ (implying a dense sub-space), $\mathbf{X}_{del}=\mathbf{X}_j$ is chosen for deletion (lines \ref{a4l10} to \ref{a4l13}). But if the associated sub-space is singleton, $\mathbf{X}_j$ is retained (e.g., $\mathbf{X}_4$ and $\mathbf{X}_3$ are retained and $\mathbf{X}_{del}=\mathbf{X}_6$ will be deleted in Fig. \ref{fig:LORD}). If all the sub-spaces with which candidates of $\mathcal{A}^{nd}_s$ are associated are singleton, the last candidate from $\mathcal{A}^{nd}_s$ is chosen for deletion (lines \ref{a4l15} to \ref{a4l17}). $\mathbf{X}_{del}$ is deleted from ($\mathcal{A}_{G}\cup\mathbf{X}_{child}$) in line \ref{a4l18} to yield the filtered $\mathcal{A}_{G}$ for the next iteration. This filtered $\mathcal{A}_{G}$ is returned from line \ref{a4l19} of Algorithm \ref{alg:filter1} to line \ref{s:filter} of main framework (Algorithm \ref{alg:framework}).
\end{enumerate}

Explicit maintenance of the three essential properties is the most important characteristics of LORD as a novel MMMOEA. While SCD explicitly accounts for solution distribution in decision space, the candidates towards the end of $\mathcal{A}^{nd}_s$ come from the larger clusters (e.g., Fig. \ref{fig:drawback_CDX} and Fig. \ref{fig:LORD}) and are more likely to be deleted. Hence, LORD implicitly takes care of the neighborhood count also as a diversity criterion. This general framework (Algorithm \ref{alg:framework}) with the filtering scheme in Algorithm \ref{alg:filter1} is called graph \underline{L}aplacian based \underline{O}ptimization using \underline{R}eference vector assisted \underline{D}ecomposition (LORD) whose performance is assessed in Section \ref{sec:Exp}. 

\subsection{Filtering Scheme of LORD-II framework}
\label{sec3d}
\begin{algorithm}[!t]
	\begin{algorithmic}[1]
		\Require $\mathcal{A}_G$: Current population; $\mathbf{X}_{child}$: Child candidate
		\Ensure $\mathcal{A}_{G}$: Filtered population of size $n_{pop}$; 
		\Procedure {FILTER}{$\mathcal{A}_G$, $\mathbf{X}_{child}$}
		\State $\mathcal{A}^{nd}\gets\mathcal{A}_G \cup \mathbf{X}_{child}$ \label{a5l1}
		\State $\mathcal{A}^{nd}_{\mathbf{F},G}=\{\mathbf{F}(\mathbf{X})|\mathbf{X}\in\mathcal{A}^{nd}\}$ \label{a5l2}
		\State $\mathcal{A}_{del}\gets\emptyset$\label{a5l3}
		\For{$k=1$ to $n_{dir}$ (for each direction)}\label{a5l4}
		\State $\mathcal{A}^{sub}_{\mathbf{F},k}\gets$ Candidates of $\mathcal{A}^{nd}_{\mathbf{F},G}$ associated with $\mathbf{W}_k$\label{a5l5}
		\If{$\left\vert\mathcal{A}^{sub}_{\mathbf{F},k}\right\vert>1$}\label{a5l6}
		\State $\mathcal{A}_{del} \gets \mathcal{A}_{del} \cup (\mathbf{X}$ with max PBI in $\mathcal{A}^{sub}_{\mathbf{F},k})$\label{a5l7}
		\EndIf\label{a5l8} 
		\EndFor\label{a5l9}
		
		\State $[\mathcal{C}_1, \cdots, \mathcal{C}_{k_\mathcal{CC}}] \gets$ Spectral clustering of $\mathcal{A}^{nd}$\label{a5l10}
		\State $I_{del}=0$, $M_{del}=0$\label{a5l11}
		\For{$j=1$ to $k_\mathcal{CC}$ (for all clusters)}\label{a5l12}
		\If{$\mathcal{A}_{del}\cap\mathcal{C}_j\neq\emptyset$}\label{a5l13}
		\If{$M_{del}< \left\vert\mathcal{C}_j\right\vert$}\label{a5l14}
		\State $I_{del}=j$, $M_{del}=\left\vert\mathcal{C}_j\right\vert$\label{a5l15}
		\EndIf\label{a5l16}
		\EndIf\label{a5l17}
		\EndFor\label{a5l18}
		\State $\mathcal{A}''_{del}\gets\mathcal{C}_{I_{del}}\cap\mathcal{A}_{del}$\label{a5l19}
		\State $\mathbf{X}_{del}\gets$ Select candidate with max PBI from $\mathcal{A}''_{del}$\label{a5l20}
		\State $\mathcal{A}_{G}\gets\mathcal{A}^{nd}-\mathbf{X}_{del}$\label{a5l21}
		\State \Return $\mathcal{A}_{G}$\label{a5l22}
		\EndProcedure
	\end{algorithmic}
	\caption{Filter for constant Population Size (LORD-II)}
	\label{alg:filter2}
\end{algorithm}
\begin{figure*}[!t]
	\centering
	\includegraphics[width=\linewidth]{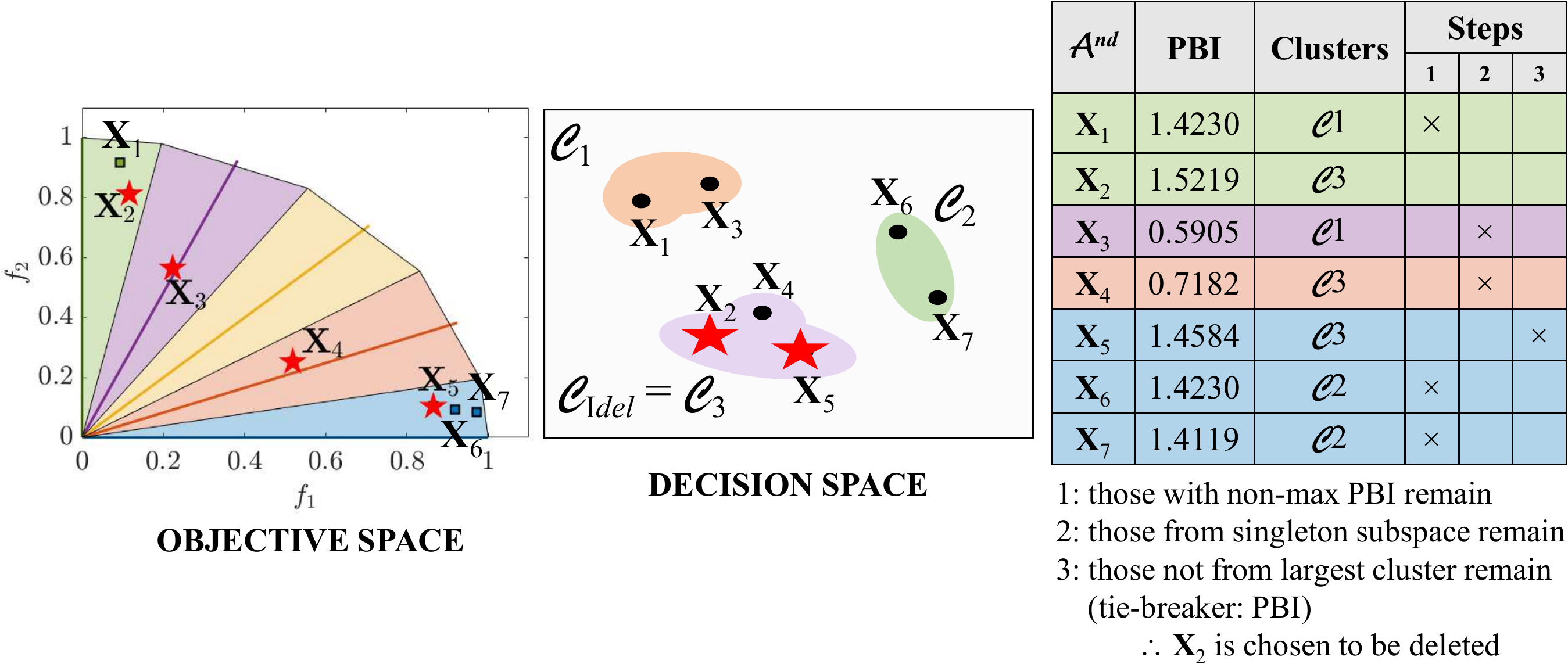}
	\caption{Filtering scheme of LORD-II on a set of solutions ($\mathcal{A}^{nd}$).}
	\label{fig:LORD_I}
\end{figure*}
With higher number of objectives, Pareto-dominance fails to provide the necessary selection pressure \cite{SWEVO_2011,UM_2014TEVCReview_PartI}. Thus, another filtering scheme is presented in Algorithm \ref{alg:filter2} for which convergence is driven by Penalty-based Boundary Intersection (PBI) instead of Pareto-dominance. As PBI based comparisons are specific to each sub-space, it can efficiently provide the necessary selection pressure. This scheme consists of the following steps:
\begin{enumerate}
	\item \emph{PBI-based selection for deletion (maintaining convergence in objective space):} The objective vectors corresponding to all candidates of $\mathcal{A}_G$ and $\mathbf{X}_{child}$ are stored in $\mathcal{A}^{nd}_{\mathbf{F},G}$ in line \ref{a5l2}. From each direction, the candidate with the maximum PBI \cite{MOEADD,NBI,ESOEA} is stored in $\mathcal{A}_{del}$ (lines \ref{a5l3} to \ref{a5l9}) as potential candidates for deletion. In Fig. \ref{fig:LORD_I}, after step 1, $\mathcal{A}_{del}=\left\{ \mathbf{X}_2,\mathbf{X}_3, \mathbf{X}_4, \mathbf{X}_5\right\}$.
	
	\item \emph{Disregarding based on association (maintaining diversity in objective space):} Deletion of a candidate from any sub-space with only one candidate would hamper the diversity in the objective space. Hence, singleton sub-spaces is not considered in $\mathcal{A}_{del}$ (lines \ref{a5l6} to \ref{a5l8}). In Fig. \ref{fig:LORD_I}, after step 2, $\mathcal{A}_{del}=\left\{ \mathbf{X}_2,\mathbf{X}_5\right\}$.
	
	\item \emph{Spectral clustering of candidates (maintaining diversity in decision space):} The candidates in $\mathcal{A}^{nd}$ are partitioned in line \ref{a5l10} using the steps mentioned in Section \ref{ss:decompose_DS}. The cardinality is noted (lines \ref{a5l11} to \ref{a5l18}) only for those clusters which has candidates from the set of potential candidates for deletion $\mathcal{A}_{del}$ (line \ref{a5l13}). Then, the candidate ($\mathbf{X}_{del}$) from the largest cluster ($\mathcal{C}_{I_{del}}$), common to $\mathcal{A}_{del}$, is to be deleted. If multiple such candidates exists (e.g., $\mathbf{X}_2$ and $\mathbf{X}_5$ in Fig. \ref{fig:LORD_I}), the smaller set $\mathcal{A}''_{del}$ is formed in line \ref{a5l19}. The candidate with the largest PBI in $\mathcal{A}''_{del}$ is deleted (lines \ref{a5l20} to \ref{a5l21}) to yield the filtered $\mathcal{A}_{G}$ for the next iteration. Thus, in Fig. \ref{fig:LORD_I}, after step 3, $\mathbf{X}_2$ is deleted.
\end{enumerate}

While the cluster size explicitly accounts for the number of optimal solutions, the spectral clustering implicitly accounts for the solution distribution in the decision space. The general framework (Algorithm \ref{alg:framework}) with this filtering scheme of Algorithm \ref{alg:filter2} is called LORD-II. 

This combination of the proposed operations allows the LORD and LORD-II to effectively address the two challenges mentioned in Section \ref{sec1}. For the first challenge, graph Laplacian based clustering handles the diversity in the decision space and reference vector based decomposition handles the diversity in the objective space. The convergence attribute is addressed differently by LORD (using Pareto-dominance) and LORD-II (using PBI based comparisons). For the second challenge, spectral clustering and cluster cardinality helps in obtaining balanced subsets in the PS.

\section{Experimental Results}
\label{sec:Exp}
For performance analysis, LORD and LORD-II are implemented in MATLAB R2018a using a 64-bit computer (8 GB RAM, Intel Core i7 @ 2.20 GHz). The experimental specifications of the benchmark MMMOPs, performance measures, and parameter settings of the proposed as well as competitor algorithms are provided in the following sub-sections. 
\subsection{Benchmark Problems}
\begin{table}[!t]
	\caption{Specifications for 2-objective MMMOPs in terms of $N$-dimensional decision space, upper and lower bounded between $\mathbf{X}^U$ and $\mathbf{X}^L$ having reference point at $\mathbf{R}_{HV}$ for HV calculation with $N_{IGD}$ number of points in the reference set for IGD evaluation and number of subsets within PS ($k_{PS}$).}
	\label{tab:params1}
	\resizebox{\linewidth}{!}{
		\begin{tabular}{ccccccc}
			\hline
			Problems & $N$ & $\mathbf{X}^L$ & $\mathbf{X}^U$ & $\mathbf{R}_{HV}$ & $N_{IGD}$ & $k_{PS}$\\
			\hline
			MMF1 & 2 & $\left[1, -1\right]$ & $\left[3, 1\right]$ & $\left[1.1, 1.1\right]$ & 400 & 2\\
			MMF1\_z & 2 & $\left[1, -1\right]$ & $\left[3, 1\right]$ & $\left[1.1, 1.1\right]$ & 400 & 2\\
			MMF1\_e & 2 & $\left[1, -20\right]$ & $\left[3, 20\right]$ & $\left[1.1, 1.1\right]$ & 400 & 2\\
			MMF2 & 2 & $\left[0, 0\right]$ & $\left[1, 1\right]$ & $\left[1.1, 1.1\right]$ & 400 & 2\\
			MMF3 & 2 & $\left[0, 0\right]$ & $\left[1, 1.5\right]$ & $\left[1.1, 1.1\right]$ & 400 & 2\\
			MMF4 & 2 & $\left[-1, 0\right]$ & $\left[1, 2\right]$ & $\left[1.1, 1.1\right]$ & 400 & 4\\
			MMF5 & 2 & $\left[1, -1\right]$ & $\left[3, 3\right]$ & $\left[1.1, 1.1\right]$ & 400 & 4\\
			MMF6 & 2 & $\left[1, -1\right]$ & $\left[3, 2\right]$ & $\left[1.1, 1.1\right]$ & 400 & 4\\
			MMF7 & 2 & $\left[1, -1\right]$ & $\left[3, 1\right]$ & $\left[1.1, 1.1\right]$ & 400 & 2\\
			MMF8 & 2 & $\left[-\pi, 0\right]$ & $\left[\pi, 9\right]$ & $\left[1.1, 1.1\right]$ & 400 & 4\\
			
			MMF9 & 2 & $\left[0.1, 0.1\right]$ & $\left[1.1, 1.1\right]$ & $\left[1.21, 11\right]$ & 400& 2\\
			MMF10 & 2 & $\left[0.1, 0.1\right]$ & $\left[1.1, 1.1\right]$ & $\left[1.21, 13.2\right]$ & 400 & 1\\
			MMF11 & 2 & $\left[0.1, 0.1\right]$ & $\left[1.1, 1.1\right]$ & $\left[1.21, 15.4\right]$ & 400 & 1\\
			MMF12 & 2 & $\left[0, 0\right]$ & $\left[1, 1\right]$ & $\left[1.54, 1.1\right]$ & 410 & 1\\
			MMF13 & 3 & $\left[0.1, 0.1, 0.1\right]$ & $\left[1.1, 1.1, 1.1\right]$ & $\left[1.54, 15.4\right]$ & 1250 & 1\\
			MMF14 & $M\geq3$ & $[0,\overset{M}{\cdots},0]$ & $[1,\overset{M}{\cdots},1]$ & $[2.2,\overset{M}{\cdots},2.2]$ & 1250 & 2\\
			MMF14\_a & $M\geq3$ & $[0,\overset{M}{\cdots},0]$ & $[1,\overset{M}{\cdots},1]$ & $[2.2,\overset{M}{\cdots},2.2]$ & 1250 & 2\\
			MMF15 & $M\geq3$ & $[0,\overset{M}{\cdots},0]$ & $[1,\overset{M}{\cdots},1]$ & $[2.5,\overset{M}{\cdots},2.5]$ & 1250 & 1\\
			MMF15\_a & $M\geq3$ & $[0,\overset{M}{\cdots},0]$ & $[1,\overset{M}{\cdots},1]$ & $[2.5,\overset{M}{\cdots},2.5]$ & 1250 & 1\\
			Omni-test & 3 & $\left[0, 0, 0\right]$ & $\left[6, 6, 6\right]$ & $\left[4.4, 4.4\right]$ & 600 & 27\\
			SYM-PART & 2 & $\left[-20, -20\right]$ & $\left[20, 20\right]$ & $\left[4.4, 4.4\right]$ & 396 & 9\\
			simple & & & & & & \\
			SYM-PART & 2 & $\left[-20, -20\right]$ & $\left[20, 20\right]$ & $\left[4.4, 4.4\right]$ & 396 & 9\\
			rotated & & & & & & \\
			\hline
	\end{tabular}}
\end{table}
\begin{table}[!t]
	\centering
	\caption{Specifications for reference vector based decomposition for problems with $M$ objectives and $N$ decision variables.}
	\label{tab:params3}
	\begin{tabular}{cccc}
		\hline
		$M$ & $p_1$ & $p_2$ & $n_{dir}$\\
		\hline
		$2$  & $100N-1$ & $0$ & $100N$\\
		$3$  & $23$ 	& $0$ & $300$\\
		$5$  & $8$  	& $0$ & $495$\\
		$8$  & $5$  	& $2$ & $828$\\
		$10$ & $4$  	& $3$ & $935$\\
		\hline
	\end{tabular}
\end{table}
The benchmark problems from CEC 2019 test suite for MMMOPs \cite{CEC19_TR} are considered with $MaxFES=5000\times N$ and $n_{pop}=100\times N$, as per \cite{CEC19_TR}. Specifications of these MMMOPS are mentioned in Table \ref{tab:params1}. It should be noted that MMF12 has discontinuous PF, hence the number of subsets in the global PS ($k_{PS}$) is one per Pareto-optimal patch. While MMF10-13, MMF15 and MMF15\_a have one global PS but these are multi-modal problems as these have local PSs close to their global PS. As MMF14-15, MMF14\_a and MMF15\_a are scalable in terms of $M$ and $N$, these are considered as MMMaOPs.

The specifications ($p_1$ and $p_2$ divisions in boundary and inside layers, respectively) for defining $n_{dir}$ reference vectors \cite{NBI,MOEADD} are mentioned in Table \ref{tab:params3}. The goal is to satisfy $n_{dir} \approxeq n_{pop} = 100N$.

Performance on the polygon MMMaOPs \cite{POLY_PROBS} with $MaxFES = 10000$ are also analyzed using the specifications from \cite{NIMMO_SWEVO19}.

\subsection{Performance Indicators}
In the objective space, Inverted Generational Distance (IGD) \cite{Coello_Review} and Hypervolume indicator (HV) \cite{HypE} are noted for CEC 2019 MMMOPs \cite{CEC19_TR} with $M=2$ to assess the convergence and diversity of MOEAs \cite{Coello_Review}. The size of the reference sets\footnote{Reference sets are obtained from \url{http://www5.zzu.edu.cn/ecilab/info/1036/1163.htm} for CEC 2019 MMMOPs \cite{CEC19_TR} and from \url{https://sites.google.com/view/nimmopt/} for polygon MMMaOPs \cite{POLY_PROBS}.} ($N_{IGD}$) for IGD evaluation and the reference points ($\mathbf{R}_{HV}$) for HV evaluation are specified in Table \ref{tab:params1} as recommended in \cite{CEC19_TR}. For polygon MMMaOPs, IGD with $N_{IGD}=5000$ is used as recommended in \cite{NIMMO_SWEVO19}. Convergence Metric (CM) \cite{aDEMO} with the same reference set as IGD and $D\_metric$ \cite{pop_dynam_indicators} are used for CEC 2019 MMMOPs \cite{CEC19_TR} with $M\geq3$ to individually assess the convergence and diversity of MOEAs.

In decision space, IGD \cite{IGDX} and Pareto-Set Proximity (PSP) \cite{MO_Ring_PSO_SCD} are used for CEC 2019 MMMOPs \cite{CEC19_TR} with $M=2$ and for polygon MMMaOPs \cite{POLY_PROBS} to assess the performance of MOEAs. For CEC 2019 MMMOPs \cite{CEC19_TR} with $M\geq3$, the fraction of non-contributing solutions\footnote{A non-contributing solution is a solution in the non-dominated set whose removal does not change the IGDX of the non-dominated set.} (NSX) \cite{ARMOEA,IGDNS} and the convergence metric of this non-contributing set (CM\_NSX) are noted. Hereafter, IGDX and IGDF represent IGD values in decision and objective space, respectively, and rHV=1/HV and rPSP=1/PSP are noted such that lower value is the better measure over all the indicators. Brief description of all the performance indicators are provided in Section 2 of the supplementary material.

\subsection{Details of Competitor Algorithms}
As DN-NSGA-II \cite{DN-NSGA2} and MO\_Ring\_PSO\_SCD \cite{MO_Ring_PSO_SCD} are MMMOEAs using non-dominated sorting with CDX, LORD is compared with these two MMMOEAs. For comparison with a standard MOEA outperforming the former MMMOEAs in the objective space, LORD is also compared with NSGA-II \cite{UM_2014TEVCReview_PartI,UM_2014TEVCReview_PartII}. For MMMOPs with $M\geq3$, LORD-II is compared with MO\_Ring\_PSO\_SCD \cite{MO_Ring_PSO_SCD} and a popular decomposition based many-objective EA (MOEA/DD) \cite{MOEADD}.

Other MMMOEAs such as Omni-Optimizer \cite{Omni}, TriMOEA\_TA\&R \cite{TriMOEA_TAR}, MM-NAEMO \cite{MMNAEMO_CEC19}, DE-TriM \cite{DETriM} and NIMMO \cite{NIMMO_SWEVO19}, have demonstrated their effectiveness only for certain kinds of test problems. These MMMOEAs are also compared with LORD and LORD-II on some CEC 2019 MMMOPs \cite{CEC19_TR} and polygon problems \cite{NIMMO_SWEVO19,POLY_PROBS}.

Most of the hyper-parameters of LORD and LORD-II are adaptive in nature, while the rest of them are set as mentioned in Table \ref{tab:specs_LORD}.
\begin{table}[!ht]
	\centering
	\caption{Recommended values of different parameters for LORD and LORD-II.}
	\label{tab:specs_LORD}
	\resizebox{\linewidth}{!}{
		\begin{tabular}{|p{2.2cm}|p{1.7cm}|p{10cm}|}
			\hline
			\textbf{Parameters} & \textbf{Values} & \textbf{Remarks}\\
			\hline
			$k_{nbr}$ & $0.2\times n_{dir}$ & Number of non-empty neighboring directions for mating pool formation (line \ref{s:k_nbr}, Algorithm \ref{alg:mating_pool}) which is easily within $0.2\times n_{dir}$ as all test cases (except MMF12) have regular PFs\\
			\hline
			$P_{mut}$ & $0.25$ & Probability of switching among reproduction methods (line \ref{s:line_9}, Algorithm \ref{alg:reproduction}) and sensitivity is analyzed in Section \ref{sec:sens_ana_LORD}\\
			\hline
			$\varepsilon_L$ & $\alpha_L\times$ diagonal of $\mathcal{D}$ with $\alpha_L=0.2$ & Threshold on inter-solution distance for formation of nearest neighbor graph (Section \ref{ss:decompose_DS}) and sensitivity is analyzed in Section \ref{sec:sens_ana_LORD}\\
			\hline
			$\mu_{F^{DE}, G=1}$, $\mu_{CR,G=1}$ and $\mu_{\eta_{c},G=1}$ & Initialized as 0.5, 0.2 and 30 & Initial mean values of reproduction parameters (line \ref{s:Init}, Algorithm \ref{alg:framework}), later adapted per generation\\
			\hline
	\end{tabular}}
\end{table}

\subsection{Parameter Sensitivity Studies}
 \label{sec:sens_ana_LORD}
Two experiments are presented to study the sensitivity of the following parameters: (1) threshold ($\varepsilon_L$) for nearest neighbor graph formation (Section \ref{ss:decompose_DS}), and (2) the probability ($P_{mut}$) of switching between DE/rand/1/bin and SBX crossover (line \ref{s:line_9}, Algorithm \ref{alg:reproduction}).

1. \textbf{Threshold for Nearest Neighbor Graph}: During spectral clustering (Section \ref{ss:decompose_DS}) in LORD and LORD-II, the formation of the nearest neighbor graph ($\mathcal{G}$) considers edges between those pairs of solutions (nodes) whose distance is less than the threshold $\varepsilon_L$. This parameter $\varepsilon_L$ is set as $\alpha_L$ $(= 0.2)$ times the longest distance in the decision space, i.e., diagonal of the box-constrained decision space, $\mathcal{D}$. For validating this value of $\alpha_L$, it is varied between 0.1 to 0.8 (10\% to 80\% of the diagonal of $\mathcal{D}$) and the performance of LORD and LORD-II are noted in Table \ref{tab:sens_LORD} for some MMMOPs with $M=2$ or $M=3$.

\begin{table}[!ht]
	\centering
	\caption{Mean IGDX and IGDF over 51 independent runs for sensitivity study of $\alpha_L$ (parameter of LORD and LORD-II) on some 2- and 3-objective MMMOPs.}
	\label{tab:sens_LORD}
	\resizebox{0.8\linewidth}{!}{
		\begin{tabular}{|c|c|cccc|cccc|}
			\hline
			& & \multicolumn{4}{c|}{\textbf{IGDX}} & \multicolumn{4}{c|}{\textbf{IGDF}}\\
			\cline{3-10}
			& $\alpha_L\rightarrow$ & $0.1$ & $0.2$ & $0.5$ & $0.8$ & $0.1$ & $0.2$ & $0.5$ & $0.8$\\
			\hline
			\multirow{8}{*}{\rotatebox[origin=c]{90}{LORD}}& MMF1 & 0.0504 & \cellcolor{black!25}0.0431 & \cellcolor{black!10}0.0479 & 0.0492 & \cellcolor{black!10}0.0028 & \cellcolor{black!25}0.0025 & \cellcolor{black!25}0.0025 & \cellcolor{black!10}0.0028\\
			& MMF2 & 0.1431 & \cellcolor{black!25}0.0180 & \cellcolor{black!10}0.0304 & 0.0366 & \cellcolor{black!10}0.0092 & \cellcolor{black!25}0.0070 & 0.0109 & 0.0173\\
			& MMF3 & 0.0459 & \cellcolor{black!25}0.0176 & \cellcolor{black!10}0.0419 & 0.0458 & \cellcolor{black!10}0.0084 & \cellcolor{black!25}0.0069 & 0.0103 & 0.0117\\
			& MMF4 & \cellcolor{black!10}0.0298 & \cellcolor{black!25}0.0251 & 0.0303 & 0.0352 & \cellcolor{black!10}0.0021 & \cellcolor{black!25}0.0018 & 0.0023 & 0.0024\\
			& MMF5 & 0.0976 & \cellcolor{black!25}0.0814 & \cellcolor{black!10}0.0943 & 0.1165 & \cellcolor{black!10}0.0025 & \cellcolor{black!25}0.0024 & \cellcolor{black!10}0.0025 & 0.0027\\
			& MMF6 & 0.0812 & \cellcolor{black!25}0.0692 & \cellcolor{black!10}0.0720 & 0.0890 & 0.0025 & \cellcolor{black!25}0.0023 & \cellcolor{black!10}0.0024 & \cellcolor{black!10}0.0024\\
			& MMF7 & \cellcolor{black!10}0.0277 & \cellcolor{black!25}0.0218 & 0.0299 & 0.0339 & \cellcolor{black!10}0.0024 & \cellcolor{black!25}0.0022 & 0.0026 & 0.0028\\
			& MMF8 & 0.1631 & \cellcolor{black!25}0.0762 & \cellcolor{black!10}0.1299 & 0.1577 & \cellcolor{black!25}0.0025 & \cellcolor{black!25}0.0025 & \cellcolor{black!25}0.0025 & \cellcolor{black!25}0.0025\\
			\hline
			\multirow{4}{*}{\rotatebox[origin=c]{90}{LORD-II}}& MMF14 & \cellcolor{black!10}0.0495 & \cellcolor{black!25}0.0443 & 0.0522 & 0.0580 & 0.0550 & \cellcolor{black!25}0.0540 & \cellcolor{black!10}0.0545 & 0.0546 \\
			& MMF14\_a & \cellcolor{black!10}0.0657 & \cellcolor{black!25}0.0576 & 0.0665 & 0.0674 & \cellcolor{black!10}0.0574 & \cellcolor{black!25}0.0561 & 0.0582 & 0.0583 \\
			& MMF15 & 0.0295 & \cellcolor{black!25}0.0287 & \cellcolor{black!10}0.0292 & 0.0293 & \cellcolor{black!10}0.0552 & \cellcolor{black!25}0.0548 & \cellcolor{black!10}0.0552 & 0.0558\\
			& MMF15\_a & \cellcolor{black!10}0.0369 & \cellcolor{black!25}0.0355 & 0.0373 & 0.0379 & \cellcolor{black!10}0.0584 & \cellcolor{black!25}0.0571 & 0.0589 & 0.0593\\
			\hline
	\end{tabular}}
\end{table}

From Table \ref{tab:sens_LORD}, the best performance is observed when $\alpha_L = 0.2$. The performance deteriorates for higher $\alpha_L$ as all the candidates in $\mathcal{A}^{nd}$ form a single cluster ($k_{\mathcal{CC}}=1$) and distinguishability of the multiple subsets in PS is lost. The performance also deteriorates for lower $\alpha_L$ as $k_{\mathcal{CC}}\to\left\vert\mathcal{A}^{nd}\right\vert$ and the candidates become independent (higher randomness).

2. \textbf{Probability of Reproduction Switching}: During the probabilistic mutation switching (Algorithm \ref{alg:reproduction}) in LORD and LORD-II, $P_{mut}$ decides between DE/rand/1/bin \cite{DE_survey} and SBX-crossover \cite{GA}. However, in either case, polynomial mutation \cite{polymut_recentref} is also executed. This parameter $P_{mut}$ is set as $0.25$ after investigating the following cases:
\begin{enumerate}
	\item $P_{mut}=0.00$: only DE/rand/1/bin is used,
	\item $P_{mut}=0.25$: DE/rand/1/bin is used more often than SBX-crossover,
	\item $P_{mut}=0.50$: DE/rand/1/bin and SBX-crossover are equally-likely to be used,
	\item $P_{mut}=0.75$: SBX-crossover is used more often than DE/rand/1/bin, and
	\item $P_{mut}=1.00$: only SBX-crossover is used.
\end{enumerate}

The performance of LORD and LORD-II are noted in Table \ref{tab:sens_switch_LORD} for some MMMOPs. From Table \ref{tab:sens_switch_LORD}, the best performance is observed when $P_{mut} = 0.25$. Hence, for exploration of the search space, DE/rand/1/bin is preferred over SBX-crossover \cite{DEvsGA} along with a switching scheme to combine the benefits of both these strategies.
\begin{table}[!ht]
	\centering
	\caption{Mean IGDX and IGDF over 51 independent runs for sensitivity study of $P_{mut}$ (parameter of LORD and LORD-II) on some 2- and 3-objective MMMOPs.}
	\label{tab:sens_switch_LORD}
	\resizebox{\linewidth}{!}{
		\begin{tabular}{|c|c|ccccc|ccccc|}
			\hline
			& & \multicolumn{5}{c|}{\textbf{IGDX}} & \multicolumn{5}{c|}{\textbf{IGDF}}\\
			\cline{3-12}
			& $P_{mut}\rightarrow$ & $0.00$ & $0.25$ & $0.50$ & $0.75$ & $1.00$ & $0.00$ & $0.25$ & $0.50$ & $0.75$ & $1.00$\\
			\hline
			\multirow{8}{*}{\rotatebox[origin=c]{90}{LORD}}& MMF1 & 0.0529 & \cellcolor{black!25}0.0431 & \cellcolor{black!10}0.0470 & 0.0472 & 0.0506 & 0.0028 & \cellcolor{black!25}0.0026 & \cellcolor{black!10}0.0027 & \cellcolor{black!10}0.0027 & \cellcolor{black!10}0.0027\\
			& MMF2 & 0.0694 & \cellcolor{black!25}0.0110 & \cellcolor{black!10}0.0169 & 0.0207 & 0.0251 & 0.0100 & \cellcolor{black!25}0.0069 & \cellcolor{black!10}0.0085 & 0.0097 & 0.0141\\
			& MMF3 & 0.0603 & 0.0275 & \cellcolor{black!25}0.0116 &\cellcolor{black!10} 0.0188 & 0.0217 & 0.0169 & \cellcolor{black!25}0.0065 & \cellcolor{black!10}0.0070 & 0.0082 & 0.0475\\
			& MMF4 & 0.0283 & \cellcolor{black!25}0.0237 & \cellcolor{black!10}0.0239 & 0.0287 & 0.0381 & \cellcolor{black!10}0.0023 & \cellcolor{black!25}0.0021 & \cellcolor{black!25}0.0021 & \cellcolor{black!10}0.0023 & 0.0025\\
			& MMF5 & 0.0923 & \cellcolor{black!10}0.0789 & \cellcolor{black!25}0.0738 & 0.0900 & 0.0904 & 0.0027 & \cellcolor{black!10}0.0025 & \cellcolor{black!25}0.0024 & 0.0026 & 0.0028\\
			& MMF6 & 0.1199 & \cellcolor{black!25}0.0693 & \cellcolor{black!10}0.0777 & 0.0827 & 0.0976 & \cellcolor{black!10}0.0025 & \cellcolor{black!25}0.0024 & \cellcolor{black!10}0.0025 & \cellcolor{black!10}0.0025 & \cellcolor{black!10}0.0025\\
			& MMF7 & 0.0240 & \cellcolor{black!25}0.0209 & 0.0229 & \cellcolor{black!10}0.0228 & 0.0278 & \cellcolor{black!10}0.0024 & \cellcolor{black!25}0.0023 & \cellcolor{black!25}0.0023 & \cellcolor{black!25}0.0023 & 0.0025\\
			& MMF8 & 0.4619 & 0.1197 & \cellcolor{black!10}0.1085 & \cellcolor{black!25}0.0737 & 0.1123 & \cellcolor{black!10}0.0026 & \cellcolor{black!25}0.0025 & \cellcolor{black!10}0.0026 & \cellcolor{black!25}0.0025 & \cellcolor{black!10}0.0026\\
			\hline
			\multirow{4}{*}{\rotatebox[origin=c]{90}{LORD-II}}& MMF14    & \cellcolor{black!10}0.0490 & \cellcolor{black!25}0.0484 & 0.0497 & 0.0494 & 0.0502 & 0.0547 & \cellcolor{black!10}0.0545 & \cellcolor{black!25}0.0542 & 0.0548 & 0.0554\\
			& MMF14\_a & 0.0617 & \cellcolor{black!25}0.0609 & \cellcolor{black!10}0.0613 & 0.0650 & 0.0671 & \cellcolor{black!10}0.0578 & \cellcolor{black!25}0.0563 & 0.0580 & 0.0589 & 0.0596\\
			& MMF15    & 0.0296 & \cellcolor{black!25}0.0288 & \cellcolor{black!10}0.0291 & 0.0292 & 0.0291 & 0.0553 & \cellcolor{black!25}0.0551 & \cellcolor{black!10}0.0552 & 0.0553 & 0.0560\\
			& MMF15\_a & 0.0374 & \cellcolor{black!10}0.0370 & \cellcolor{black!25}0.0364 & 0.0372 & 0.0378 & \cellcolor{black!25}0.0588 & \cellcolor{black!25}0.0588 & \cellcolor{black!10}0.0593 & 0.0594 & 0.0606\\
			\hline
	\end{tabular}}
\end{table}

\subsection{Comparison of LORD and LORD-II with Other MMMOEAs}
Five sets of experiments are conducted to compare the performance of LORD and LORD-II with other MMMOEAs.

1) \textbf{Experiment-I: Comparison on CEC 2019 Test Suite}: For 2-objective MMMOPs, the performance of LORD in decision and objective spaces are presented in Tables \ref{tab:DS_2obj} and \ref{tab:OS_2obj}, respectively. For $M$-objective MMMOPs (with $M\geq$3), the performance of LORD-II is presented in Table \ref{tab:res_Mobj}. 

The plots of the estimated PSs and PFs (median run) are presented in Section 3 of the supplementary material. All the results are statistically validated using the Wilcoxon's rank-sum test \cite{ARMOEA} under the null hypothesis ($H_0$) that the performance of LORD (or LORD-II) is equivalent to other MMMOEAs. The statistical significance is indicated using three signs: $+$ denoting LORD (or LORD-II) is superior, $-$ denoting the competitor EA is superior, and $\sim$ indicating the algorithms are equivalent.

\begin{table*}[!t]
	\caption{Mean $\pm$ Standard Deviation (Significance) of rPSP and IGDX for $2$-objective MMMOPs over 51 Runs.}
	\label{tab:DS_2obj}
	\resizebox{\linewidth}{!}{
	\begin{tabular}{|c|c|c|c|c|c|c|c|c|}
		\hline
		& \multicolumn{4}{c|}{rPSP=IGDX/cov\_rate} & \multicolumn{4}{c|}{IGDX}\\
		\cline{2-9}
		Problems & LORD & MO\_Ring\_PSO\_SCD & DN-NSGA-II & NSGA-II & LORD & MO\_Ring\_PSO\_SCD & DN-NSGA-II & NSGA-II\\
		\hline
		MMF1 & \cellcolor{black!25}0.0441 $\pm$ & \cellcolor{black!10}0.0489 $\pm$ & 0.0957 $\pm$ & 0.0652 $\pm$ & \cellcolor{black!25}0.0431 $\pm$ & \cellcolor{black!10}0.0485 $\pm$ & 0.0939 $\pm$ & 0.0645 $\pm$ \\
		& \cellcolor{black!25}0.0044 & \cellcolor{black!10}0.0018 ($+$) & 0.0146 ($+$) & 0.0103 ($+$) & \cellcolor{black!25}0.0044 & \cellcolor{black!10}0.0017 ($+$) & 0.0141 ($+$) & 0.0098 ($+$)\\			
		\hline
		MMF1\_z & \cellcolor{black!10}0.0356 $\pm$ & \cellcolor{black!25}0.0354 $\pm$ & 0.0822 $\pm$ & 0.3892 $\pm$ & \cellcolor{black!10}0.0351 $\pm$ & \cellcolor{black!25}0.0352 $\pm$ & 0.0805 $\pm$ & 0.2606 $\pm$\\
		& \cellcolor{black!10}0.0069 & \cellcolor{black!25}0.0019 ($\sim$) & 0.0166 ($+$) & 0.3913 ($+$) & \cellcolor{black!10}0.0075 & \cellcolor{black!25}0.0018 ($\sim$) & 0.0157 ($+$) & 0.1608 ($+$)\\
		\hline
		MMF1\_e & \cellcolor{black!10}0.8894 $\pm$ & \cellcolor{black!25}0.5501 $\pm$ & 1.7201 $\pm$ & 14.0870 $\pm$ & \cellcolor{black!10}0.7499 $\pm$ & \cellcolor{black!25}0.4738 $\pm$ & 1.1536 $\pm$ & 3.0324 $\pm$\\
		& \cellcolor{black!10}0.1466 & \cellcolor{black!25}0.1276 ($-$) & 1.2086 ($+$) & 8.1289 ($+$) & \cellcolor{black!10}0.4192 & \cellcolor{black!25}0.0847 ($-$) & 0.5095 ($+$) & 0.7634 ($+$)\\
		\hline
		MMF2 & \cellcolor{black!25}0.0219 $\pm$ & \cellcolor{black!10}0.0444 $\pm$ & 0.1356 $\pm$ & 0.0766 $\pm$ & \cellcolor{black!25}0.0180 $\pm$ & \cellcolor{black!10}0.0416 $\pm$ & 0.1121 $\pm$ & 0.0650 $\pm$\\
		& \cellcolor{black!25}0.0108 & \cellcolor{black!10}0.0113 ($+$) & 0.0805 ($+$) & 0.0402 ($+$) & \cellcolor{black!25}0.0093 & \cellcolor{black!10}0.0103 ($+$) & 0.0525 ($+$) & 0.0300 ($+$)\\
		\hline
		MMF3 & \cellcolor{black!25}0.0200 $\pm$ & \cellcolor{black!10}0.0294 $\pm$ & 0.1249 $\pm$ & 0.0785 $\pm$ & \cellcolor{black!25}0.0176 $\pm$ & \cellcolor{black!10}0.0276 $\pm$ & 0.0968 $\pm$ & 0.0661 $\pm$\\
		& \cellcolor{black!25}0.0105 & \cellcolor{black!10}0.0074 ($+$) & 0.1291 ($+$) & 0.0416 ($+$) & \cellcolor{black!25}0.0080 & \cellcolor{black!10}0.0061 ($+$) & 0.0632 ($+$) & 0.0311 ($+$)\\
		\hline
		MMF4 & \cellcolor{black!25}0.0253 $\pm$ & \cellcolor{black!10}0.0274 $\pm$ & 0.0854 $\pm$ & 0.1066 $\pm$ & \cellcolor{black!25}0.0251 $\pm$ & \cellcolor{black!10}0.0271 $\pm$ & 0.0849 $\pm$ & 0.1004 $\pm$\\
		& \cellcolor{black!25}0.0036 & \cellcolor{black!10}0.0014 ($+$) & 0.0232 ($+$) & 0.0468 ($+$) & \cellcolor{black!25}0.0039 & \cellcolor{black!10}0.0014 ($+$) & 0.0230 ($+$) & 0.0411 ($+$)\\
		\hline
		MMF5 & \cellcolor{black!25}0.0814 $\pm$ & \cellcolor{black!10}0.0864 $\pm$ & 0.1788 $\pm$ & 0.1525 $\pm$ & \cellcolor{black!25}0.0814 $\pm$ & \cellcolor{black!10}0.0857 $\pm$ & 0.1763 $\pm$ & 0.1478 $\pm$\\
		& \cellcolor{black!25}0.0080 & \cellcolor{black!10}0.0045 ($+$) & 0.0179 ($+$) & 0.0296 ($+$) & \cellcolor{black!25}0.0074 & \cellcolor{black!10}0.0044 ($+$) & 0.0165 ($+$) & 0.0265 ($+$)\\
		\hline
		MMF6 & \cellcolor{black!25}0.0692 $\pm$ & \cellcolor{black!10}0.0741 $\pm$ & 0.1453 $\pm$ & 0.1410 $\pm$ & \cellcolor{black!25}0.0692 $\pm$ & \cellcolor{black!10}0.0736 $\pm$ & 0.1433 $\pm$ & 0.1372 $\pm$\\
		& \cellcolor{black!25}0.0104 & \cellcolor{black!10}0.0044 ($+$) & 0.0176 ($+$) & 0.0272 ($+$) & \cellcolor{black!25}0.0104 & \cellcolor{black!10}0.0042 ($+$) & 0.0173 ($+$) & 0.0251 ($+$)\\
		\hline
		MMF7 & \cellcolor{black!25}0.0219 $\pm$ & \cellcolor{black!10}0.0264 $\pm$ & 0.0535 $\pm$ & 0.0452 $\pm$ & \cellcolor{black!25}0.0218 $\pm$ & \cellcolor{black!10}0.0262 $\pm$ & 0.0524 $\pm$ & 0.0420 $\pm$\\
		& \cellcolor{black!25}0.0044 & \cellcolor{black!10}0.0014 ($+$) & 0.0098 ($+$) & 0.0132 ($+$) & \cellcolor{black!25}0.0025 & \cellcolor{black!10}0.0014 ($+$) & 0.0092 ($+$) & 0.0106 ($+$)\\
		\hline
		MMF8 & \cellcolor{black!10}0.0745 $\pm$ & \cellcolor{black!25}0.0679 $\pm$ & 0.2969 $\pm$ & 0.9348 $\pm$ & \cellcolor{black!10}0.0762 $\pm$ & \cellcolor{black!25}0.0673 $\pm$ & 0.2860 $\pm$ & 0.7198 $\pm$\\
		& \cellcolor{black!10}0.0452 & \cellcolor{black!25}0.0049 ($\sim$) & 0.1120 ($+$) & 0.4682 ($+$) & \cellcolor{black!10}0.0504 & \cellcolor{black!25}0.0048 ($\sim$) & 0.1078 ($+$) & 0.3034 ($+$)\\			
		\hline
		MMF9 & \cellcolor{black!25}0.0047 $\pm$ & \cellcolor{black!10}0.0079 $\pm$ & 0.0229 $\pm$ & 1.7445 $\pm$ & \cellcolor{black!25}0.0046 $\pm$ & \cellcolor{black!10}0.0079 $\pm$ & 0.0229 $\pm$ & 0.1783 $\pm$\\
		& \cellcolor{black!25}0.0002 & \cellcolor{black!10}0.0005 ($+$) & 0.0081 ($+$) & 1.9877 ($+$) & \cellcolor{black!25}0.0002 & \cellcolor{black!10}0.0005 ($+$) & 0.0081 ($+$) & 0.0740 ($+$)\\
		\hline
		MMF10 & \cellcolor{black!25}0.0018 $\pm$ & \cellcolor{black!10}0.0293 $\pm$ & 0.1426 $\pm$ & 0.0398 $\pm$ & \cellcolor{black!25}0.0018 $\pm$ & \cellcolor{black!10}0.0276 $\pm$ & 0.1295 $\pm$ & 0.0398 $\pm$\\
		& \cellcolor{black!25}0.0007 & \cellcolor{black!10}0.0113 ($+$) & 0.0834 ($+$) & 0.1184 ($\sim$) & \cellcolor{black!25}0.0009 & \cellcolor{black!10}0.0092 ($+$) & 0.0747 ($+$) & 0.1184 ($\sim$)\\
		\hline
		MMF11 & \cellcolor{black!10}0.0029 $\pm$ & 0.0055 $\pm$ & 0.0045 $\pm$ & \cellcolor{black!25}0.0027 $\pm$ & \cellcolor{black!10}0.0029 $\pm$ & 0.0054 $\pm$ & 0.0045 $\pm$ & \cellcolor{black!25}0.0027 $\pm$\\
		& \cellcolor{black!10}0.0002 & 0.0003 ($+$) & 0.0003 ($+$) & \cellcolor{black!25}0.0003 ($-$) & \cellcolor{black!10}0.0002 & 0.0003 ($+$) & 0.0003 ($+$) & \cellcolor{black!25}0.0003 ($-$)\\
		\hline
		MMF12 & \cellcolor{black!25}0.0013 $\pm$ & 0.0038 $\pm$ & 0.0090 $\pm$ & \cellcolor{black!10}0.0013 $\pm$ & \cellcolor{black!25}0.0013 $\pm$ & 0.0038 $\pm$ & 0.0090 $\pm$ & \cellcolor{black!10}0.0013 $\pm$\\
		& \cellcolor{black!25}0.0001 & 0.0003 ($+$) & 0.0159 ($+$) & \cellcolor{black!10}0.0002 ($\sim$) & \cellcolor{black!25}0.0001 & 0.0003 ($+$) & 0.0159 ($+$) & \cellcolor{black!10}0.0002 ($\sim$)\\
		\hline
		MMF13 & \cellcolor{black!25}0.0243 $\pm$ & \cellcolor{black!10}0.0317 $\pm$ & 0.0614 $\pm$ & 0.1492 $\pm$ & \cellcolor{black!25}0.0242 $\pm$ & \cellcolor{black!10}0.0314 $\pm$ & 0.0609 $\pm$ & 0.0880 $\pm$\\
		& \cellcolor{black!25}0.0039 & \cellcolor{black!10}0.0014 ($+$) & 0.0070 ($+$) & 0.0652 ($+$) & \cellcolor{black!25}0.0039 & \cellcolor{black!10}0.0013 ($+$) & 0.0064 ($+$) & 0.0173 ($+$)\\
		\hline
		Omni- & \cellcolor{black!25}0.0754 $\pm$ & \cellcolor{black!10}0.3946 $\pm$ & 1.4390 $\pm$ & 1.8176 $\pm$ & \cellcolor{black!25}0.0706 $\pm$ & \cellcolor{black!10}0.3907 $\pm$ & 1.4159 $\pm$ & 1.4210 $\pm$\\
		test & \cellcolor{black!25}0.0242 & \cellcolor{black!10}0.0939 ($+$) & 0.2069 ($+$) & 0.6886 ($+$) & \cellcolor{black!25}0.0215 & \cellcolor{black!10}0.0927 ($+$) & 0.1986 ($+$) & 0.3726 ($+$)\\
		\hline
		SYM-PART & \cellcolor{black!25}0.0556 $\pm$ & \cellcolor{black!10}0.1741 $\pm$ & 4.1590 $\pm$ & 113.0044 $\pm$ & \cellcolor{black!25}0.0549 $\pm$ & \cellcolor{black!10}0.1733 $\pm$ & 4.0657 $\pm$ & 6.8332 $\pm$\\
		simple & \cellcolor{black!25}0.0145 & \cellcolor{black!10}0.0301 ($+$) & 0.8683 ($+$) & 131.2343 ($+$) & \cellcolor{black!25}0.0130 & \cellcolor{black!10}0.0300 ($+$) & 0.7040 ($+$) & 1.8906 ($+$)\\ 
		\hline
		SYM-PART & \cellcolor{black!25}0.1730 $\pm$ & \cellcolor{black!10}0.3142 $\pm$ & 5.5941 $\pm$ & 13.9239 $\pm$ & \cellcolor{black!25}0.1558 $\pm$ & \cellcolor{black!10}0.2926 $\pm$ & 3.7659 $\pm$ & 5.4249 $\pm$\\
		rotated & \cellcolor{black!25}0.0743 & \cellcolor{black!10}0.3533 ($+$) & 3.6017 ($+$) & 12.8588 ($+$) & \cellcolor{black!25}0.0760 & \cellcolor{black!10}0.2938 ($+$) & 1.2478 ($+$) & 1.9790 ($+$)\\
		\hline
		Sum-up & $+/-/\sim$ & 15/1/2 & 18/0/0 & 15/1/2 & $+/-/\sim$ & 15/1/2 & 18/0/0 & 15/1/2\\
		\hline
	\end{tabular}}
\end{table*}
\begin{table*}[!t]
	\caption{Mean $\pm$ Standard Deviation (Significance) of rHV and IGDF for $2$-objective MMMOPs over 51 Runs.}
	\label{tab:OS_2obj}
	\resizebox{\linewidth}{!}{
		\begin{tabular}{|c|c|c|c|c|c|c|c|c|}
			\hline
			& \multicolumn{4}{c|}{rHV=1/HV} & \multicolumn{4}{c|}{IGDF}\\
			\cline{2-9}
			Problems & LORD & MO\_Ring\_PSO\_SCD & DN-NSGA-II & NSGA-II & LORD & MO\_Ring\_PSO\_SCD & DN-NSGA-II & NSGA-II\\
			\hline
			MMF1 & \cellcolor{black!25}1.0737 $\pm$ & 1.1484 $\pm$ & 1.1495 $\pm$ & \cellcolor{black!10}1.0738 $\pm$ & \cellcolor{black!25}0.0025 $\pm$ & 0.0037 $\pm$ & 0.0043 $\pm$ & \cellcolor{black!10}0.0028 $\pm$\\
			& \cellcolor{black!25}0.0008 & 0.0005 ($+$) & 0.0014 ($+$) & \cellcolor{black!10}0.0006 ($\sim$) & \cellcolor{black!25}0.0002 & 0.0002 ($+$) & 0.0005 ($+$) & \cellcolor{black!10}0.0004 ($+$)\\
			\hline
			MMF1\_z & \cellcolor{black!25}1.0731 $\pm$ & 1.1483 $\pm$ & 1.1484 $\pm$ & \cellcolor{black!10}1.1255 $\pm$ & \cellcolor{black!25}0.0022 $\pm$ & \cellcolor{black!10}0.0036 $\pm$ & 0.0036 $\pm$ & 0.0396 $\pm$\\
			& \cellcolor{black!25}0.0008 & 0.0005 ($+$) & 0.0009 ($+$) & \cellcolor{black!10}0.0615 ($+$) & \cellcolor{black!25}0.0001 & \cellcolor{black!10}0.0002 ($+$) & 0.0004 ($+$) & 0.0496 ($+$)\\
			\hline
			MMF1\_e & \cellcolor{black!25}1.0751 $\pm$ & 1.1861 $\pm$ & 1.2080 $\pm$ & \cellcolor{black!10}1.1058 $\pm$ & \cellcolor{black!25}0.0029 $\pm$ & \cellcolor{black!10}0.0119 $\pm$ & 0.0276 $\pm$ & 0.0250 $\pm$\\
			& \cellcolor{black!25}0.0021 & 0.0173 ($+$) & 0.0387 ($+$) & \cellcolor{black!10}0.0180 ($+$) & \cellcolor{black!25}0.0006 & \cellcolor{black!10}0.0017 ($+$) & 0.0207 ($+$) & 0.0139 ($+$)\\
			\hline
			MMF2 & \cellcolor{black!25}1.0817 $\pm$ & 1.1848 $\pm$ & 1.1944 $\pm$ & \cellcolor{black!10}1.1168 $\pm$ & \cellcolor{black!25}0.0070 $\pm$ &\cellcolor{black!10} 0.0207 $\pm$ & 0.0325 $\pm$ & 0.0300 $\pm$\\
			& \cellcolor{black!25}0.0120 & 0.0059 ($+$) & 0.0322 ($+$) & \cellcolor{black!10}0.0280 ($+$) & \cellcolor{black!25}0.0031 & \cellcolor{black!10}0.0034 ($+$) & 0.0238 ($+$) & 0.0182 ($+$)\\
			\hline
			MMF3 & \cellcolor{black!25}1.0792 $\pm$ & 1.1739 $\pm$ & 1.1873 $\pm$ & \cellcolor{black!10}1.1089 $\pm$ & \cellcolor{black!25}0.0069 $\pm$ & \cellcolor{black!10}0.0154 $\pm$ & 0.0263 $\pm$ & 0.0229 $\pm$\\
			& \cellcolor{black!25}0.0322 & 0.0043 ($+$) & 0.0398 ($+$) & \cellcolor{black!10}0.0212 ($+$) & \cellcolor{black!25}0.0023 & \cellcolor{black!10}0.0025 ($+$) & 0.0308 ($+$) & 0.0126 ($+$)\\
			\hline
			MMF4 & \cellcolor{black!25}1.5234 $\pm$ & 1.8620 $\pm$ & 1.8577 $\pm$ & \cellcolor{black!10}1.5241 $\pm$ & \cellcolor{black!25}0.0018 $\pm$ & 0.0037 $\pm$ & 0.0032 $\pm$ & \cellcolor{black!10}0.0024 $\pm$\\
			& \cellcolor{black!25}0.0003 & 0.0021 ($+$) & 0.0012 ($+$) & \cellcolor{black!10}0.0004 ($+$) & \cellcolor{black!25}0.0002 & 0.0004 ($+$) & 0.0002 ($+$) & \cellcolor{black!10}0.0002 ($+$)\\
			\hline
			MMF5 & \cellcolor{black!25}1.0734 $\pm$ & 1.1485 $\pm$ & 1.1488 $\pm$ & \cellcolor{black!10}1.0739 $\pm$ & \cellcolor{black!25}0.0024 $\pm$ & 0.0037 $\pm$ & 0.0039 $\pm$ & \cellcolor{black!10}0.0028 $\pm$\\
			& \cellcolor{black!25}0.0006 & 0.0006 ($+$) & 0.0015 ($+$) & \cellcolor{black!10}0.0003 ($+$) & \cellcolor{black!25}0.0001 & 0.0001 ($+$) & 0.0007 ($+$) & \cellcolor{black!10}0.0002 ($+$)\\
			\hline
			MMF6 & \cellcolor{black!25}1.0732 $\pm$ & 1.1483 $\pm$ & 1.1486 $\pm$ & \cellcolor{black!10}1.0738 $\pm$ & \cellcolor{black!25}0.0023 $\pm$ & 0.0035 $\pm$ & 0.0036 $\pm$ & \cellcolor{black!10}0.0026 $\pm$\\
			& \cellcolor{black!25}0.0003 & 0.0009 ($+$) & 0.0016 ($+$) & \cellcolor{black!10}0.0006 ($+$) & \cellcolor{black!25}0.0001 & 0.0002 ($+$) & 0.0003 ($+$) & \cellcolor{black!10}0.0002 ($+$)\\
			\hline
			MMF7 & \cellcolor{black!25}1.0731 $\pm$ & 1.1484 $\pm$ & 1.1498 $\pm$ & \cellcolor{black!10}1.0736 $\pm$ & \cellcolor{black!25}0.0022 $\pm$ & 0.0037 $\pm$ & 0.0039 $\pm$ & \cellcolor{black!10}0.0027 $\pm$\\
			& \cellcolor{black!25}0.0002 & 0.0009 ($+$) & 0.0011 ($+$) & \cellcolor{black!10}0.0003 ($+$) & \cellcolor{black!25}0.0001 & 0.0003 ($+$) & 0.0003 ($+$) & \cellcolor{black!10}0.0003 ($+$)\\
			\hline
			MMF8 & \cellcolor{black!25}1.7915 $\pm$ & 2.4065 $\pm$ & 2.3813 $\pm$ & \cellcolor{black!10}1.7920 $\pm$ & \cellcolor{black!25}0.0025 $\pm$ & 0.0048 $\pm$ & \cellcolor{black!10}0.0040 $\pm$ & \cellcolor{black!25}0.0025 $\pm$\\
			& \cellcolor{black!25}0.0012 & 0.0164 ($+$) & 0.0025 ($+$) & \cellcolor{black!10}0.0014 ($\sim$) & \cellcolor{black!25}0.0001 & 0.0002 ($+$) & \cellcolor{black!10}0.0004 ($+$) & \cellcolor{black!25}0.0001 ($\sim$)\\ 
			\hline
			MMF9 & \cellcolor{black!25}0.0820 $\pm$ & \cellcolor{black!10}0.1034 $\pm$ & \cellcolor{black!10}0.1034 $\pm$ & \cellcolor{black!25}0.0820 $\pm$ & \cellcolor{black!25}0.0085 $\pm$ & 0.0160 $\pm$ & 0.0141 $\pm$ & \cellcolor{black!10}0.0108 $\pm$\\
			& \cellcolor{black!25}0.0000 & \cellcolor{black!10}0.0000 ($+$) & \cellcolor{black!10}0.0000 ($+$) & \cellcolor{black!25}0.0000 ($\sim$) & \cellcolor{black!25}0.0007 & 0.0014 ($+$) & 0.0012 ($+$) & \cellcolor{black!10}0.0007 ($+$)\\
			\hline
			MMF10 & \cellcolor{black!25}0.0678 $\pm$ & 0.0679 $\pm$ & 0.0680 $\pm$ & \cellcolor{black!10}0.0678 $\pm$ & \cellcolor{black!25}0.0061 $\pm$ & 0.1128 $\pm$ & 0.1446 $\pm$ & \cellcolor{black!10}0.0074 $\pm$\\
			& \cellcolor{black!25}0.0000 & 0.0000 ($+$) & 0.0001 ($+$) & \cellcolor{black!10}0.0001 ($\sim$) & \cellcolor{black!25}0.0009 & 0.0230 ($+$) & 0.0660 ($+$) & \cellcolor{black!10}0.0000 ($+$)\\
			\hline
			MMF11 & \cellcolor{black!25}0.0581 $\pm$ & \cellcolor{black!25}0.0581 $\pm$ & \cellcolor{black!25}0.0581 $\pm$ & \cellcolor{black!25}0.0581 $\pm$ & \cellcolor{black!25}0.0082 $\pm$ & 0.0176 $\pm$ & 0.0136 $\pm$ & \cellcolor{black!10}0.0107 $\pm$\\
			& \cellcolor{black!25}0.0000 & \cellcolor{black!25}0.0000 ($\sim$) & \cellcolor{black!25}0.0000 ($\sim$) & \cellcolor{black!25}0.0000 ($\sim$) & \cellcolor{black!25}0.0004 & 0.0018 ($+$) & 0.0014 ($+$) & \cellcolor{black!10}0.0008 ($+$)\\
			\hline
			MMF12 & \cellcolor{black!10}0.5431 $\pm$ & 0.5452 $\pm$ & 0.5598 $\pm$ & \cellcolor{black!25}0.5430 $\pm$ & \cellcolor{black!25}0.0020 $\pm$ & \cellcolor{black!10}0.0068 $\pm$ & 0.0110 $\pm$ & \cellcolor{black!25}0.0020 $\pm$\\
			& \cellcolor{black!10}0.0000 & 0.0014 ($+$) & 0.0492 ($\sim$) & \cellcolor{black!25}0.0000 ($-$) & \cellcolor{black!25}0.0001 &  \cellcolor{black!10}0.0006 ($+$) & 0.0187 ($+$) & \cellcolor{black!25}0.0001 ($\sim$)\\
			\hline
			MMF13 & \cellcolor{black!25}0.0444 $\pm$ & \cellcolor{black!25}0.0444 $\pm$ & \cellcolor{black!25}0.0444 $\pm$ & \cellcolor{black!25}0.0444 $\pm$ & \cellcolor{black!25}0.0063 $\pm$ & 0.0264 $\pm$ & 0.0121 $\pm$ & \cellcolor{black!10}0.0089 $\pm$\\
			& \cellcolor{black!25}0.0000 & \cellcolor{black!25}0.0000 ($\sim$) & \cellcolor{black!25}0.0000 ($\sim$) & \cellcolor{black!25}0.0000 ($\sim$) & \cellcolor{black!25}0.0014 & 0.0076 ($+$) & 0.0036 ($+$) & \cellcolor{black!10}0.0014 ($+$)\\
			\hline
			Omni- & 0.0518 $\pm$ & \cellcolor{black!10}0.0190 $\pm$ & \cellcolor{black!25}0.0189 $\pm$ & 0.0518 $\pm$ & \cellcolor{black!10}0.0091 $\pm$ & 0.0422 $\pm$ & \cellcolor{black!25}0.0080 $\pm$ & 0.0100 $\pm$\\
			test & 0.0000 & \cellcolor{black!10}0.0000 ($-$) & \cellcolor{black!25}0.0000 ($-$) & 0.0000 ($\sim$) & \cellcolor{black!10}0.0015 & 0.0034 ($+$) & \cellcolor{black!25}0.0005 ($-$) & 0.0021 ($\sim$)\\
			\hline
			SYM-PART & \cellcolor{black!25}0.0520 $\pm$ & 0.0605 $\pm$ & \cellcolor{black!10}0.0601 $\pm$ & \cellcolor{black!25}0.0520 $\pm$ & 0.0165 $\pm$ & 0.0419 $\pm$ & \cellcolor{black!10}0.0127 $\pm$ & \cellcolor{black!25}0.0109 $\pm$\\
			simple & \cellcolor{black!25}0.0000 & 0.0001 ($+$) & \cellcolor{black!10}0.0000 ($+$) & \cellcolor{black!25}0.0000 ($\sim$) & 0.0039 & 0.0044 ($+$) & \cellcolor{black!10}0.0014 ($-$) & \cellcolor{black!25}0.0013 ($-$)\\
			\hline
			SYM-PART & \cellcolor{black!25}0.0520 $\pm$ & 0.0606 $\pm$ & \cellcolor{black!10}0.0601 $\pm$ & \cellcolor{black!25}0.0520 $\pm$ & 0.0178 $\pm$ & 0.0467 $\pm$ & \cellcolor{black!25}0.0152 $\pm$ & \cellcolor{black!10}0.0159 $\pm$\\
			rotated & \cellcolor{black!25}0.0000 & 0.0001 ($+$) & \cellcolor{black!10}0.0000 ($+$) & \cellcolor{black!25}0.0000 ($\sim$) & 0.0047 & 0.0058 ($+$) & \cellcolor{black!25}0.0022 ($-$) & \cellcolor{black!10}0.0040 ($\sim$)\\
			\hline
			Sum-up & $+/-/\sim$ & 15/1/2 & 14/1/3 & 8/1/9 & $+/-/\sim$ & 18/0/0 & 15/3/0 & 13/1/4\\
			\hline
	\end{tabular}}
\end{table*}
\begin{table*}[!t]
	\caption{Mean $\pm$ Standard Deviation (Significance) of NSX, CM\_NSX, D\_metric and CM for $M$-objective ($M\geq3$) MMMOPs over 51 Runs.}
	\label{tab:res_Mobj}
	\resizebox{\linewidth}{!}{
		\begin{tabular}{|c|c|c|c|c|c|c|c|c|c|c|c|c|}
			\hline
			& \multicolumn{3}{c|}{NSX} & \multicolumn{3}{c|}{CM\_NSX} & \multicolumn{3}{c|}{D\_metric} & \multicolumn{3}{c|}{CM}\\
			\cline{2-13}
			Problems ($M$) & LORD-II & MO\_Ring\_ & MOEA/DD & LORD-II & MO\_Ring\_ & MOEA/DD & LORD-II & MO\_Ring\_ & MOEA/DD & LORD-II & MO\_Ring\_ & MOEA/DD\\
			& & PSO\_SCD & & & PSO\_SCD & & & PSO\_SCD & & & PSO\_SCD & \\
			\hline
			MMF14 (3) & \cellcolor{black!25}0.0068 $\pm$ & 0.2400 $\pm$ & \cellcolor{black!10}0.0133 $\pm$ & \cellcolor{black!25}0.0203 $\pm$ & 0.1078 $\pm$ & \cellcolor{black!10}0.0228 $\pm$ & \cellcolor{black!25}0.0000 $\pm$ & \cellcolor{black!10}21.3266 $\pm$ & \cellcolor{black!25}0.0000 $\pm$ & \cellcolor{black!25}0.0419 $\pm$ & \cellcolor{black!10}0.1083 $\pm$ & \cellcolor{black!25}0.0419 $\pm$\\
			& \cellcolor{black!25}0.0052 & 0.0262($+$) & \cellcolor{black!10}0.0024 ($+$) & \cellcolor{black!25}0.0018 & 0.0088 ($+$) & \cellcolor{black!10}0.0019 ($+$) & \cellcolor{black!25}0.0000 & \cellcolor{black!10}2.0086 ($+$) & \cellcolor{black!25}0.0000 ($\sim$) & \cellcolor{black!25}0.0003 & \cellcolor{black!10}0.0130 ($+$) & \cellcolor{black!25}0.0002 ($\sim$)\\
			\hline
			MMF14\_a (3) & \cellcolor{black!25}0.0251 $\pm$ & 0.2533 $\pm$ & \cellcolor{black!10}0.1333 $\pm$ & \cellcolor{black!25}0.1498 $\pm$ & 0.1534 $\pm$ & \cellcolor{black!10}0.2481 $\pm$ & \cellcolor{black!25}0.0000 $\pm$ & \cellcolor{black!10}22.2752 $\pm$ & \cellcolor{black!25}0.0000 $\pm$ & \cellcolor{black!25}0.0435 $\pm$ & 0.0949 $\pm$ & \cellcolor{black!10}0.0438 $\pm$\\
			& \cellcolor{black!25}0.0290 & 0.0320 ($+$) & \cellcolor{black!10}0.0259 ($+$) & \cellcolor{black!25}0.0518 & 0.0147 ($+$) & \cellcolor{black!10}0.0103 ($+$) & \cellcolor{black!25}0.0000 & \cellcolor{black!10}2.2816 ($+$) & \cellcolor{black!25}0.0000 ($\sim$) & \cellcolor{black!25}0.0007 & 0.0149 ($+$) & \cellcolor{black!10}0.0010 ($\sim$)\\
			\hline
			MMF15 (3) & \cellcolor{black!25}0.0205 $\pm$ & 0.4033 $\pm$ & \cellcolor{black!10}0.0400 $\pm$ & \cellcolor{black!25}0.0209 $\pm$ & 0.2356 $\pm$ & \cellcolor{black!10}0.0265 $\pm$ & \cellcolor{black!25}0.0000 $\pm$ & \cellcolor{black!10}24.4073$\pm$ & \cellcolor{black!25}0.0000 $\pm$ & \cellcolor{black!25}0.0422 $\pm$ & 0.1471 $\pm$ & \cellcolor{black!10}0.0426 $\pm$\\
			& \cellcolor{black!25}0.0073 & 0.0361 ($+$) & \cellcolor{black!10}0.0024 ($+$) & \cellcolor{black!25}0.0010 & 0.0251 ($+$) & \cellcolor{black!10}0.0039 ($\sim$) & \cellcolor{black!25}0.0000 & \cellcolor{black!10}3.8341 ($+$) & \cellcolor{black!25}0.0000 ($\sim$) & \cellcolor{black!25}0.0004 & 0.0161 ($+$) & \cellcolor{black!10}0.0004 ($\sim$)\\
			\hline
			MMF15\_a (3) & \cellcolor{black!25}0.0179 $\pm$ & 0.3400 $\pm$ & \cellcolor{black!10}0.0567 $\pm$ & \cellcolor{black!25}0.0270 $\pm$ & 0.2069 $\pm$ & \cellcolor{black!10}0.0454 $\pm$ & \cellcolor{black!25}0.0000 $\pm$ & \cellcolor{black!10}22.5315$\pm$ & \cellcolor{black!25}0.0000 $\pm$ & \cellcolor{black!25}0.0445 $\pm$ & 0.1322 $\pm$ & \cellcolor{black!10}0.0449 $\pm$\\
			& \cellcolor{black!25}0.0076 & 0.0262 ($+$) & \cellcolor{black!10}0.0024 ($+$) & \cellcolor{black!25}0.0180 & 0.0263 ($+$) & \cellcolor{black!10}0.0098 ($+$) & \cellcolor{black!25}0.0000 & \cellcolor{black!10}3.5845($+$) & \cellcolor{black!25}0.0000 ($\sim$) & \cellcolor{black!25}0.0007 & 0.0207 ($+$) & \cellcolor{black!10}0.0008 ($\sim$)\\
			\hline
			
			MMF14 (5) & \cellcolor{black!25}0.4838 $\pm$ & \cellcolor{black!10}0.5140 $\pm$ & 0.5152 $\pm$ & \cellcolor{black!25}0.1830 $\pm$ & 0.2363 $\pm$ & \cellcolor{black!10}0.1874 $\pm$ & \cellcolor{black!25}0.0000 $\pm$ & \cellcolor{black!10}43.9023 $\pm$ & \cellcolor{black!25}0.0000 $\pm$ & \cellcolor{black!10}0.0590 $\pm$ & 0.4121 $\pm$ & \cellcolor{black!25}0.0587 $\pm$\\
			& \cellcolor{black!25}0.0125 & \cellcolor{black!10}0.0168 ($+$) & 0.0057 ($+$) & \cellcolor{black!25}0.0015 & 0.0047 ($+$) & \cellcolor{black!10}0.0015 ($+$) & \cellcolor{black!25}0.0000 & \cellcolor{black!10}4.6565 ($+$) & \cellcolor{black!25}0.0000 ($\sim$) & \cellcolor{black!10}0.0020 & 0.0177 ($+$) & \cellcolor{black!25}0.0015 ($\sim$)\\
			\hline
			MMF14\_a (5) & \cellcolor{black!25}0.4855 $\pm$ & \cellcolor{black!10}0.4960 $\pm$ & 0.5232 $\pm$ & \cellcolor{black!25}0.2080 $\pm$ & 0.2809 $\pm$ & \cellcolor{black!10}0.2474 $\pm$ & \cellcolor{black!25}0.0000 $\pm$ & 46.7494 $\pm$ & \cellcolor{black!10}0.9428 $\pm$ & \cellcolor{black!25}0.0781 $\pm$ & 0.3659 $\pm$ & \cellcolor{black!10}0.0827 $\pm$\\
			& \cellcolor{black!25}0.0141 & \cellcolor{black!10}0.0175 ($\sim$) & 0.0029 ($+$) & \cellcolor{black!25}0.0155 & 0.0041 ($+$) & \cellcolor{black!10}0.0044 ($+$) & \cellcolor{black!25}0.0000 & 2.8893 ($+$) & \cellcolor{black!10}0.8165 ($+$) & \cellcolor{black!25}0.0022 & 0.0115 ($+$) & \cellcolor{black!10}0.0021 ($+$)\\
			\hline
			MMF15 (5) & \cellcolor{black!25}0.4959 $\pm$ & 0.5600 $\pm$ & \cellcolor{black!10}0.5172 $\pm$ & \cellcolor{black!25}0.1559 $\pm$ & 0.4118 $\pm$ & \cellcolor{black!10}0.1602 $\pm$ & \cellcolor{black!25}0.0000 $\pm$ & \cellcolor{black!10}43.6883 $\pm$ & \cellcolor{black!25}0.0000 $\pm$ & \cellcolor{black!10}0.0654 $\pm$ & 0.4610 $\pm$ & \cellcolor{black!25}0.0625 $\pm$\\
			& \cellcolor{black!25}0.0076 & 0.0155 ($+$) & \cellcolor{black!10}0.0014 ($+$) & \cellcolor{black!25}0.0015 & 0.0152 ($+$) & \cellcolor{black!10}0.0005 ($\sim$) & \cellcolor{black!25}0.0000 & \cellcolor{black!10}3.9734 ($+$) & \cellcolor{black!25}0.0000 ($\sim$) & \cellcolor{black!10}0.0017 & 0.0152 ($+$) & \cellcolor{black!25}0.0025 ($-$)\\
			\hline
			MMF15\_a (5) & \cellcolor{black!25}0.4969 $\pm$ & 0.5660 $\pm$ & \cellcolor{black!10}0.5232 $\pm$ & \cellcolor{black!25}0.1783 $\pm$ & 0.3672 $\pm$ & \cellcolor{black!10}0.1938 $\pm$ & \cellcolor{black!25}0.0000 $\pm$ & 45.2327 $\pm$ & \cellcolor{black!10}0.9428 $\pm$ & \cellcolor{black!25}0.0954 $\pm$ & 0.4339 $\pm$ & \cellcolor{black!10}0.0961 $\pm$\\
			& \cellcolor{black!25}0.0086 & 0.0155 ($+$) & \cellcolor{black!10}0.0071 ($+$) & \cellcolor{black!25}0.0062 & 0.0128 ($+$) & \cellcolor{black!10}0.0078 ($+$) & \cellcolor{black!25}0.0000 & 3.4695 ($+$) & \cellcolor{black!10}0.8165 ($+$) & \cellcolor{black!25}0.0045 & 0.0162 ($+$) & \cellcolor{black!10}0.0053 ($\sim$)\\
			\hline
			
			MMF14 (8) & \cellcolor{black!25}0.2772 $\pm$ & 0.5663 $\pm$ & \cellcolor{black!10}0.3088 $\pm$ & \cellcolor{black!25}0.4025 $\pm$ & 0.4386 $\pm$ & \cellcolor{black!10}0.4178 $\pm$ & \cellcolor{black!25}0.0000 $\pm$ & 101.1673 $\pm$ & \cellcolor{black!10}5.3833 $\pm$ & \cellcolor{black!25}0.1332 $\pm$ & 0.6277 $\pm$ & \cellcolor{black!10}0.1456 $\pm$\\
			& \cellcolor{black!25}0.0012 & 0.0129 ($+$) & \cellcolor{black!10}0.0018 ($+$) & \cellcolor{black!25}0.0011 & 0.0070 ($+$) & \cellcolor{black!10}0.0011 ($+$) & \cellcolor{black!25}0.0000 & 4.3256 ($+$) & \cellcolor{black!10}0.0000 ($+$) & \cellcolor{black!25}0.0002 & 0.0154 ($+$) & \cellcolor{black!10}0.0016 ($+$)\\
			\hline
			MMF14\_a (8) & \cellcolor{black!25}0.2796 $\pm$ & 0.5588 $\pm$ & \cellcolor{black!10}0.2900 $\pm$ & \cellcolor{black!25}0.4222 $\pm$ & 0.4480 $\pm$ & \cellcolor{black!10}0.4335 $\pm$ & \cellcolor{black!25}0.0000 $\pm$ & 106.8644 $\pm$ & \cellcolor{black!10}5.3833 $\pm$ & \cellcolor{black!25}0.1742 $\pm$ & 0.5917 $\pm$ & \cellcolor{black!10}0.1817 $\pm$\\
			& \cellcolor{black!25}0.0008 & 0.0114 ($+$) & \cellcolor{black!10}0.0062 ($+$) & \cellcolor{black!25}0.0004 & 0.0045 ($+$) & \cellcolor{black!10}0.0021 ($+$) & \cellcolor{black!25}0.0000 & 3.5223 ($+$) & \cellcolor{black!10}0.0000 ($+$) & \cellcolor{black!25}0.0044 & 0.0149 ($+$) & \cellcolor{black!10}0.0044 ($+$)\\
			\hline
			MMF15 (8) & \cellcolor{black!25}0.2524 $\pm$ & 0.6438 $\pm$ & \cellcolor{black!10}0.2713 $\pm$ & \cellcolor{black!25}0.3537 $\pm$ & 0.5312 $\pm$ & \cellcolor{black!10}0.3815 $\pm$ & \cellcolor{black!25}0.0000 $\pm$ & 99.5550 $\pm$ & \cellcolor{black!10}5.4810 $\pm$ & \cellcolor{black!25}0.1288 $\pm$ & 0.6767 $\pm$ & \cellcolor{black!10}0.1508 $\pm$\\
			& \cellcolor{black!25}0.0002 & 0.0175 ($+$) & \cellcolor{black!10}0.0018 ($+$) & \cellcolor{black!25}0.0052 & 0.0059 ($+$) & \cellcolor{black!10}0.0015 ($+$) & \cellcolor{black!25}0.0000 & 3.0343 ($+$) & \cellcolor{black!10}0.1382 ($+$) & \cellcolor{black!25}0.0038 & 0.0128 ($+$) & \cellcolor{black!10}0.0008 ($+$)\\
			\hline
			MMF15\_a (8) & \cellcolor{black!25}0.2430 $\pm$ & 0.6113 $\pm$ & \cellcolor{black!10}0.2688 $\pm$ & \cellcolor{black!25}0.3797 $\pm$ & 0.4885 $\pm$ & \cellcolor{black!10}0.3886 $\pm$ & \cellcolor{black!25}0.0000 $\pm$ & 103.5948 $\pm$ & \cellcolor{black!10}5.5787 $\pm$ & \cellcolor{black!25}0.2014 $\pm$ & 0.6488 $\pm$ & \cellcolor{black!10}0.2122 $\pm$\\
			& \cellcolor{black!25}0.0137 & 0.0093 ($+$) & \cellcolor{black!10}0.0027 ($+$) & \cellcolor{black!25}0.0115 & 0.0078 ($+$) & \cellcolor{black!10}0.0068 ($\sim$) & \cellcolor{black!25}0.0000 & 3.5180 ($+$) & \cellcolor{black!10}0.0000 ($+$) & \cellcolor{black!25}0.0025 & 0.0151 ($+$) & \cellcolor{black!10}0.0284 ($+$)\\
			\hline
			
			MMF14 (10) & \cellcolor{black!25}0.2595 $\pm$ & 0.5880 $\pm$ & \cellcolor{black!10}0.2620 $\pm$ & \cellcolor{black!25}0.5088 $\pm$ & 0.5584 $\pm$ & \cellcolor{black!10}0.5356 $\pm$ & \cellcolor{black!25}14.1331 $\pm$ & 132.4681 $\pm$ & \cellcolor{black!10}38.9838 $\pm$ & \cellcolor{black!25}0.2200 $\pm$ & 0.6575 $\pm$ & \cellcolor{black!10}0.2504 $\pm$\\
			& \cellcolor{black!25}0.0005 & 0.0101 ($+$) & \cellcolor{black!10}0.0038 ($+$) & \cellcolor{black!25}0.0087 & 0.0046 ($+$) & \cellcolor{black!10}0.0085 ($+$) & \cellcolor{black!25}2.5516 & 4.6720 ($+$) & \cellcolor{black!10}0.7255 ($+$) & \cellcolor{black!25}0.0037 & 0.0129 ($+$) & \cellcolor{black!10}0.0002 ($+$)\\
			\hline
			MMF14\_a (10) & \cellcolor{black!25}0.2717 $\pm$ & 0.5930 $\pm$ & \cellcolor{black!10}0.2718 $\pm$ & \cellcolor{black!25}0.5165 $\pm$ & 0.5699 $\pm$ & \cellcolor{black!10}0.5393 $\pm$ & \cellcolor{black!25}21.9290 $\pm$ & 137.8081 $\pm$ & \cellcolor{black!10} 41.7357 $\pm$ & \cellcolor{black!25}0.2554 $\pm$ & 0.6311 $\pm$ & \cellcolor{black!10}0.2966 $\pm$\\
			& \cellcolor{black!25}0.0106 & 0.0134 ($+$) & \cellcolor{black!10}0.0083 ($\sim$) & \cellcolor{black!25}0.0008 & 0.0036 ($+$) & \cellcolor{black!10}0.0063 ($+$) & \cellcolor{black!25}0.4837 & 3.8224 ($+$) & \cellcolor{black!10}0.5083 ($+$) & \cellcolor{black!25}0.0065 & 0.0128 ($+$) & \cellcolor{black!10}0.0041 ($+$)\\
			\hline
			MMF15 (10) & \cellcolor{black!10}0.2557 $\pm$ & 0.6590 $\pm$ & \cellcolor{black!25}0.2481 $\pm$ & \cellcolor{black!25}0.4516 $\pm$ & 0.6180 $\pm$ & \cellcolor{black!10}0.4850 $\pm$ & \cellcolor{black!25}17.6340 $\pm$ & 131.8568 $\pm$ & \cellcolor{black!10}36.4571 $\pm$ & \cellcolor{black!25}0.2219 $\pm$ & 0.7072 $\pm$ & \cellcolor{black!10}0.2535 $\pm$\\
			& \cellcolor{black!10}0.0023 & 0.0123 ($+$) & \cellcolor{black!25}0.0030 ($-$) & \cellcolor{black!25}0.0096 & 0.0071 ($+$) & \cellcolor{black!10}0.0005 ($+$) & \cellcolor{black!25}1.4436 & 4.0914 ($+$) & \cellcolor{black!10}9.3681 ($+$) & \cellcolor{black!25}0.0118 & 0.0109 ($+$) & \cellcolor{black!10}0.0072 ($+$)\\
			\hline
			MMF15\_a (10) & \cellcolor{black!10}0.2664 $\pm$ & 0.6350 $\pm$ & \cellcolor{black!25}0.2652 $\pm$ & \cellcolor{black!25}0.4731 $\pm$ & 0.5780 $\pm$ & \cellcolor{black!10}0.4903 $\pm$ & \cellcolor{black!25}23.2171 $\pm$ & 133.7104 $\pm$ & \cellcolor{black!10}39.7423 $\pm$ & \cellcolor{black!25}0.2672 $\pm$ & 0.6812 $\pm$ & \cellcolor{black!10}0.3193 $\pm$\\
			& \cellcolor{black!10}0.0143 & 0.0119 ($+$) & \cellcolor{black!25}0.0098 ($\sim$) & \cellcolor{black!25}0.0061 & 0.0048 ($+$) & \cellcolor{black!10}0.0006 ($+$) & \cellcolor{black!25}2.4365 & 2.8705 ($+$) & \cellcolor{black!10}1.7614 ($+$) & \cellcolor{black!25}0.0037 & 0.0092 ($+$) & \cellcolor{black!10}0.0007 ($+$)\\
			\hline
			Sum-up & $+/-/\sim$ & 15/0/1 & 13/1/2 & $+/-/\sim$ & 16/0/0 & 13/0/3 & $+/-/\sim$ & 16/0/0 & 10/0/6 & $+/-/\sim$ & 16/0/0 & 9/1/6\\
			\hline
	\end{tabular}}
\end{table*}

From Tables \ref{tab:DS_2obj} and \ref{tab:OS_2obj}, the following insights are obtained for LORD:
\begin{itemize}
	\item LORD is superior to DN-NSGA-II \cite{DN-NSGA2} as DN-NSGA-II neglects the solution diversity in the objective space. Thus, the solution distribution also suffers in the decision space by the neighborhood property \cite{NAEMO}.
	\item While NSGA-II is the second-best in the objective space (Table \ref{tab:OS_2obj}), it neglects the solution diversity in the decision space and thus, gets outperformed by LORD (Table \ref{tab:DS_2obj}).
	\item While MO\_Ring\_PSO\_SCD is the second-best in the decision space (Table \ref{tab:DS_2obj}), it often gets trapped in the local optima (as seen from estimated PSs and PFs in the supplementary material) leading to poor performance for some MMMOPs (e.g., MMF11 and MMF12). As LORD efficiently addresses the crowding illusion problem (Section \ref{ss:decompose_DS}), it has superior performance in most cases.
	\item The performance of LORD remains consistent (Tables \ref{tab:DS_2obj} and \ref{tab:OS_2obj}), even for high $k_{PS}$ (e.g., Omni-test with $k_{PS}=27$). It can successfully overcome the local optima (e.g., MMF10) and thus, also, acts as an excellent MOEA. The similarity between rPSP (=IGDX/cov\_rate) and IGDX values imply that cover rate (cov\_rate) is nearly equal to one (ideal value) \cite{MO_Ring_PSO_SCD}.
\end{itemize}

\begin{figure*}[!t]
	\centering
	\resizebox{\linewidth}{!}{
		\begin{tabular}{|c|cc|cc|}
			\hline
			& \multicolumn{2}{c|}{3-objective problems} & \multicolumn{2}{c|}{8-objective problems}\\
			\cline{2-5}
			& \multicolumn{2}{c|}{Estimated Pareto-optimal Set (PS)} & \multicolumn{2}{c|}{Estimated Pareto-optimal Set (PS) Projected}\\
			& \multicolumn{2}{c|}{\space} & \multicolumn{2}{c|}{on Last Two Dimensions ($x_8$ vs. $x_7$)} \\
			\hline
			\rotatebox[origin=c]{90}{LORD-II} &
			\raisebox{-.5\height}{\includegraphics[width=0.25\linewidth]{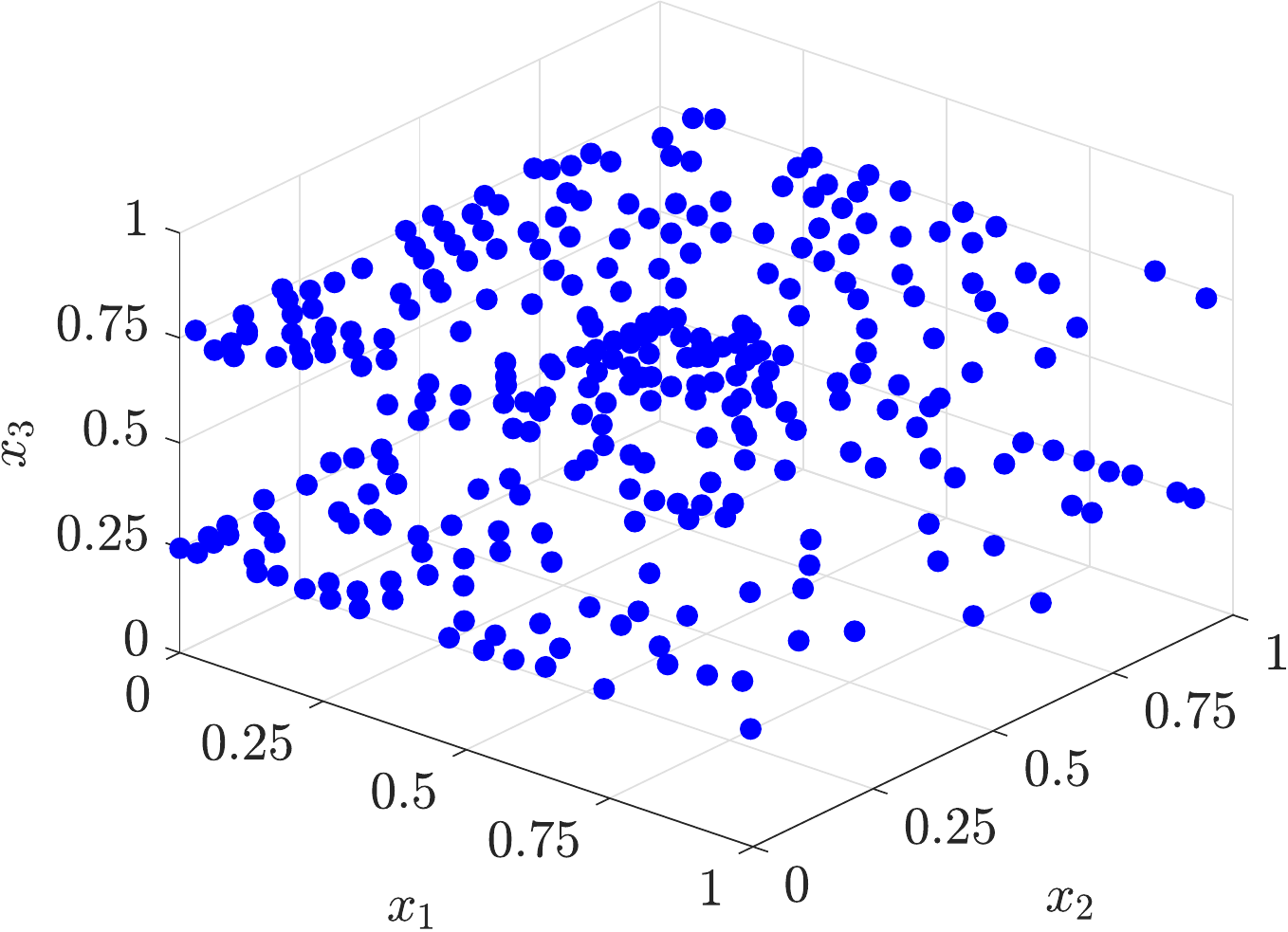}} & \raisebox{-.5\height}{\includegraphics[width=0.25\linewidth]{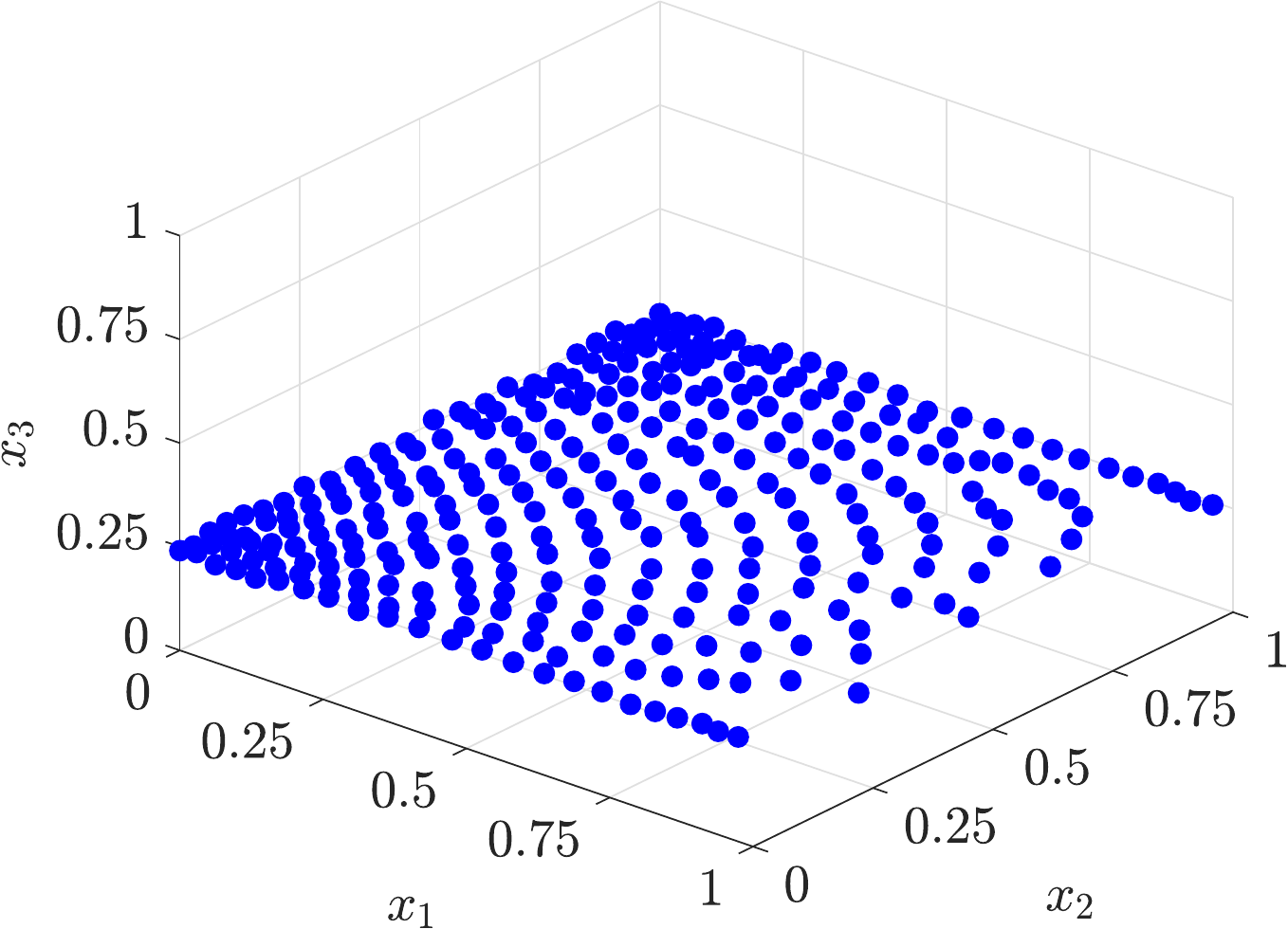}} & \raisebox{-.5\height}{\includegraphics[width=0.25\linewidth]{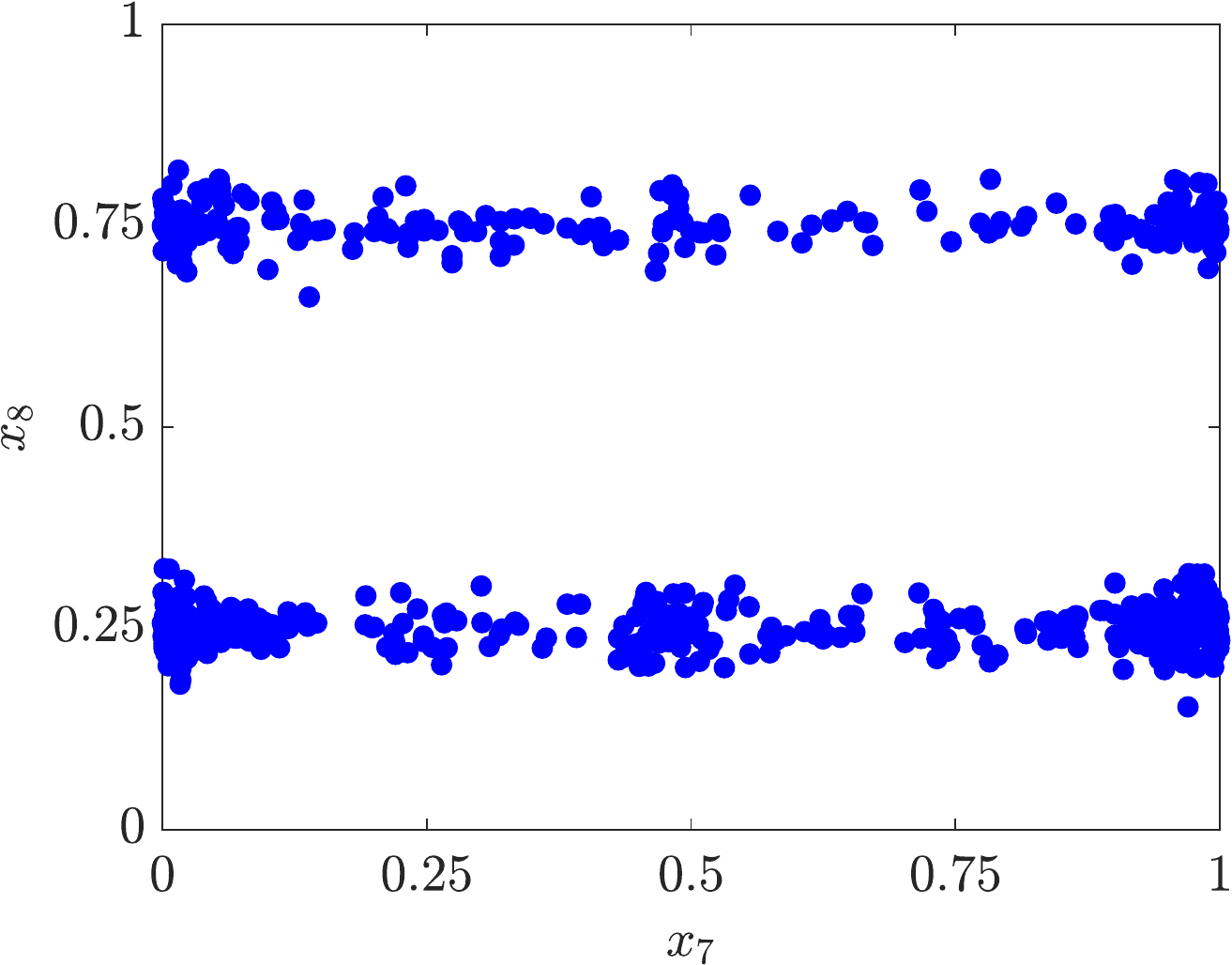}} & \raisebox{-.5\height}{\includegraphics[width=0.25\linewidth]{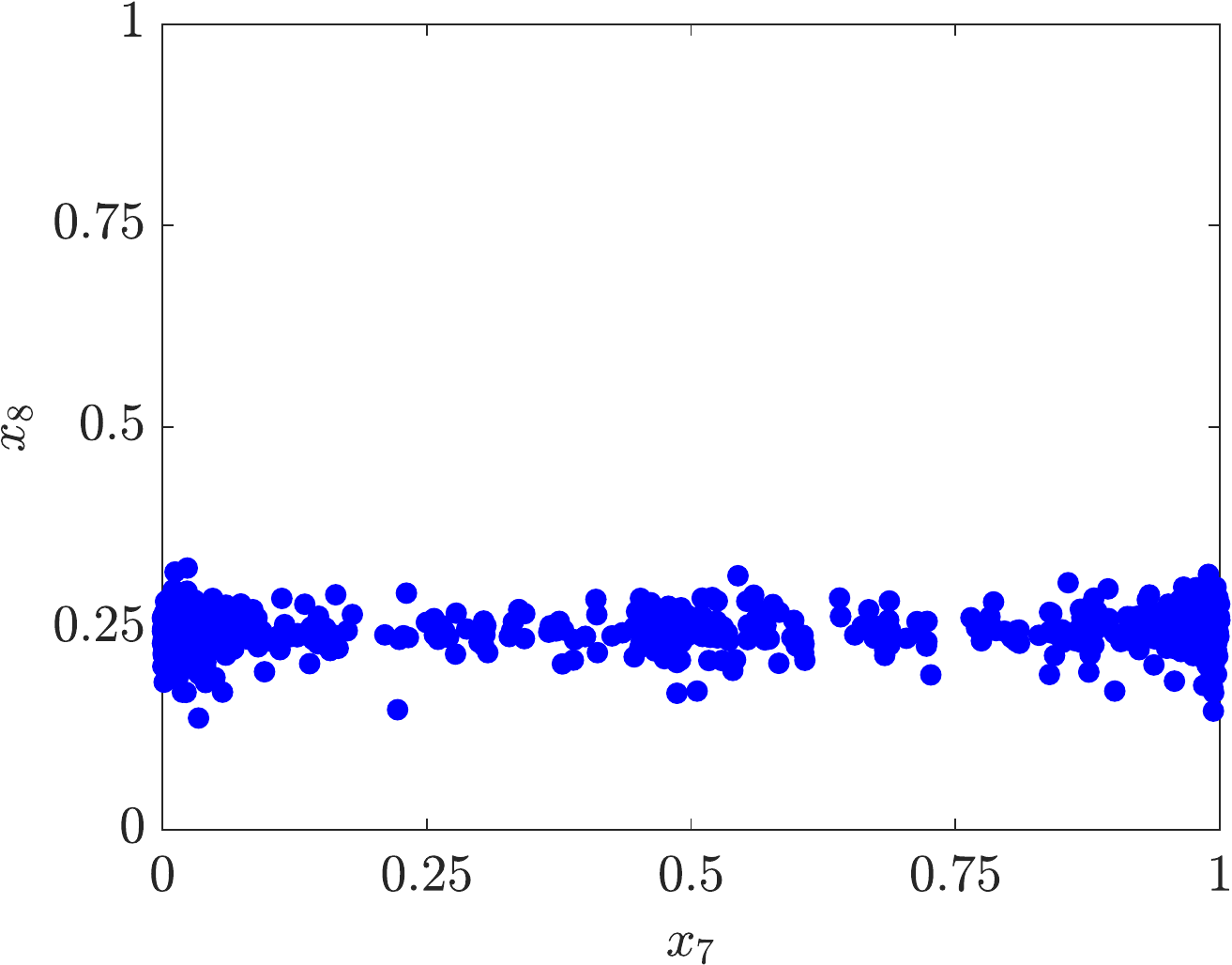}} \\
			& (a) MMF14 & (b) MMF15 & (c) MMF14 & (d) MMF15\\
			\hline
			\rotatebox[origin=c]{90}{MO\_Ring\_PSO\_SCD} &
			\raisebox{-.5\height}{\includegraphics[width=0.25\linewidth]{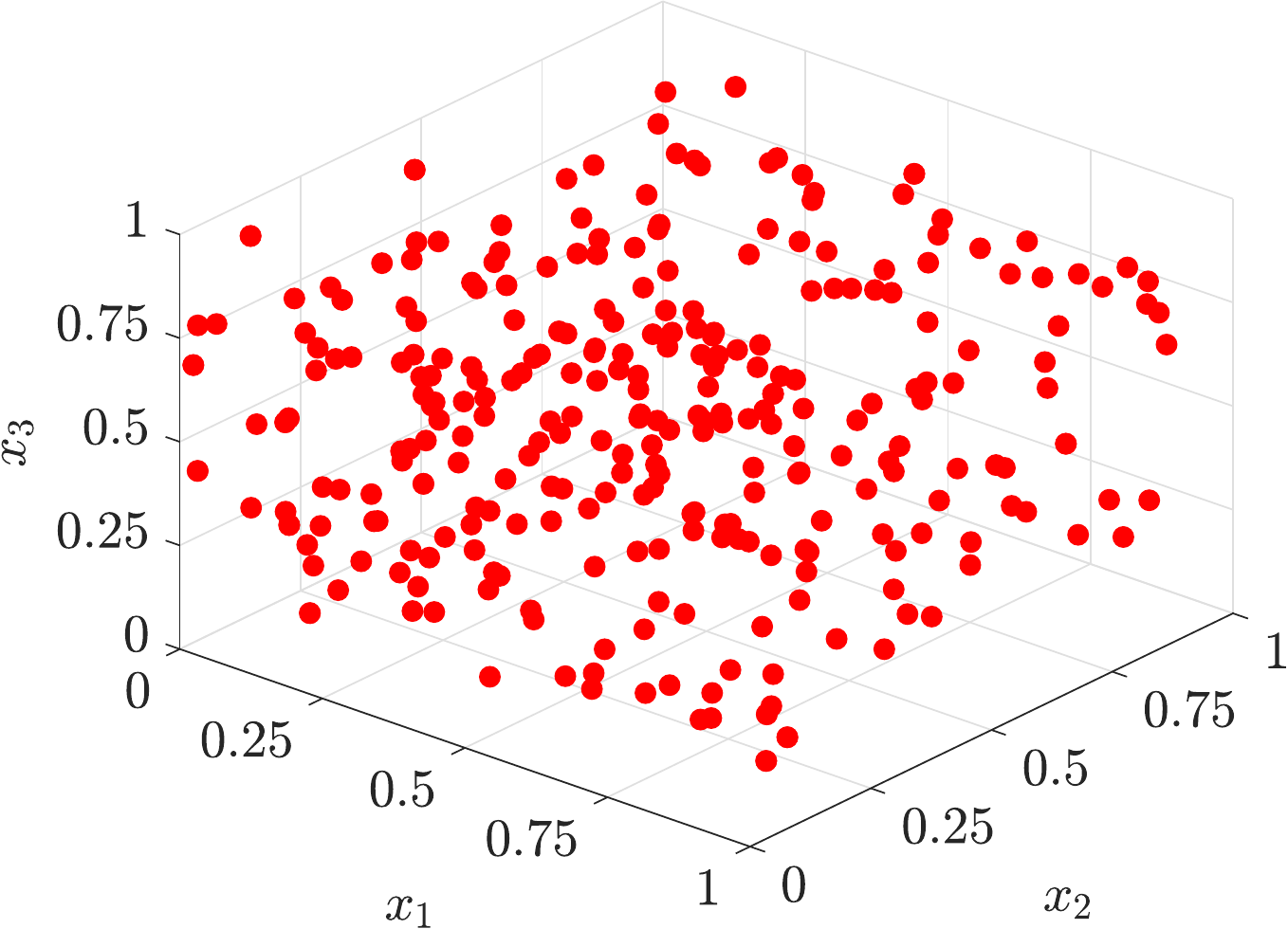}} & \raisebox{-.5\height}{\includegraphics[width=0.25\linewidth]{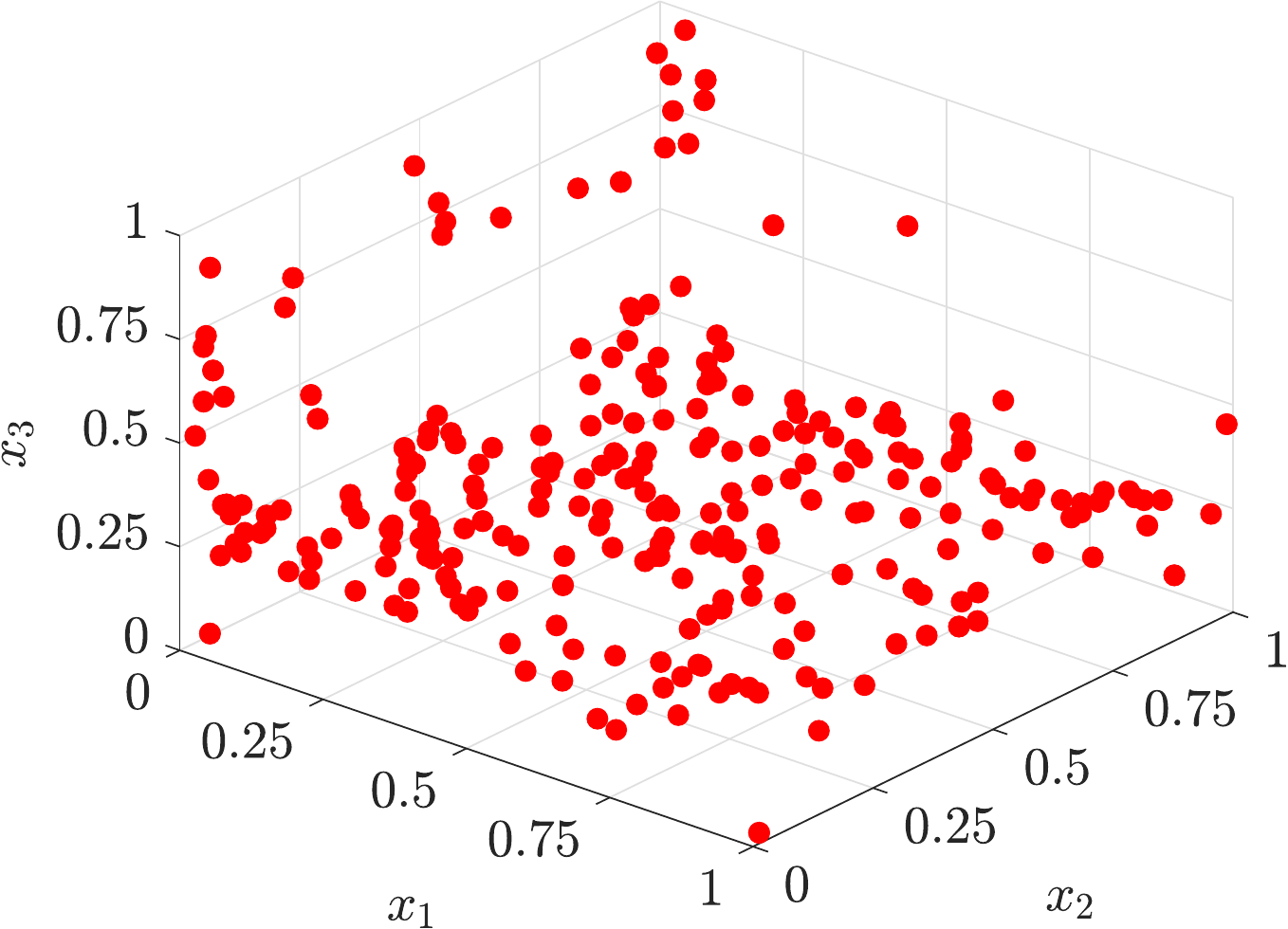}} & \raisebox{-.5\height}{\includegraphics[width=0.25\linewidth]{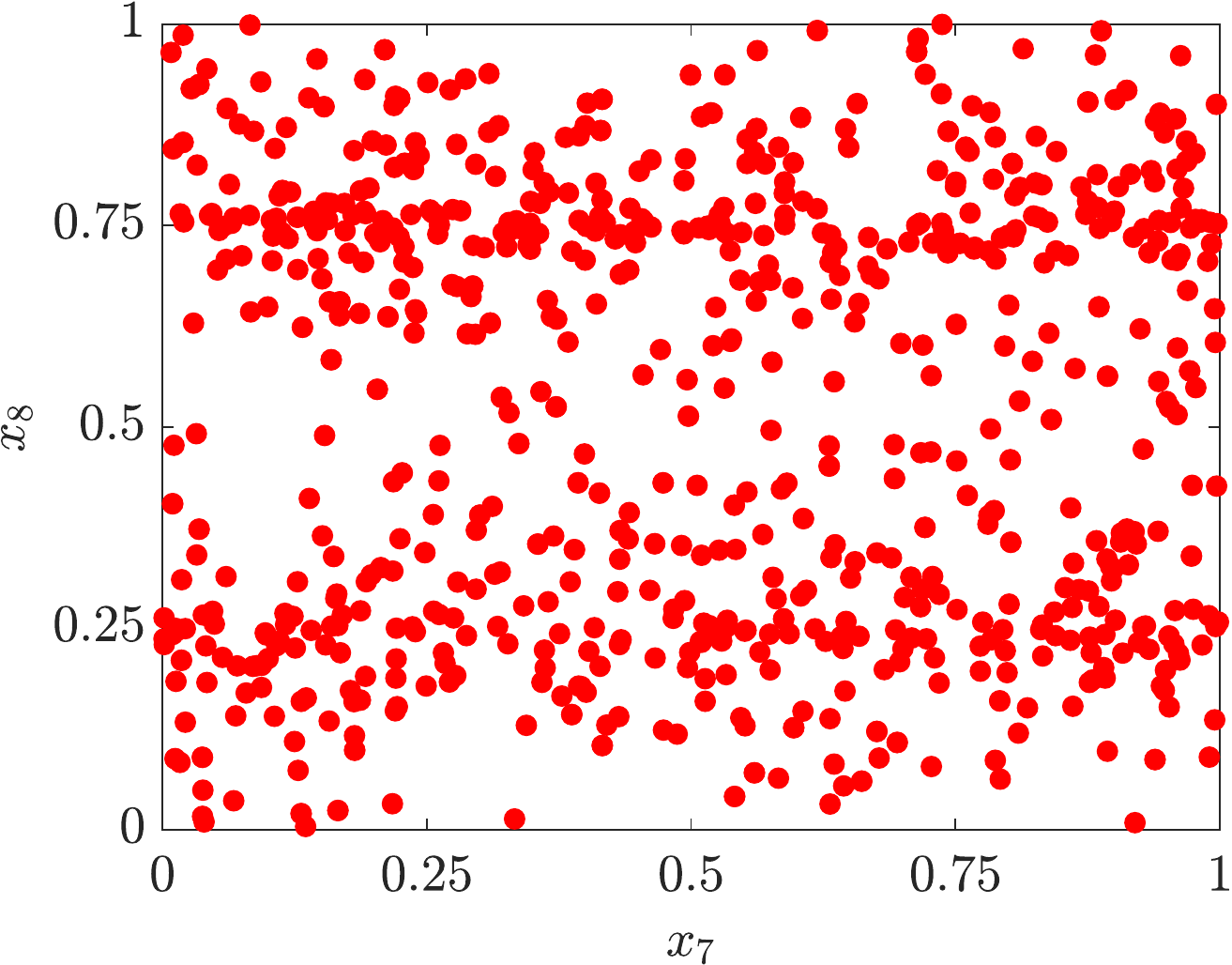}} & \raisebox{-.5\height}{\includegraphics[width=0.25\linewidth]{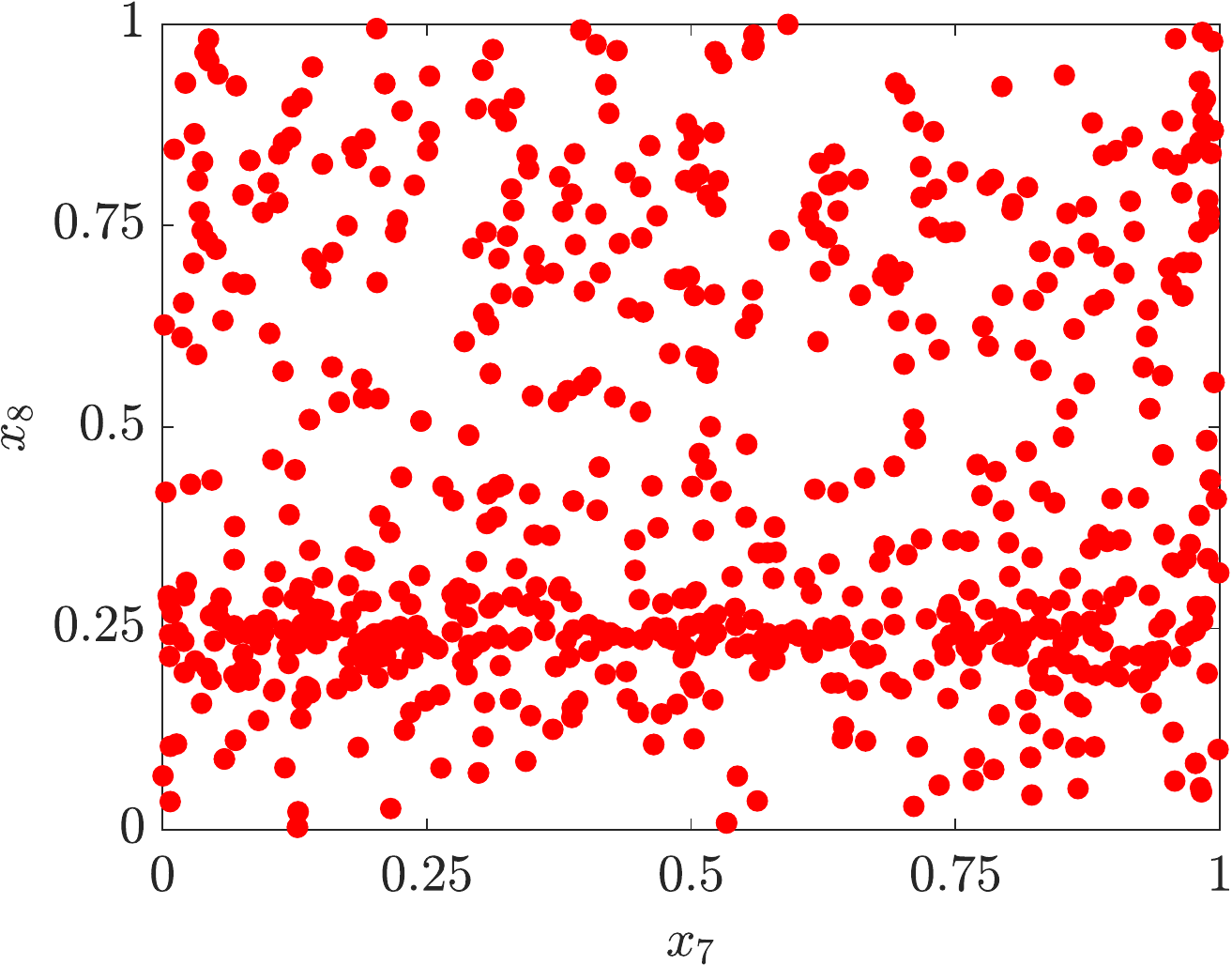}} \\
			& (e) MMF14 & (f) MMF15 & (g) MMF14 & (h) MMF15 \\
			\hline
	\end{tabular}}
	\caption{Visualizations of Pareto-optimal Set (PS) 3- and 8-objective MMF14 and MMF15 problems for median runs by LORD-II (top row) and MO\_Ring\_PSO\_SCD (bottom row).}
	\label{fig:LORD2_vis1}
\end{figure*}
\begin{figure*}[!t]
	\centering
	\resizebox{\linewidth}{!}{
		\begin{tabular}{|cc|cc|}
			\hline
			\multicolumn{2}{|c|}{LORD-II} & \multicolumn{2}{c|}{MO\_Ring\_PSO\_SCD}\\
			\hline
			\raisebox{-.5\height}{\includegraphics[width=0.35\linewidth]{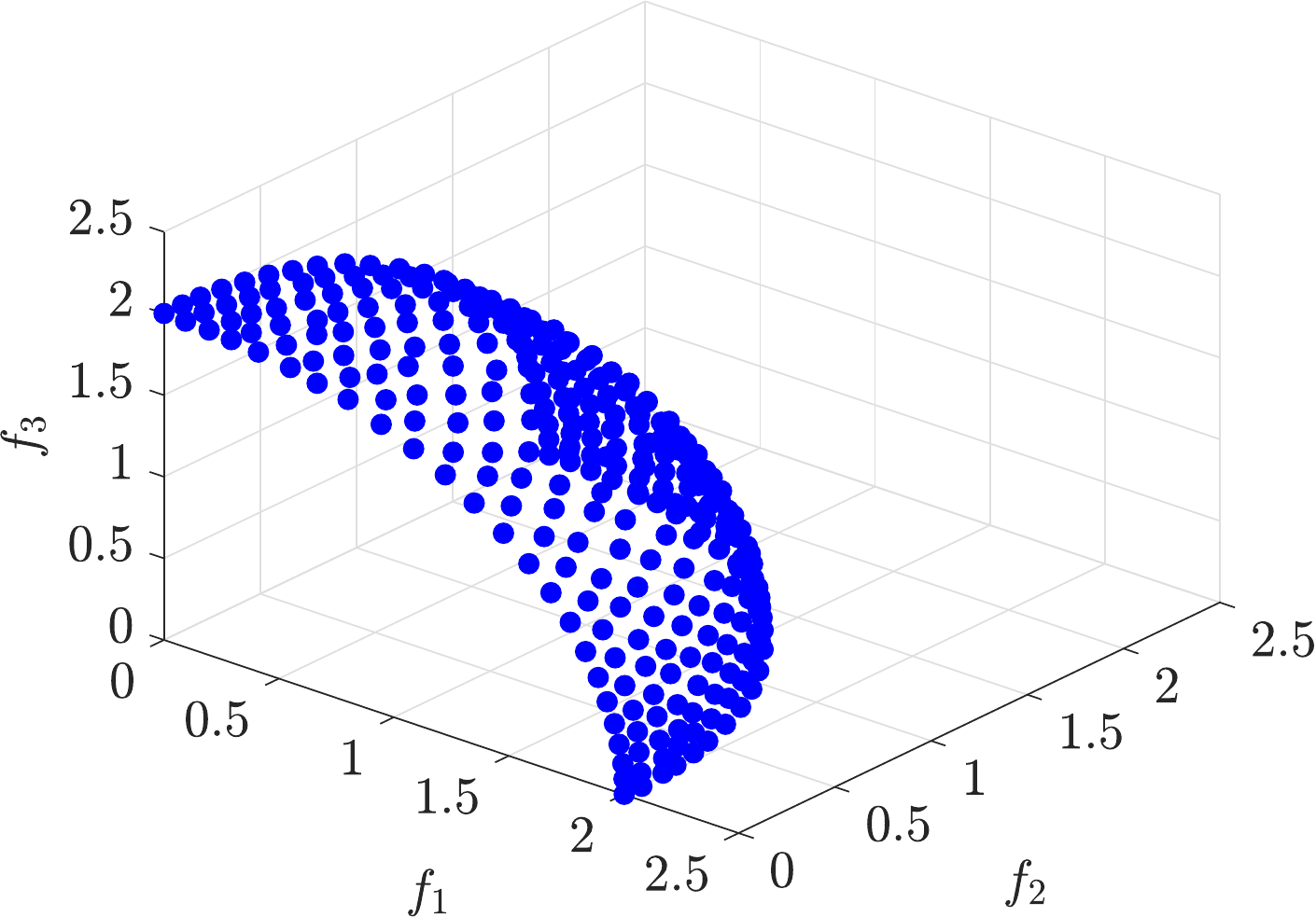}} & \raisebox{-.5\height}{\includegraphics[width=0.25\linewidth]{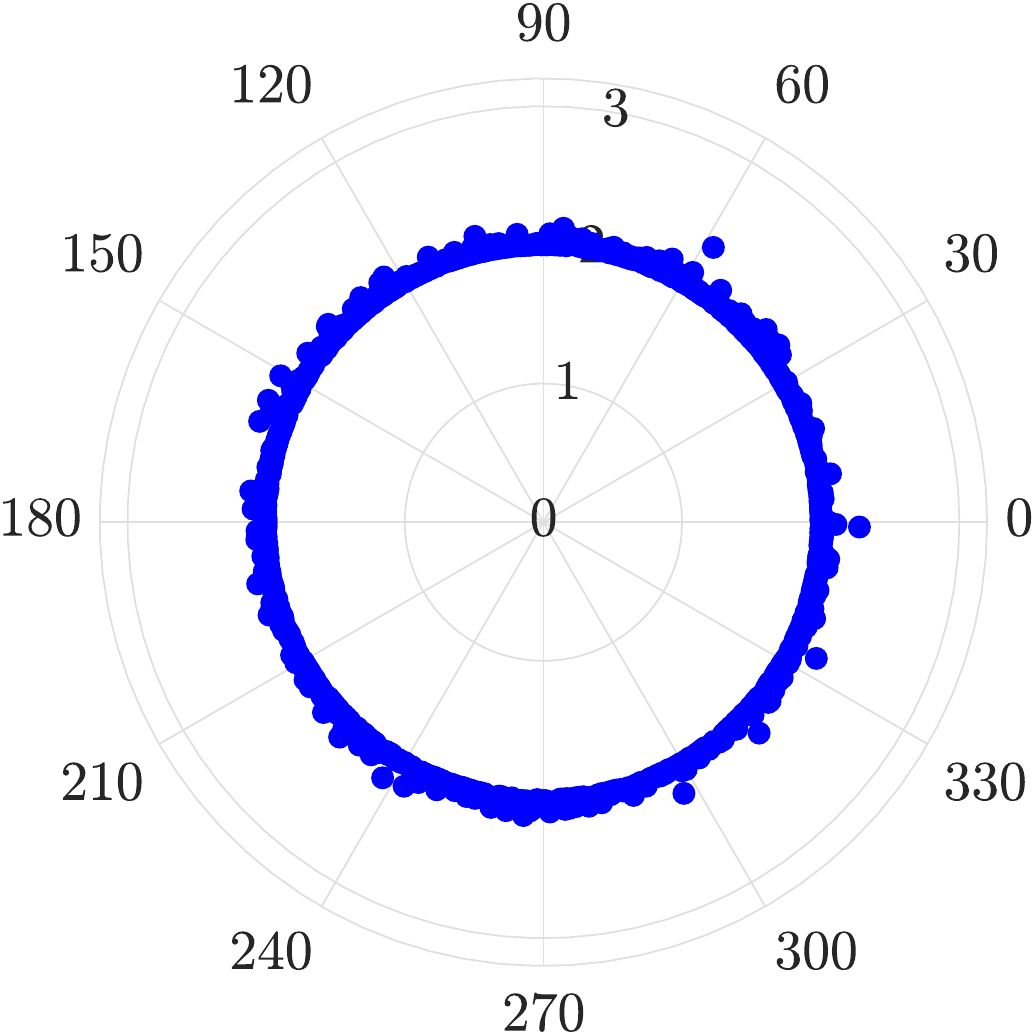}} & \raisebox{-.5\height}{\includegraphics[width=0.35\linewidth]{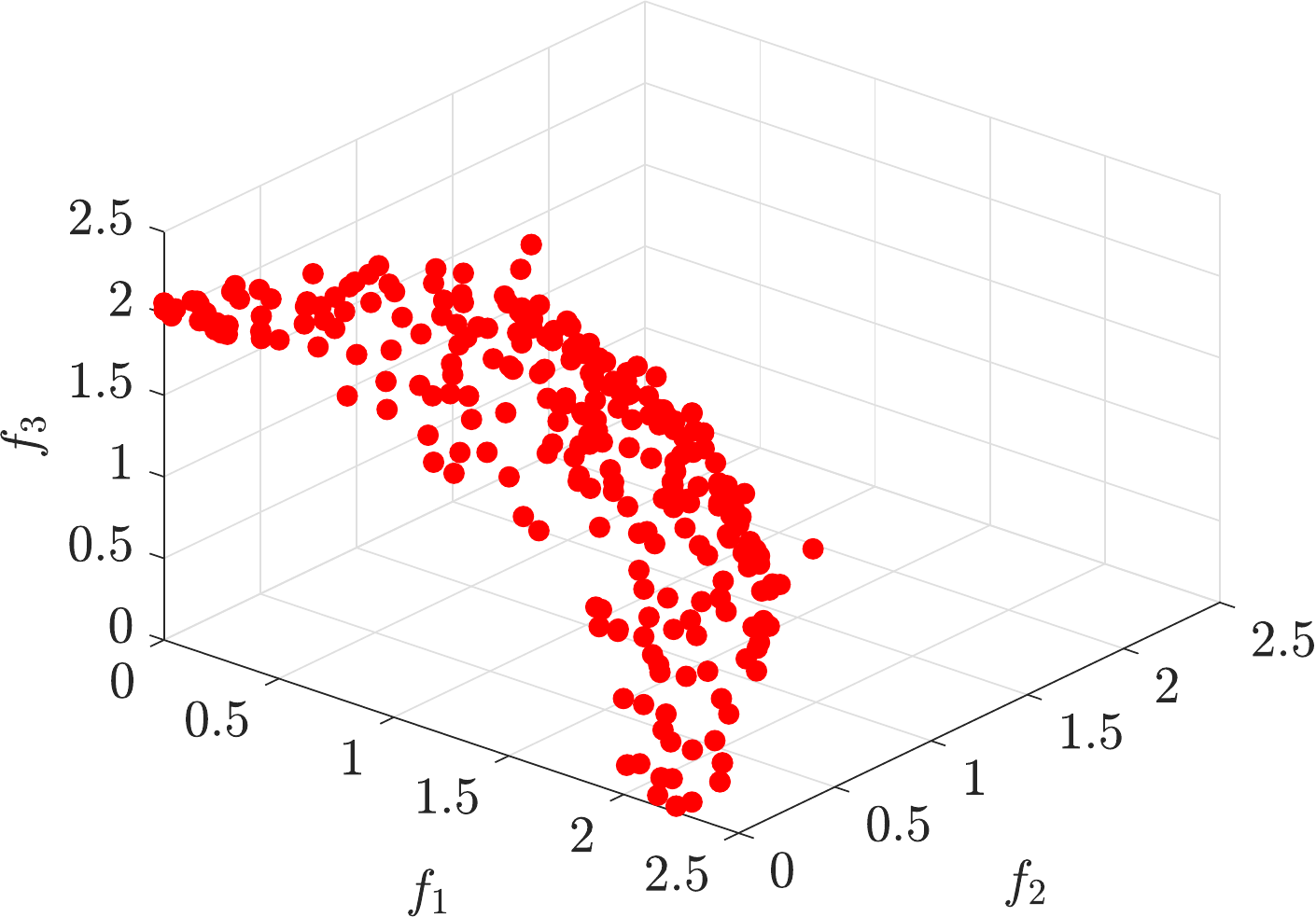}} & \raisebox{-.5\height}{\includegraphics[width=0.25\linewidth]{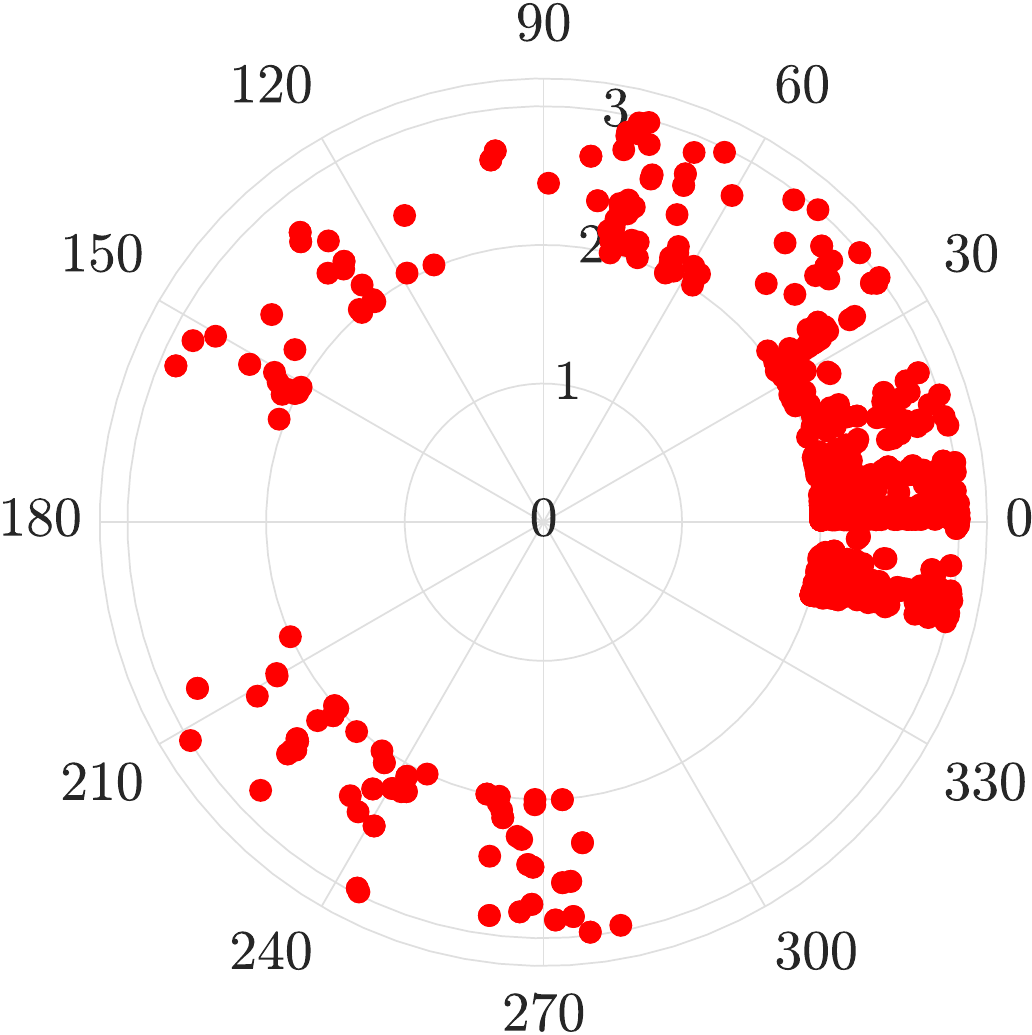}} \\
			(a) $M=3$ & (b) $M=8$ & (c) $M=3$ & (d) $M=8$\\
			\hline
	\end{tabular}}
	\caption{Visualizations of estimated Pareto-Front (PF) of $M$-objective MMF14 problem for median runs by LORD-II and MO\_Ring\_PSO\_SCD. Results are similar for MMF15 as well.}
	\label{fig:LORD2_vis2}
\end{figure*}
The following insights can be obtained for LORD-II from Table \ref{tab:res_Mobj}:
\begin{itemize}
	\item In the objective space (Table \ref{tab:res_Mobj}), both LORD-II and MOEA/DD have similar performance for 3-objective problems. For 5-objective problems, LORD-II is marginally outperformed in only one case by MOEA/DD. For 8- and 10-objective problems, LORD-II is superior. In all cases, LORD-II outperforms MO\_Ring\_PSO\_SCD in both convergence (CM) and diversity ($D\_metric$) as also seen in Fig. \ref{fig:LORD2_vis2}.
	 
	\item In the decision space (Table \ref{tab:res_Mobj}), LORD-II maintains superiority.
	
	\item The estimated PS and PF from LORD-II (Figs. \ref{fig:LORD2_vis1} and \ref{fig:LORD2_vis2}, respectively) demonstrate excellent convergence and diversity. The results from MO\_Ring\_PSO\_SCD deteriorate severely with an increase in dimension. In contrast to MO\_Ring\_PSO\_SCD (Fig. \ref{fig:LORD2_vis2}d), the polar plot \cite{radial_plot} from LORD-II (Fig. \ref{fig:LORD2_vis2}b) converges all solutions to a near-global PF, forming a uniformly distributed circle for 8-objective MMMaOPs. 
\end{itemize}

2) \textbf{Experiment-II: Comparison with Reference Vector Assisted MMMOEAs}: The performance of two recent reference-vector assisted MMMOEAs (DE-TriM \cite{DETriM} and MM-NAEMO \cite{MMNAEMO_CEC19}) are compared with LORD and LORD-II in Table \ref{tab:ref_EAs_LORD} on CEC 2019 MMMOPs \cite{CEC19_TR}. The results of this experiment are also compared with MO\_Ring\_PSO\_SCD to fairly assess the relative rankings of algorithms. Each of these algorithms (DE-TriM, MM-NAEMO and MO\_Ring\_PSO\_SCD) are set up using the parameters recommended in \cite{DETriM}, \cite{MMNAEMO_CEC19} and \cite{MO_Ring_PSO_SCD}, respectively.
\begin{table}[!ht]
	\centering
	\caption{Mean of IGDX and IGDF over 51 independent runs for comparing reference-vector guided MMMOEAs on $2$- and $3$-objective MMMOPs.}
	\label{tab:ref_EAs_LORD}
	\resizebox{\linewidth}{!}{
		\begin{tabular}{|c|c|c|c|c|c|c|c|c|}
			\hline
			& \multicolumn{4}{c|}{\textbf{IGDX}} & \multicolumn{4}{c|}{\textbf{IGDF}}\\
			\cline{2-9}
			\textbf{2-objective} & & & & \textbf{MO\_Ring\_} & & & & \textbf{MO\_Ring\_}\\
			\textbf{Problems} & \textbf{LORD} & \textbf{DE-TriM} & \textbf{MM-NAEMO} & \textbf{PSO\_SCD} & \textbf{LORD} & \textbf{DE-TriM} & \textbf{MM-NAEMO} & \textbf{PSO\_SCD}\\
			\hline
			MMF1 & \cellcolor{black!25}0.0431 & \cellcolor{black!10}0.0465 ($+$) & 0.0486 ($+$) & 0.0485 ($+$) & \cellcolor{black!25}0.0025 & \cellcolor{black!10}0.0026 ($\sim$) & 0.0040 ($+$) & 0.0037 ($+$)\\
			MMF1\_z & \cellcolor{black!10}0.0351 & 0.0503 ($+$) & \cellcolor{black!25}0.0347 ($\sim$) & 0.0352 ($\sim$) & \cellcolor{black!25}0.0022 & \cellcolor{black!10}0.0026 ($+$) & 0.0035 ($+$) & 0.0036 ($+$)\\
			MMF1\_e & 0.7499 & 2.8757 ($+$) & \cellcolor{black!25}0.4115 ($-$) & \cellcolor{black!10}0.4738 ($-$) & \cellcolor{black!25}0.0029 & \cellcolor{black!25}0.0029 ($\sim$) & \cellcolor{black!10}0.0051 ($+$) & 0.0119 ($+$)\\
			MMF2 & \cellcolor{black!10}0.0180 & 0.0505 ($+$) & \cellcolor{black!25}0.0118 ($-$) & 0.0416 ($+$) & \cellcolor{black!10}0.0070 & \cellcolor{black!25}0.0035 ($-$) & 0.0083 ($+$) & 0.0207 ($+$)\\
			MMF3 & \cellcolor{black!10}0.0176 & 0.0235 ($+$) & \cellcolor{black!25}0.0137 ($-$) & 0.0276 ($+$) & \cellcolor{black!10}0.0069 & \cellcolor{black!25}0.0047 ($-$) & 0.0085 ($+$) & 0.0154 ($+$)\\
			MMF4 & \cellcolor{black!10}0.0251 & \cellcolor{black!25}0.0211 ($-$) & 0.0312 ($+$) & 0.0271 ($+$) & \cellcolor{black!25}0.0018 & \cellcolor{black!10}0.0025 ($+$) & 0.0033 ($+$) & 0.0037 ($+$)\\ 
			MMF5 & \cellcolor{black!25}0.0814 & 0.0892 ($+$) & 0.0871 ($+$) & \cellcolor{black!10}0.0857 ($+$) & \cellcolor{black!25}0.0024 & \cellcolor{black!10}0.0027 ($+$) & 0.0037 ($+$) & 0.0037 ($+$)\\
			MMF6 & \cellcolor{black!25}0.0692 & 0.0756 ($+$) & 0.0743 ($+$) & \cellcolor{black!10}0.0736 ($+$) & \cellcolor{black!25}0.0023 & \cellcolor{black!10}0.0025 ($\sim$) & 0.0036 ($+$) & 0.0035 ($+$)\\
			MMF7 & \cellcolor{black!10} 0.0218 & \cellcolor{black!25}0.0201 ($-$) & 0.0229 ($+$) & 0.0262 ($+$) & \cellcolor{black!25}0.0022 & \cellcolor{black!10}0.0025 ($+$) & 0.0035 ($+$) & 0.0037 ($+$)\\ 
			MMF8 & \cellcolor{black!10}0.0762 & 0.0989 ($+$) & 0.3348 ($+$) & \cellcolor{black!25}0.0673 ($\sim$) & \cellcolor{black!25}0.0025 & \cellcolor{black!10}0.0029 ($+$) & 0.0037 ($+$) & 0.0048 ($+$)\\
			MMF9 & \cellcolor{black!25}0.0046 & 0.0787 ($+$) & \cellcolor{black!10}0.0048 ($\sim$) & 0.0079 ($+$) & \cellcolor{black!25}0.0085 & \cellcolor{black!10}0.0119 ($+$) & 0.0479 ($+$) & 0.0160 ($+$)\\
			MMF10 & \cellcolor{black!25}0.0018 & \cellcolor{black!25}0.0018 ($\sim$) & \cellcolor{black!10}0.0121 ($+$) & 0.0276 ($+$) & \cellcolor{black!25}0.0061 & \cellcolor{black!10}0.0080 ($+$) & 0.0639 ($+$) & 0.1128 ($+$)\\
			MMF11 & \cellcolor{black!25}0.0029 & \cellcolor{black!10}0.0036 ($+$) & 0.0418 ($+$) & 0.0054 ($+$) & \cellcolor{black!25}0.0082 & \cellcolor{black!10}0.0109 ($+$) & 0.0931 ($+$) & 0.0176 ($+$)\\
			MMF12 & \cellcolor{black!25}0.0013 & \cellcolor{black!25}0.0013 ($\sim$) & 0.0050 ($+$) & \cellcolor{black!10}0.0038 ($+$) & \cellcolor{black!25}0.0020 & \cellcolor{black!10}0.0021 ($\sim$) & 0.0196 ($+$) & 0.0068 ($+$)\\
			MMF13 & \cellcolor{black!25}0.0242 & 0.0368 ($+$) & 0.1878 ($+$) & \cellcolor{black!10}0.0314 ($+$) & \cellcolor{black!25}0.0063 & \cellcolor{black!10}0.0094 ($+$) & 0.1059 ($+$) & 0.0264 ($+$)\\
			Omni-test & \cellcolor{black!25}0.0706 & \cellcolor{black!10}0.0732 ($+$) & 0.1511 ($+$) & 0.3907 ($+$) & \cellcolor{black!25}0.0091 & \cellcolor{black!10}0.0125 ($+$) & 0.0130 ($+$) & 0.0422 ($+$)\\
			SYM-PART-simple & \cellcolor{black!10}0.0549 & 0.0740 ($+$) & 0.1115 ($+$) & \cellcolor{black!25}0.0300 ($-$) & \cellcolor{black!10}0.0165 & \cellcolor{black!25}0.0101 ($-$) & 0.0472 ($+$) & 0.0419 ($+$)\\
			SYM-PART-rotated & \cellcolor{black!25}0.1558 & \cellcolor{black!10}0.1885 ($+$) & 0.7586 ($+$) & 0.2926 ($+$) & \cellcolor{black!10}0.0178 & \cellcolor{black!25}0.0125 ($-$) & 0.0395 ($+$) & 0.0467 ($+$)\\
			\hline
			\multicolumn{2}{|c|}{LORD vs. others ($+/-/\sim$)} & 14/2/2 & 13/3/2 & 14/2/2 & ($+/-/\sim$) & 10/4/4 & 18/0/0 & 18/0/0\\
			\hline
			\textbf{3-objective} & & & & \textbf{MO\_Ring\_} & & & & \textbf{MO\_Ring\_}\\
			\textbf{Problems} & \textbf{LORD-II} & \textbf{DE-TriM} & \textbf{MM-NAEMO} & \textbf{PSO\_SCD} & \textbf{LORD-II} & \textbf{DE-TriM} & \textbf{MM-NAEMO} & \textbf{PSO\_SCD}\\
			\hline
			MMF14 & \cellcolor{black!25}0.0443 & 0.0558 ($+$) & \cellcolor{black!10}0.0465 ($+$) & 0.0539 ($+$) & \cellcolor{black!25}0.0540 & \cellcolor{black!10}0.0749 ($+$) & 0.0808 ($+$) & 0.0801 ($+$)\\ 
			MMF14\_a & \cellcolor{black!25}0.0576 & 0.0676 ($+$) & 0.0663 ($+$) & \cellcolor{black!10}0.0613 ($+$) & \cellcolor{black!25}0.0561 & 0.0809 ($+$) & 0.0791 ($+$) & \cellcolor{black!10}0.0789 ($+$)\\
			MMF15 & \cellcolor{black!25}0.0287 & \cellcolor{black!10} 0.0361 ($+$) & 0.0518 ($+$) & 0.0419 ($+$) & \cellcolor{black!25}0.0548 & \cellcolor{black!10}0.0787 ($+$) & 0.1113 ($+$) & 0.0854 ($+$)\\
			MMF15\_a & \cellcolor{black!25}0.0355 & 0.0503 ($+$) & 0.0848 ($+$) & \cellcolor{black!10}0.0452 ($+$) & \cellcolor{black!25}0.0571 & 0.0951 ($+$) & 0.1263 ($+$) & \cellcolor{black!10}0.0841 ($+$)\\
			\hline
			\multicolumn{2}{|c|}{LORD-II vs. others ($+/-/\sim$)} & 4/0/0 & 4/0/0 & 4/0/0 & ($+/-/\sim$) & 4/0/0 & 4/0/0 & 4/0/0\\
			\hline
	\end{tabular}}
\end{table}

From Table \ref{tab:ref_EAs_LORD}, LORD-II is noted to have the best performance in all cases and LORD is noted to have the best or the second-best performance in both objective and decision spaces for most of the cases. Unlike other MMMOEAs \cite{MO_Ring_PSO_SCD,MMNAEMO_CEC19,ZoneS} which yield poor convergence and/or diversity in objective space in order to improve the performance in decision space, LORD and LORD-II perform satisfactorily in both the spaces and competitively outperform the other reference vector assisted MMMOEAs.

3) \textbf{Experiment-III: Comparison on Polygon MMMaOPs}: Similar to \cite{NIMMO_SWEVO19}, the mean IGDX and IGDF of LORD-II are compared with NIMMO on Polygon test problems as both the MMMOEAs are designed for MMMaOPs. The results of MO\_Ring\_PSO\_SCD \cite{MO_Ring_PSO_SCD}, Omni-Optimizer \cite{Omni} and TriMOEA\_TA\&R \cite{TriMOEA_TAR} are also compared. 

The performance of these MMMOEAs are noted on $M$-objective polygon and rotated polygon MMMaOPs \cite{POLY_PROBS} in Tables \ref{tab:poly_igdx_LORD} (using IGDX) and \ref{tab:poly_igdf_LORD} (using IGDF). This experiment considers the specifications mentioned in Table \ref{tab:specs_poly_LORD} as recommended in \cite{NIMMO_SWEVO19}. The performance values of the other MMMOEAs (except LORD-II) are also noted from \cite{NIMMO_SWEVO19}. The remaining parameters of LORD-II are set up as specified in Table \ref{tab:specs_LORD}.
\begin{table}[!t]
	\centering
	\caption{Specifications for the experiment conducted on polygon and rotated polygon problems according to recommendations in \cite{NIMMO_SWEVO19}.}
	\label{tab:specs_poly_LORD}
	\resizebox{0.85\linewidth}{!}{
		\begin{tabular}{|c|c|c|c|}
			\hline
			\multicolumn{2}{|c|}{\space} & & \textbf{Used by NIMMO, TriMOEA\_TA\&R,}\\
			\multicolumn{2}{|c|}{\textbf{Parameters}} & \textbf{Used by LORD-II} & \textbf{MO\_Ring\_PSO\_SCD, and} \\
			\multicolumn{2}{|c|}{\space} & & \textbf{Omni-Optimizer in \cite{NIMMO_SWEVO19}}\\
			\hline
			\multirow{4}{*}{\rotatebox[origin=c]{90}{$n_{pop}$}} & 3-obj & 210 & 210\\
			& 5-obj & 210 & 210\\
			& 8-obj & 156 & 156\\ 
			& 10-obj & 210 & 210\\
			\hline
			\multicolumn{2}{|c|}{\#runs} & 31 & 31\\
			\hline
			\multicolumn{2}{|c|}{$MaxFES$} & 10000 & 10000\\
			\hline
			\multicolumn{2}{|c|}{$N_{IGD}$} & 5000 & 5000\\
			\hline
	\end{tabular}}
\end{table}
\begin{table}[!ht]
	\centering
	\caption{Mean IGDX over 31 independent runs for comparing LORD-II on $M$-objective polygon and rotated polygon (RPolygon) problems.}
	\label{tab:poly_igdx_LORD}
	\resizebox{0.8\linewidth}{!}{
		\begin{tabular}{|c|c|c|c|c|c|}
			\hline
			& & & \textbf{TriMOEA} & \textbf{MO\_Ring\_} & \textbf{Omni-}\\
			$M$\textbf{-Problems} & \textbf{LORD-II} & \textbf{NIMMO} & \textbf{\_TA\&R} & \textbf{PSO\_SCD} & \textbf{Optimizer}\\
			\hline
			3-Polygon & \cellcolor{black!25}0.0054 & \cellcolor{black!10}0.0056 ($+$) & 0.0063 ($+$) & 0.0091 ($+$) & 0.0083 ($+$)\\
			3-RPolygon & \cellcolor{black!10}0.0064 & \cellcolor{black!25}0.0059 ($-$) & 0.0295 ($+$) & 0.0090 ($+$) & 0.0085 ($+$)\\
			5-Polygon & \cellcolor{black!25}0.0055 & \cellcolor{black!10}0.0070 ($+$) & 0.0162 ($+$) & 0.0113 ($+$) & 0.0110 ($+$)\\
			5-RPolygon & \cellcolor{black!25}0.0062 & \cellcolor{black!10}0.0074 ($+$) & 0.0400 ($+$) & 0.0113 ($+$) & 0.0110 ($+$)\\
			8-Polygon & \cellcolor{black!25}0.0046 & \cellcolor{black!10}0.0089 ($+$) & 0.0136 ($+$) & 0.0143 ($+$) & 0.0140 ($+$)\\
			8-RPolygon & \cellcolor{black!25}0.0051 & \cellcolor{black!10}0.0093 ($+$) & 0.0747 ($+$) & 0.0144 ($+$) & 0.0138 ($+$)\\
			10-Polygon & \cellcolor{black!25}0.0044 & \cellcolor{black!10}0.0072 ($+$) & 0.0123 ($+$) & 0.0120 ($+$) & 0.0112 ($+$)\\
			10-RPolygon & \cellcolor{black!25}0.0053 & \cellcolor{black!10}0.0076 ($+$) & 0.0404 ($+$) & 0.0118 ($+$) & 0.0112 ($+$)\\
			\hline
			\multicolumn{2}{|c|}{LORD-II vs. others ($+/-/\sim$)} & 7/1/0 & 8/0/0 & 8/0/0 & 8/0/0\\
			\hline
	\end{tabular}}
\end{table}
\begin{table}[!ht]
	\centering
	\caption{Mean IGDF over 31 independent runs for comparing LORD-II on $M$-objective polygon and rotated polygon (RPolygon) problems.}
	\label{tab:poly_igdf_LORD}
	\resizebox{0.8\linewidth}{!}{
		\begin{tabular}{|c|c|c|c|c|c|}
			\hline
			& & & \textbf{TriMOEA} & \textbf{MO\_Ring\_} & \textbf{Omni-}\\
			$M$\textbf{-Problems} & \textbf{LORD-II} & \textbf{NIMMO} & \textbf{\_TA\&R} & \textbf{PSO\_SCD} & \textbf{Optimizer}\\
			\hline
			3-Polygon & \cellcolor{black!25}0.0023 & \cellcolor{black!10}0.0025 ($+$) & 0.0040 ($+$) & 0.0034 ($+$) & 0.0028 ($+$)\\
			3-RPolygon & \cellcolor{black!25}0.0023 & \cellcolor{black!10}0.0025 ($+$) & 0.0046 ($+$) & 0.0034 ($+$) & 0.0028 ($+$)\\
			5-Polygon & \cellcolor{black!25}0.0031 & \cellcolor{black!10}0.0044 ($+$) & 0.0149 ($+$) & 0.0057 ($+$) & 0.0051 ($+$)\\
			5-RPolygon & \cellcolor{black!25}0.0030 & \cellcolor{black!10}0.0044 ($+$) & 0.0149 ($+$) & 0.0058 ($+$) & 0.0052 ($+$)\\
			8-Polygon & \cellcolor{black!25}0.0031 & \cellcolor{black!10}0.0069 ($+$) & 0.0180 ($+$) & 0.0092 ($+$) & 0.0082 ($+$)\\
			8-RPolygon & \cellcolor{black!25}0.0032 & \cellcolor{black!10}0.0069 ($+$) & 0.0190 ($+$) & 0.0093 ($+$) & 0.0083 ($+$)\\
			10-Polygon & \cellcolor{black!25}0.0033 & \cellcolor{black!10}0.0064 ($+$) & 0.0204 ($+$) & 0.0087 ($+$) & 0.0074 ($+$)\\
			10-RPolygon & \cellcolor{black!25}0.0034 & \cellcolor{black!10}0.0064 ($+$) & 0.0185 ($+$) & 0.0086 ($+$) & 0.0075 ($+$)\\
			\hline
			\multicolumn{2}{|c|}{LORD-II vs. others ($+/-/\sim$)} & 8/0/0 & 8/0/0 & 8/0/0 & 8/0/0\\
			\hline
	\end{tabular}}
\end{table}

From Tables \ref{tab:poly_igdx_LORD} and \ref{tab:poly_igdf_LORD}, LORD-II is observed to be superior in both decision and objective spaces, respectively. The performance of all MMMOEAs (except TriMOEA\_TA\&R) are unaffected due to rotation. However, the IGDX values of TriMOEA\_TA\&R are widely different (poorer) for rotated polygon problems from those of the polygon problems (Table \ref{tab:poly_igdx_LORD}). This difference arises as TriMOEA\_TA\&R considers only the number of solutions as the diversity criteria and neglects the solution distribution in the decision space \cite{TriMOEA_TAR}.

The estimated PSs from LORD-II are shown in Table \ref{tab:poly_perf} from which the following observations are noted:
\begin{itemize}
	\item For all the 8 instances, LORD-II converges to global surfaces without any outliers.
	\item The number of solutions per subset is relatively uniform over the 9 subsets in PS.
	\item For both polygon and rotated polygon problems, the shape of the polygon is properly replicated for 3- and 5-objective problems. For 8- and 10-objective problem, a near-spherical blob (of unidentifiable shape) is formed at each of the subsets in PS.
\end{itemize}
\begin{table*}[!t]
	\caption{Cartesian coordinate plots of the 2-dimensional PSs of $M$-objective polygon and rotated polygon problems.}
	\label{tab:poly_perf}
	\resizebox{\linewidth}{!}{
		\begin{tabular}{c}
			\includegraphics{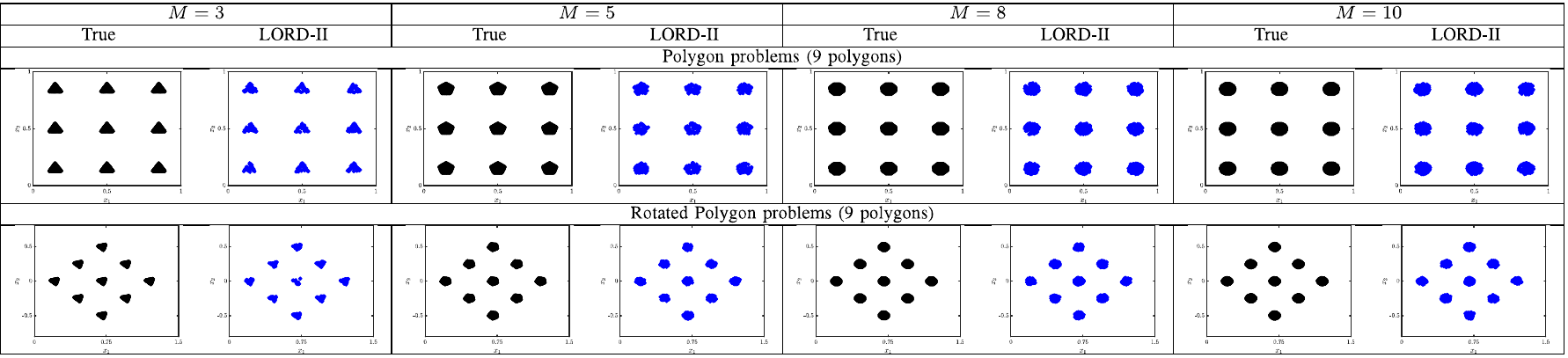}
	\end{tabular}}
\end{table*}

4) \textbf{Experiment-IV: Comparison of LORD and LORD-II with CEC 2019 winner:} To deal with the reduced selection pressure of dominance-based filtering approach of LORD, LORD-II is designed for higher number of objectives using PBI function. However, the performance of LORD-II on problems with small number of objectives has not been tested so far. Thus, this experiment compares the performance of LORD, LORD-II and CEC 2019 competition winner \cite{CEC19_winner} (MMO-Clustering PSO\footnote{As mentioned in \url{http://www5.zzu.edu.cn/ecilab/info/1036/1211.htm} with implementation obtained from Dr. Weiwei Zhang (anqikeli@163.com).}). It also uses $n_{pop} = 100N$ and $MaxFES = 5000N$ and considers the rPSP and rHV values for comparison. The performance of all the three MMMOEAs for 2-objective MMMOPs are specified in Table \ref{tab:comp_new} and validated using Wilcoxon's rank-sum test \cite{ARMOEA}.
\begin{table*}[!t]
	\caption{Mean $\pm$ Standard Deviation (Rank) of rPSP and rHV for $2$-objective MMMOPs over 51 Runs.}
	\label{tab:comp_new}
	\resizebox{\linewidth}{!}{
		\begin{tabular}{|c|c|c|c|c|c|c|}
			\hline
			& \multicolumn{3}{c|}{rPSP=IGDX/cov\_rate} & \multicolumn{3}{c|}{rHV=1/HV}\\
			\cline{2-7}
			& LORD & LORD-II & MMO-Clustering & LORD & LORD-II & MMO-Clustering\\
			Problems & & & PSO & & & PSO\\
			\hline
			MMF1 & \cellcolor{black!10}0.0441 $\pm$ 0.0044 & 0.0562 $\pm$ 0.0024 ($+$) & \cellcolor{black!25}0.0332 $\pm$ 0.0016 ($-$) & \cellcolor{black!25}1.0737 $\pm$ 0.0008 & 1.1480 $\pm$ 0.0004 ($+$) & \cellcolor{black!10}1.1456 $\pm$ 0.0003 ($+$)\\
			\hline
			MMF1\_z & \cellcolor{black!10}0.0356 $\pm$ 0.0069 & 0.0653 $\pm$ 0.0066 ($+$) & \cellcolor{black!25}0.0237 $\pm$ 0.0009 ($-$) & \cellcolor{black!25}1.0731 $\pm$ 0.0008 & 1.1466 $\pm$ 0.0005 ($+$) & \cellcolor{black!10}1.1454 $\pm$ 0.0002 ($+$)\\
			\hline
			MMF1\_e & \cellcolor{black!10}0.8894 $\pm$ 0.1466 & 1.6368 $\pm$ 0.5607 ($+$) & \cellcolor{black!25}0.5479 $\pm$ 0.0859 ($-$) & \cellcolor{black!25}1.0751 $\pm$ 0.0021 & \cellcolor{black!10}1.1532 $\pm$ 0.0018 ($+$) & 1.2560 $\pm$ 0.0376 ($+$)\\
			\hline
			MMF2 & \cellcolor{black!25}0.0219 $\pm$ 0.0108 & \cellcolor{black!10}0.0284 $\pm$ 0.0041 ($+$) & 0.1294 $\pm$ 0.0314 ($+$) & \cellcolor{black!25}1.0817 $\pm$ 0.0120 & \cellcolor{black!10}1.2168 $\pm$ 0.0067 ($+$) & 1.2605 $\pm$ 0.0177 ($+$)\\
			\hline
			MMF3 & \cellcolor{black!25}0.0200 $\pm$ 0.0105 & \cellcolor{black!10}0.0346 $\pm$ 0.0108 ($+$) & 0.0465 $\pm$ 0.0047 ($+$) & \cellcolor{black!25}1.0792 $\pm$ 0.0322 & 1.2527 $\pm$ 0.0689 ($+$) & \cellcolor{black!10}1.1920 $\pm$ 0.0042 ($+$)\\
			\hline
			MMF4 & \cellcolor{black!10}0.0253 $\pm$ 0.0036 & 0.0362 $\pm$ 0.0131 ($+$) & \cellcolor{black!25}0.0128 $\pm$ 0.0002 ($-$) & \cellcolor{black!25}1.5234 $\pm$ 0.0003 & 1.8545 $\pm$ 0.0005 ($+$) & \cellcolor{black!10}1.8510 $\pm$ 0.0010 ($+$)\\
			\hline
			MMF5 & \cellcolor{black!10}0.0814 $\pm$ 0.0080 & 0.0922 $\pm$ 0.0043 ($+$) & \cellcolor{black!25}0.0556 $\pm$ 0.0017 ($-$) & \cellcolor{black!25}1.0734 $\pm$ 0.0006 & 1.1481 $\pm$ 0.0002 ($+$) & \cellcolor{black!10}1.1456 $\pm$ 0.0004 ($+$)\\
			\hline
			MMF6 & \cellcolor{black!10}0.0692 $\pm$ 0.0104 & 0.0923 $\pm$ 0.0084 ($+$) & \cellcolor{black!25}0.0431 $\pm$ 0.0016 ($-$) & \cellcolor{black!25}1.0732 $\pm$ 0.0003 & 1.1472 $\pm$ 0.0012 ($+$) & \cellcolor{black!10}1.1453 $\pm$ 0.0004 ($+$)\\
			\hline
			MMF7 & \cellcolor{black!10}0.0219 $\pm$ 0.0044 & 0.0303 $\pm$ 0.0020  ($+$) & \cellcolor{black!25}0.0142 $\pm$ 0.0004 ($-$) & \cellcolor{black!25}1.0731 $\pm$ 0.0002 & \cellcolor{black!10}1.1458 $\pm$ 0.0005 ($+$) & 1.1445 $\pm$ 0.0004 ($+$)\\
			\hline
			MMF8 & \cellcolor{black!10}0.0745 $\pm$ 0.0452 & 0.0917 $\pm$ 0.0392 ($+$) & \cellcolor{black!25}0.0654 $\pm$ 0.0061 ($-$) & \cellcolor{black!25}1.7915 $\pm$ 0.0012 & \cellcolor{black!10}2.3794 $\pm$ 0.6116 ($+$) & 2.4027 $\pm$ 0.0060 ($+$)\\
			\hline
			MMF9 & \cellcolor{black!25}0.0047 $\pm$ 0.0002 & 0.0097 $\pm$ 0.0464 ($+$) & \cellcolor{black!10}0.0056 $\pm$ 0.0002 ($+$) & \cellcolor{black!25}0.0820 $\pm$ 0.0000 & 0.1172 $\pm$ 0.0005 ($+$) & \cellcolor{black!10}0.1032 $\pm$ 0.0000 ($+$)\\
			\hline
			MMF10 & \cellcolor{black!25}0.0018 $\pm$ 0.0007 & \cellcolor{black!10}0.0043 $\pm$ 0.0003 ($+$) & 0.0233 $\pm$ 0.0029 ($+$) & \cellcolor{black!25}0.0678 $\pm$ 0.0000 & 0.0811 $\pm$ 0.0002 ($+$) & \cellcolor{black!10}0.0797 $\pm$ 0.0001 ($+$)\\
			\hline
			MMF11 & \cellcolor{black!25}0.0029 $\pm$ 0.0002 & \cellcolor{black!10}0.0045 $\pm$ 0.0001 ($+$) & 0.0046 $\pm$ 0.0001 ($+$) & \cellcolor{black!25}0.0581 $\pm$ 0.0000 & 0.0738 $\pm$ 0.0002 ($+$) & \cellcolor{black!10}0.0690 $\pm$ 0.0000 ($+$)\\
			\hline
			MMF12 & \cellcolor{black!25}0.0013 $\pm$ 0.0001 & \cellcolor{black!10}0.0053 $\pm$ 0.0003 ($+$) & 0.0055 $\pm$ 0.0005 ($+$) & \cellcolor{black!25}0.5431 $\pm$ 0.0000 & 0.8312 $\pm$ 0.0041 ($+$) & \cellcolor{black!10}0.6686 $\pm$ 0.0115 ($+$)\\
			\hline
			MMF13 & \cellcolor{black!25}0.0243 $\pm$ 0.0039 & 0.2723 $\pm$ 0.0631 ($+$) & \cellcolor{black!10}0.0325 $\pm$ 0.0010 ($+$) & \cellcolor{black!25}0.0444 $\pm$ 0.0000 & \cellcolor{black!10}0.0550 $\pm$ 0.0002 ($+$) & 0.0544 $\pm$ 0.0000 ($+$)\\
			\hline
			Omni-test & \cellcolor{black!25}0.0754 $\pm$ 0.0242 & \cellcolor{black!10}0.1215 $\pm$ 0.0424 ($+$) & 0.5470 $\pm$ 0.0476 ($+$) & \cellcolor{black!10}0.0518 $\pm$ 0.0000 & \cellcolor{black!25}0.0190 $\pm$ 0.0000 ($-$) & \cellcolor{black!25}0.0190 $\pm$ 0.0000 ($-$)\\
			\hline
			SYM-PART & \cellcolor{black!25}0.0556 $\pm$ 0.0145 & \cellcolor{black!10}0.1191 $\pm$ 0.0206 ($+$) & 0.2801 $\pm$ 0.0370 ($+$) & \cellcolor{black!25}0.0520 $\pm$ 0.0000 & \cellcolor{black!10}0.0601 $\pm$ 0.0000 ($+$) & 0.0610 $\pm$ 0.0001 ($+$)\\
			simple & \cellcolor{black!25} & \cellcolor{black!10} & & \cellcolor{black!25} & \cellcolor{black!10} & \\
            \hline
            SYM-PART & \cellcolor{black!25}0.1730 $\pm$ 0.0743 & \cellcolor{black!10}0.2252 $\pm$ 0.0389 ($+$) & 0.2729 $\pm$ 0.0262 ($+$) & \cellcolor{black!25}0.0520 $\pm$ 0.0000 & \cellcolor{black!10}0.0602 $\pm$ 0.0000 ($+$) & 0.0609 $\pm$ 0.0001 ($+$)\\
            rotated & \cellcolor{black!25} & \cellcolor{black!10} & & \cellcolor{black!25} & \cellcolor{black!10} & \\
            \hline
            \multicolumn{2}{|c|}{LORD vs. others ($+$/$-$/$\sim$)}& 18/0/0 & 10/8/0 & ($+$/$-$/$\sim$) & 17/1/0 & 17/1/0\\
            \hline
	\end{tabular}}
\end{table*}

According to Table \ref{tab:comp_new}, LORD demonstrates superior performance in 17 out of 18 cases in the objective space. In the decision space, LORD outperforms LORD-II in all the 18 cases and the CEC 2019 winner in 10 out of 18 cases. Thus, LORD can be considered as a very robust algorithm for 2-objective MMMOPs.    

5) \textbf{Experiment-V: Comparison by Variation in Population Size}: While a large population size ($n_{pop}$) is a necessity for MMMOPs (as mentioned in Section \ref{sec1}), standard MOEAs such as MOEA/DD may have poor performance due to a large $n_{pop}$. For a fair assessment on the superiority of LORD-II, this experiment compares LORD-II with MOEA/DD using both small $n_{pop}$ (as per the optimal setting of MOEA/DD in \cite{MOEADD}) and large $n_{pop}$ ($=100\times N$ as per the recommendation of CEC 2019 MMMOPs \cite{CEC19_TR}) in Table \ref{tab:MOEADD_npop_runs_LORD}, from which the following insights are obtained:
\begin{itemize}
	\item LORD-II is superior even for small $n_{pop}$.
	\item While MOEA/DD never outperforms LORD-II in the decision space, the former is marginally superior for a few cases (one out of 16 cases for small $n_{pop}$ and two out of 16 cases for large $n_{pop}$) in the objective space.
	\item A large $n_{pop}$ improves IGDX and IGDF regardless of the effectiveness of the underlying algorithm \cite{NIMMO_SWEVO19}. This behavior is also observed in Table \ref{tab:MOEADD_npop_runs_LORD} for both LORD-II and MOEA/DD. However, since the superiority of LORD-II against MOEA/DD is also established for a small $n_{pop}$, these results indeed reflect the efficient synergism of various strategies in the evolutionary framework of LORD-II.
\end{itemize}

\begin{table*}[!ht]
	\centering
	\caption{Mean of IGDX and IGDF over 51 independent runs with different population sizes ($n_{pop}$) for $M$-objective MMMOPs.}
	\label{tab:MOEADD_npop_runs_LORD}
	\resizebox{\linewidth}{!}{
		\begin{tabular}{|c|c|c|c|c|c|c|c|c|c|c|c|}
			\hline
			& & \multicolumn{5}{c|}{\textbf{Recommended Population Size for MOEA/DD in \cite{MOEADD}}} & \multicolumn{5}{c|}{\textbf{Recommended Population Size for MMMOPs in \cite{CEC19_TR}}}\\
			\cline{3-12}
			\textbf{Problems} & $M$ & & \multicolumn{2}{c|}{\textbf{IGDX}} & \multicolumn{2}{c|}{\textbf{IGDF}} & & \multicolumn{2}{c|}{\textbf{IGDX}} & \multicolumn{2}{c|}{\textbf{IGDF}} \\
			\cline{4-7}\cline{9-12}
			& & $n_{pop}$ & \textbf{LORD-II} & \textbf{MOEA/DD} & \textbf{LORD-II} & \textbf{MOEA/DD} & $n_{pop}$ & \textbf{LORD-II} & \textbf{MOEA/DD} & \textbf{LORD-II} & \textbf{MOEA/DD}\\
			\hline
			MMF14 & 3 & 91 & \cellcolor{black!25}0.0832 & 0.2150 ($+$) & \cellcolor{black!25}0.1044 & 0.1045 ($\sim$) & 300 & \cellcolor{black!25}0.0443 & 0.0671 ($+$) & \cellcolor{black!25}0.0540 & 0.0555 ($+$)\\
			MMF14\_a & 3 & 91 & \cellcolor{black!25}0.1150 & 0.2076 ($+$) & \cellcolor{black!25}0.1044 & 0.1045 ($\sim$) & 300 & \cellcolor{black!25}0.0576 & 0.0780 ($+$) & \cellcolor{black!25}0.0561 & 0.0568 ($+$)\\
			MMF15 & 3 & 91 & \cellcolor{black!25}0.0514 & 0.0522 ($\sim$) & \cellcolor{black!25}0.1055 & 0.1056 ($\sim$) & 300 & \cellcolor{black!25}0.0287 & 0.0295 ($+$) & \cellcolor{black!25}0.0548 & 0.0562 ($+$)\\
			MMF15\_a & 3 & 91 & \cellcolor{black!25}0.0638 & 0.0705 ($+$) & \cellcolor{black!25}0.1056 & 0.1144 ($+$) & 300 & \cellcolor{black!25}0.0355 & 0.0357 ($\sim$) & \cellcolor{black!25}0.0571 & 0.0607 ($+$)\\
			\hline
			MMF14 & 5 & 210 & \cellcolor{black!25}0.3070 & 0.3314 ($+$) & \cellcolor{black!25}0.3125 & 0.3136 ($+$) & 495 & \cellcolor{black!25}0.2448 & 0.2554 ($+$) & \cellcolor{black!25}0.0564 & 0.0598 ($+$)\\
			MMF14\_a & 5 & 210 & \cellcolor{black!25}0.3283 & 0.4083 ($+$) & \cellcolor{black!25}0.3129 & 0.3135 ($\sim$) & 495 & \cellcolor{black!25}0.2670 & 0.2846 ($+$) & \cellcolor{black!25}0.0752 & 0.0839 ($+$)\\
			MMF15 & 5 & 210 & \cellcolor{black!25}0.2460 & 0.2652 ($+$) & \cellcolor{black!25}0.3155 & 0.3167 ($+$) & 495 & \cellcolor{black!25}0.1960 & 0.2032 ($+$) & \cellcolor{black!25}0.0602 & 0.0645 ($+$)\\
			MMF15\_a & 5 & 210 & \cellcolor{black!25}0.2695 & 0.2963 ($+$) & 0.3181 & \cellcolor{black!25}0.3168 ($-$) & 495 & \cellcolor{black!25}0.2155 & 0.2230 ($+$) & \cellcolor{black!25}0.0895 & 0.0999 ($+$)\\
			\hline
			MMF14 & 8 & 156 & \cellcolor{black!25}0.6864 & 0.7006 ($+$) & \cellcolor{black!25}0.7233 & 0.7244 ($+$) & 828 & \cellcolor{black!25}0.5621 & 0.5857 ($+$) & \cellcolor{black!25}0.1445 & 0.1494 ($+$)\\
			MMF14\_a & 8 & 156 & \cellcolor{black!25}0.6851 & 0.7363 ($+$) & \cellcolor{black!25}0.7225 & 0.7241 ($+$) & 828 & \cellcolor{black!25}0.5725 & 0.5936 ($+$) & \cellcolor{black!25}0.1776 & 0.1905 ($+$)\\
			MMF15 & 8 & 156 & \cellcolor{black!25}0.6086 & 0.6263 ($+$) & \cellcolor{black!25}0.7270 & 0.7291 ($+$) & 828 & \cellcolor{black!25}0.5146 & 0.5586 ($+$) & 0.1503 & \cellcolor{black!25}0.1486 ($-$)\\
			MMF15\_a & 8 & 156 & 0.6543 & \cellcolor{black!25}0.6539 ($\sim$) & \cellcolor{black!25}0.7277 & 0.7289 ($+$) & 828 & \cellcolor{black!25}0.5315 & 0.5498 ($+$) & 0.2195 & \cellcolor{black!25}0.2159 ($\sim$)\\
			\hline
			MMF14 & 10 & 275 & \cellcolor{black!25}0.8404 & 0.8847 ($+$) & \cellcolor{black!25}0.6811 & 0.6864 ($+$) & 935 & \cellcolor{black!25}0.7088 & 0.7373 ($+$) & 0.3463 & \cellcolor{black!25}0.3102 ($\sim$)\\
			MMF14\_a & 10 & 275 & \cellcolor{black!25}0.8374 & 0.8972 ($+$) & \cellcolor{black!25}0.6839 & 0.6907 ($+$) & 935 & \cellcolor{black!25}0.6869 & 0.7241 ($+$) & \cellcolor{black!25}0.4296 & 0.4317 ($\sim$)\\
			MMF15 & 10 & 275 & \cellcolor{black!25}0.7787 & 0.8105 ($+$) & \cellcolor{black!25}0.6864 & 0.6903 ($+$) & 935 & \cellcolor{black!25}0.6469 & 0.6731 ($+$) & 0.3561 & \cellcolor{black!25}0.2984 ($-$)\\
			MMF15\_a & 10 & 275 & \cellcolor{black!25}0.8074 & 0.8246 ($+$) & \cellcolor{black!25}0.6913 & 0.6940 ($+$) & 935 & \cellcolor{black!25}0.6712 & 0.6848 ($+$) & \cellcolor{black!25}0.4375 & 0.4384 ($\sim$)\\
			\hline
			\multicolumn{4}{|c|}{LORD-II vs. MOEA/DD ($+/-/\sim$)} & 14/0/2 & ($+/-/\sim$) & 11/1/4 & \multicolumn{2}{c|}{($+/-/\sim$)} & 15/0/1 & ($+/-/\sim$) & 10/2/4 \\
			\hline
	\end{tabular}}
\end{table*}

Thus, it is evident that the improved performance is an attribute of the algorithmic framework of LORD-II and not of the large $n_{pop}$.

\subsection{Scalability Study on LORD-II framework}
As most of the MMMOEAs are not tested on problems scalable in candidate dimension ($N$) \cite{TriMOEA_TAR}, the scalability of LORD-II is established by studying its performance in Table \ref{tab:scale_LORD} with variations in $N$ using a 3-objective MMF14 problem for $N = \{3, 10, 30, 50, 100\}$.
\begin{table}[!ht]
	\centering
	\caption{Mean of rPSP, IGDX, rHV and IGDF for $3$-objective MMF14 (with different candidate dimensions, $N$) over 51 Independent Runs of LORD-II.}
	\label{tab:scale_LORD}
	\resizebox{0.4\linewidth}{!}{
		\begin{tabular}{|c|cc|cc|}
			\hline
			$N$ & rPSP & IGDX & rHV & IGDF\\
			\hline
			$3$		& 0.0449 & 0.0443 & 1.0395 & 0.0540\\
			$10$	& 0.5928 & 0.5838 & 1.0414 & 0.0013\\
			$30$	& 2.8270 & 1.5038 & 1.0402 & 0.0001\\
			$50$	& 2.1513 & 2.1258 & 1.0405 & 0.0001\\
			$100$	& 3.2476 & 3.1807 & 1.0406 & 0.0000\\
			\hline
	\end{tabular}}
\end{table}

From Table \ref{tab:scale_LORD}, the following observations are noted:
\begin{itemize}
	\item As the number of objectives ($M$) does not change, the performance of LORD-II remains unaffected in the objective space as noted from the absence of any significant increase in rHV and IGDF. 
	\item For small $N$, the performance in the decision space deteriorates only linearly (not exponentially) with an increase in $N$. For example, IGDX increases 34 times when $N$ is increased from 3 to 30. However, with further increase in $N$, the deterioration in performance is even less drastic. For example, IGDX only doubles when $N$ is increased from 30 to 100.
\end{itemize}

Thus, LORD-II, using decomposition of decision and objective spaces, works efficiently even for high-dimensional MMMOPs, i.e., LORD-II is scalable with problem size.

\section{Conclusion and Future Research Directions}
 \label{sec5}
As most of the existing MMMOEAs evaluates crowding distance over the entire decision space, its analysis exhibits a major disadvantage which is identified as the crowding illusion problem (Section \ref{sec_illusion}). To mitigate the adverse effects of this problem for MMMOPs, a novel evolutionary framework is presented in this manuscript. It is a first of its kind algorithm to attempt decomposition of decision space using graph Laplacian based clustering for maintaining the diversity of solutions in that space. It uses reference vectors to partition the objective space for maintaining diversity in the objective space. The proposed algorithm has two different versions which differently impart the selection pressure on the population. The first version (LORD) is for MMMOPs with small number of objectives, which eliminates the maximally crowded solution from the last non-dominated rank. The second version (LORD-II) is for problems with higher number of objectives which eliminates the candidate with maximal PBI, from the maximally large cluster. During elimination of candidate, LORD and LORD-II try to ensure that the removal does not occur from the sub-spaces (defined by reference vectors) with only one associated candidate. The proposed frameworks have been tested over several MMMOPs \cite{CEC19_TR} and MMMaOPs \cite{CEC19_TR,POLY_PROBS} and their performance have been compared with recent state-of-the-art algorithms to establish their efficacy.

Inspite of their superior performance, LORD and LORD-II have the following disadvantages; reducing the effects of which could be considered for future extensions of this work:
\begin{enumerate}
	\item Spectral clustering solves the preliminary purpose of decomposing the decision space and is effective for problems like MMF2, MMF3, SYM-PART, etc. However, for problems like MMF1, MMF6, MMF7, etc., where overlap exists along certain dimensions within each subset of PS \cite{CEC19_TR}, spectral clustering does not completely eliminate the crowding illusion problem (Section \ref{sec_illusion}). Hence, the search for better decomposition methods (with adaptive threshold/parameters) in the decision space forms a potential future work.
	
	\item As done in the proposed work as well as in most of the previous works such as \cite{MOEA/D-AD,DETriM}, the mating pool selection is performed from neighboring candidates in the objective space. Nonetheless, if an approximate partitioning of the decision space could be done to yield the bounding box of each subset of PS, intra- and inter-subset mating could further be studied for enhanced diversity of solutions in the decision space. Moreover, different subsets of the PS of an MMMOP have correlated structures. Such information on correlation could be incorporated to design new perturbation operators for guiding the search in MMMOPs.
	
	\item It should be noted that the research on MMMOPs has just started developing. Hence, besides designing more algorithms to deal with MMMOPs, other challenges in this new direction involve developing novel decision-making strategies (selecting one out of multiple equivalent solutions from the PS mapping to a certain solution in the PF), analyzing more practical problems, designing more difficult benchmark problems (e.g., CEC 2020 test suite \cite{CEC20_TR}), validating existing approaches on such problems as well as designing new performance measures (independent of the true PS) to study the convergence and diversity in the decision space for practical problems. 
\end{enumerate}

\bibliographystyle{unsrt}
\bibliography{RefDB1}

\begin{thebibliography}{10}

\bibitem{SWEVO_2011}
Aimin Zhou, Bo-Yang Qu, Hui Li, Shi-Zheng Zhao, Ponnuthurai~Nagaratnam
  Suganthan, and Qingfu Zhang.
\newblock Multiobjective evolutionary algorithms: A survey of the state of the
  art.
\newblock {\em Swarm and Evolutionary Computation}, 1(1):32--49, 2011.

\bibitem{UM_2014TEVCReview_PartI}
Anirban Mukhopadhyay, Ujjwal Maulik, Sanghamitra Bandyopadhyay, and Carlos
  Artemio~Coello Coello.
\newblock A survey of multiobjective evolutionary algorithms for data mining:
  Part {I}.
\newblock {\em IEEE Transactions on Evolutionary Computation}, 18(1):4--19, Feb
  2014.

\bibitem{Review_MMMOP}
Ryoji Tanabe and Hisao Ishibuchi.
\newblock A review of evolutionary multi-modal multi-objective optimization.
\newblock {\em IEEE Transactions on Evolutionary Computation}, pages 1--9,
  2019.

\bibitem{CEC19_TR}
J.~J. Liang, B.~Y. Qu, D.~W. Gong, and C.~T. Yue.
\newblock Problem definitions and evaluation criteria for the {CEC} 2019
  special session on multimodal multiobjective optimization.
\newblock {\em Technical Report, Computational Intelligence Laboratory,
  Zhengzhou University}, 2019.

\bibitem{Appl2_MMMOP}
Fumiya Kudo, Tomohiro Yoshikawa, and Takeshi Furuhashi.
\newblock A study on analysis of design variables in pareto solutions for
  conceptual design optimization problem of hybrid rocket engine.
\newblock In {\em 2011 IEEE Congress of Evolutionary Computation (CEC)}, pages
  2558--2562, June 2011.

\bibitem{Appl1_MMMOP}
C.~T. Yue, J.~J. Liang, B.~Y. Qu, K.~J. Yu, and H.~Song.
\newblock Multimodal multiobjective optimization in feature selection.
\newblock In {\em 2019 IEEE Congress on Evolutionary Computation (CEC)}, pages
  302--309, June 2019.

\bibitem{Appl3_MMMOP}
Jafar Jamal, Roberto Montemanni, David Huber, Marco Derboni, and Andrea~E
  Rizzoli.
\newblock A multi-modal and multi-objective journey planner for integrating
  carpooling and public transport.
\newblock {\em Journal of Traffic and Logistics Engineering}, 5(2):68--72,
  December 2017.

\bibitem{DETriM}
Monalisa Pal and Sanghamitra Bandyopadhyay.
\newblock Differential evolution for multi-modal multi-objective problems.
\newblock In {\em Proceedings of the Genetic and Evolutionary Computation
  Conference Companion}, GECCO '19, pages 1399--1406, New York, NY, USA, 2019.
  ACM.

\bibitem{NSGA2}
K.~Deb, A.~Pratap, S.~Agarwal, and T.~Meyarivan.
\newblock A fast and elitist multiobjective genetic algorithm: {NSGA-II}.
\newblock {\em IEEE Transactions on Evolutionary Computation}, 6(2):182--197,
  2002.

\bibitem{thDEA}
Y.~Yuan, X.~Hua, W.~Bo, and Y.~Xin.
\newblock A new dominance relation-based evolutionary algorithm for
  many-objective optimization.
\newblock {\em IEEE Transactions on Evolutionary Computation}, 20(1):16--37,
  2016.

\bibitem{HypE}
Johannes Bader and Eckart Zitzler.
\newblock {H}yp{E}: An algorithm for fast hypervolume-based many-objective
  optimization.
\newblock {\em Evolutionary computation}, 19(1):45--76, 2011.

\bibitem{GDE-MOEA}
Adriana Menchaca-Mendez and Carlos A~Coello Coello.
\newblock {GDE-MOEA}: a new moea based on the generational distance indicator
  and $\varepsilon$-dominance.
\newblock In {\em 2015 IEEE Congress on Evolutionary Computation (CEC)}, pages
  947--955. IEEE, 2015.

\bibitem{MOEA/D}
Z.~Qingfu and L.~Hui.
\newblock {MOEA/D}: A multiobjective evolutionary algorithm based on
  decomposition.
\newblock {\em IEEE Transactions on Evolutionary Computation}, 11(6):712--731,
  2007.

\bibitem{NSGA3}
Kalyanmoy Deb and Himanshu Jain.
\newblock An evolutionary many-objective optimization algorithm using
  reference-point-based nondominated sorting approach, part {I}: Solving
  problems with box constraints.
\newblock {\em IEEE Transactions on Evolutionary Computation}, 18(4):577--601,
  2014.

\bibitem{MOEADD}
Ke~Li, Kalyanmoy Deb, Qingfu Zhang, and Sam Kwong.
\newblock An evolutionary many-objective optimization algorithm based on
  dominance and decomposition.
\newblock {\em IEEE Transactions on Evolutionary Computation}, 19(5):694--716,
  2015.

\bibitem{NBI}
I.~Das and J.~E. Dennis.
\newblock Normal-boundary intersection: A new method for generating the pareto
  surface in nonlinear multicriteria optimization problems.
\newblock {\em SIAM Journal on Optimization}, 8(3):631--657, 1998.

\bibitem{Sys_Decomp}
Qi~Kang, Xinyao Song, MengChu Zhou, and Li~Li.
\newblock A collaborative resource allocation strategy for decomposition-based
  multiobjective evolutionary algorithms.
\newblock {\em IEEE Transactions on Systems, Man, and Cybernetics: Systems},
  49(12):2416--2423, Dec 2019.

\bibitem{Omni}
Kalyanmoy Deb and Santosh Tiwari.
\newblock Omni-optimizer: A procedure for single and multi-objective
  optimization.
\newblock In Carlos~A. Coello~Coello, Arturo Hern{\'a}ndez~Aguirre, and Eckart
  Zitzler, editors, {\em Evolutionary Multi-Criterion Optimization}, pages
  47--61, Berlin, Heidelberg, 2005. Springer Berlin Heidelberg.

\bibitem{Lit21}
K.~P. Chan and T.~Ray.
\newblock An evolutionary algorithm to maintain diversity in the parametric and
  the objective space.
\newblock In {\em International Conference on Computational Robotics and
  Autonomous Systems (CIRAS), Centre for Intelligent Control, National
  University of Singapore}, 2005.

\bibitem{IGDX}
Aimin Zhou, Qingfu Zhang, and Yaochu Jin.
\newblock Approximating the set of pareto-optimal solutions in both the
  decision and objective spaces by an estimation of distribution algorithm.
\newblock {\em IEEE Transactions on Evolutionary Computation},
  13(5):1167--1189, 2009.

\bibitem{DN-NSGA2}
J.~J. Liang, C.~T. Yue, and B.~Y. Qu.
\newblock Multimodal multi-objective optimization: A preliminary study.
\newblock In {\em 2016 IEEE Congress on Evolutionary Computation (CEC)}, pages
  2454--2461. IEEE, 2016.

\bibitem{GECCO19_poster}
Mahrokh Javadi, Heiner Zille, and Sanaz Mostaghim.
\newblock Modified crowding distance and mutation for multimodal
  multi-objective optimization.
\newblock In {\em Proceedings of the Genetic and Evolutionary Computation
  Conference Companion}, GECCO '19, pages 211--212, New York, NY, USA, 2019.
  ACM.

\bibitem{MO_Ring_PSO_SCD}
Caitong Yue, Boyang Qu, and Jing Liang.
\newblock A multiobjective particle swarm optimizer using ring topology for
  solving multimodal multiobjective problems.
\newblock {\em IEEE Transactions on Evolutionary Computation}, 22(5):805--817,
  Oct 2018.

\bibitem{ZoneS}
Qinqin Fan and Xuefeng Yan.
\newblock Solving multimodal multiobjective problems through zoning search.
\newblock {\em IEEE Transactions on Systems, Man, and Cybernetics: Systems},
  pages 1--12, 2019.

\bibitem{MOEA/D-AD}
Ryoji Tanabe and Hisao Ishibuchi.
\newblock A decomposition-based evolutionary algorithm for multi-modal
  multi-objective optimization.
\newblock In Anne Auger, Carlos~M. Fonseca, Nuno Louren{\c{c}}o, Penousal
  Machado, Lu{\'i}s Paquete, and Darrell Whitley, editors, {\em Parallel
  Problem Solving from Nature -- PPSN XV}, pages 249--261, Cham, 2018. Springer
  International Publishing.

\bibitem{ADA_2020}
Ryoji Tanabe and Hisao Ishibuchi.
\newblock A framework to handle multimodal multiobjective optimization in
  decomposition-based evolutionary algorithms.
\newblock {\em IEEE Transactions on Evolutionary Computation}, 24(4):720--734,
  2020.

\bibitem{TriMOEA_TAR}
Yiping Liu, Gary~G. Yen, and Dunwei Gong.
\newblock A multi-modal multi-objective evolutionary algorithm using
  two-archive and recombination strategies.
\newblock {\em IEEE Transactions on Evolutionary Computation}, 23(4):660--674,
  Aug 2019.

\bibitem{MMNAEMO_CEC19}
K.~Maity, R.~Sengupta, and S.~Saha.
\newblock {MM}-{NAEMO} : Multimodal neighborhood-sensitive archived
  evolutionary many-objective optimization algorithm.
\newblock In {\em 2019 IEEE Congress on Evolutionary Computation (CEC)}, pages
  286--294, June 2019.

\bibitem{NIMMO_SWEVO19}
Ryoji Tanabe and Hisao Ishibuchi.
\newblock A niching indicator-based multi-modal many-objective optimizer.
\newblock {\em Swarm and Evolutionary Computation}, 49:134 -- 146, 2019.

\bibitem{POLY_PROBS}
Hisao Ishibuchi, Naoya Akedo, and Yusuke Nojima.
\newblock A many-objective test problem for visually examining diversity
  maintenance behavior in a decision space.
\newblock In {\em Proceedings of the 13th Annual Conference on Genetic and
  Evolutionary Computation}, GECCO '11, pages 649--656, New York, NY, USA,
  2011. ACM.

\bibitem{NAEMO}
Raunak Sengupta, Monalisa Pal, Sriparna Saha, and Sanghamitra Bandyopadhyay.
\newblock {NAEMO}: Neighborhood-sensitive archived evolutionary many-objective
  optimization algorithm.
\newblock {\em Swarm and Evolutionary Computation}, 46:201 -- 218, 2019.

\bibitem{polymut_recentref}
Li-Min Li, Kang-Di Lu, Guo-Qiang Zeng, Lie Wu, and Min-Rong Chen.
\newblock A novel real-coded population-based extremal optimization algorithm
  with polynomial mutation: A non-parametric statistical study on continuous
  optimization problems.
\newblock {\em Neurocomputing}, 174:577 -- 587, 2016.

\bibitem{Sys_DE}
Xianpeng Wang, Zhiming Dong, and Lixin Tang.
\newblock Multiobjective differential evolution with personal archive and
  biased self-adaptive mutation selection.
\newblock {\em IEEE Transactions on Systems, Man, and Cybernetics: Systems},
  pages 1--13, 2018.

\bibitem{RSA}
Raunak Sengupta and Sriparna Saha.
\newblock Reference point based archived many objective simulated annealing.
\newblock {\em Information Sciences}, 467:725 -- 749, 2018.

\bibitem{DEMO}
Tea Robi{\v{c}} and Bogdan Filipi{\v{c}}.
\newblock {DEMO}: Differential evolution for multiobjective optimization.
\newblock In C.~A.~C. Coello, A.~H. Aguirre, and E.~Zitzler, editors, {\em
  Evolutionary Multi-Criterion Optimization}, pages 520--533, Berlin,
  Heidelberg, 2005. Springer Berlin Heidelberg.

\bibitem{SaDE}
A.~K. Qin and P.~N. Suganthan.
\newblock Self-adaptive differential evolution algorithm for numerical
  optimization.
\newblock In {\em Evolutionary Computation, 2005. The 2005 IEEE Congress on},
  volume~2, pages 1785--1791. IEEE, 2005.

\bibitem{SpecClus}
Ulrike von Luxburg.
\newblock A tutorial on spectral clustering.
\newblock {\em Statistics and Computing}, 17(4):395--416, Dec 2007.

\bibitem{Aparajita_SpecClus}
Aparajita Khan and Pradipta Maji.
\newblock Approximate graph laplacians for multimodal data clustering.
\newblock {\em IEEE Transactions on Pattern Analysis and Machine Intelligence},
  pages 1--16, 2019.

\bibitem{Cheeger1}
Jeff Cheeger.
\newblock A lower bound for the smallest eigenvalue of the {L}aplacian.
\newblock In Robert~C. Gunning, editor, {\em Problems in analysis (Papers
  dedicated to Salomon Bochner, 1969)}, pages 195--199. Princeton Univ. Press,
  Princeton, NJ, 1970.

\bibitem{Cheeger2}
Peter Buser.
\newblock {\"U}ber eine ungleichung von {C}heeger ({O}n an inequality of
  {C}heeger).
\newblock {\em Mathematische Zeitschrift (in German)}, 158(3):245--252, Oct
  1978.

\bibitem{kmeans}
Michael~B. Cohen, Sam Elder, Cameron Musco, Christopher Musco, and Madalina
  Persu.
\newblock Dimensionality reduction for k-means clustering and low rank
  approximation.
\newblock In {\em Proceedings of the Forty-seventh Annual {ACM} Symposium on
  Theory of Computing}, STOC '15, pages 163--172, New York, NY, USA, 2015. ACM.

\bibitem{spec_versus_kmeans}
Abdelkarim~Ben Ayed, Mohamed~Ben Halima, and Adel~M Alimi.
\newblock Adaptive fuzzy exponent cluster ensemble system based feature
  selection and spectral clustering.
\newblock In {\em 2017 IEEE International Conference on Fuzzy Systems
  (FUZZ-IEEE)}, pages 1--6. IEEE, 2017.

\bibitem{SS_ndSort}
Sumit Mishra, Mondal. Samrat, and Sriparna Saha.
\newblock Fast implementation of steady-state {NSGA-II}.
\newblock In {\em 2016 IEEE Congress on Evolutionary Computation (CEC)}, pages
  3777--3784, July 2016.

\bibitem{ESOEA}
Monalisa Pal and Sanghamitra Bandyopadhyay.
\newblock {ESOEA}: Ensemble of single objective evolutionary algorithms for
  many-objective optimization.
\newblock {\em Swarm and Evolutionary Computation}, 50:100511, 2019.

\bibitem{Coello_Review}
C.~A.~C. Coello.
\newblock Recent results and open problems in evolutionary multiobjective
  optimization.
\newblock In C.~Mart{\'i}n-Vide, R.~Neruda, and M.~A. Vega-Rodr{\'i}guez,
  editors, {\em Theory and Practice of Natural Computing}, pages 3--21, Cham,
  2017. Springer International Publishing.

\bibitem{aDEMO}
S.~Bandyopadhyay and A.~Mukherjee.
\newblock An algorithm for many-objective optimization with reduced objective
  computations: A study in differential evolution.
\newblock {\em IEEE Transactions on Evolutionary Computation}, 19(3):400--413,
  2015.

\bibitem{pop_dynam_indicators}
Raunak Sengupta, Monalisa Pal, Sriparna Saha, and Sanghamitra Bandyopadhyay.
\newblock Population dynamics indicators for evolutionary many-objective
  optimization.
\newblock In {\em Progress in Advanced Computing and Intelligent Engineering},
  pages 261--271. Springer, 2019.

\bibitem{ARMOEA}
Ye~Tian, Ran Cheng, Xingyi Zhang, Fan Cheng, and Yaochu Jin.
\newblock An indicator based multi-objective evolutionary algorithm with
  reference point adaptation for better versatility.
\newblock {\em IEEE Transactions on Evolutionary Computation}, 2017.

\bibitem{IGDNS}
Ye~Tian, Xingyi Zhang, Ran Cheng, and Yaochu Jin.
\newblock A multi-objective evolutionary algorithm based on an enhanced
  inverted generational distance metric.
\newblock In {\em 2016 IEEE Congress on Evolutionary Computation (CEC)}, pages
  5222--5229, July 2016.

\bibitem{UM_2014TEVCReview_PartII}
Anirban Mukhopadhyay, Ujjwal Maulik, Sanghamitra Bandyopadhyay, and Carlos
  Artemio~Coello Coello.
\newblock Survey of multiobjective evolutionary algorithms for data mining:
  Part {II}.
\newblock {\em IEEE Transactions on Evolutionary Computation}, 18(1):20--35,
  Feb 2014.

\bibitem{DE_survey}
S.~Das, S.~S. Mullick, and P.~N. Suganthan.
\newblock Recent advances in differential evolution – an updated survey.
\newblock {\em Swarm and Evolutionary Computation}, 27:1 -- 30, 2016.

\bibitem{GA}
K.~Deb and R.~B. Agrawal.
\newblock Simulated binary crossover for continuous search space.
\newblock {\em Complex systems}, 9(2):115--148, 1995.

\bibitem{DEvsGA}
Tea Tu{\v{s}}ar and Bogdan Filipi{\v{c}}.
\newblock Differential evolution versus genetic algorithms in multiobjective
  optimization.
\newblock In Shigeru Obayashi, Kalyanmoy Deb, Carlo Poloni, Tomoyuki Hiroyasu,
  and Tadahiko Murata, editors, {\em Evolutionary Multi-Criterion
  Optimization}, pages 257--271, Berlin, Heidelberg, 2007. Springer Berlin
  Heidelberg.

\bibitem{radial_plot}
Zhenan He and Gary~G Yen.
\newblock Visualization and performance metric in many-objective optimization.
\newblock {\em IEEE Transactions on Evolutionary Computation}, 20(3):386--402,
  June 2016.

\bibitem{CEC19_winner}
C.T. Yue, J.J. Liang, P.N. Suganthan, B.Y. Qu, K.J. Yu, and S.~Liu.
\newblock {MMOGA} for solving multimodal multiobjective optimization problems
  with local pareto sets.
\newblock In {\em 2020 IEEE Congress on Evolutionary Computation (CEC)}, pages
  1--8. IEEE, 2020.

\bibitem{CEC20_TR}
J.~J. Liang, P.~N. Suganthan, B.~Y. Qu, D.~W. Gong, and C.~T. Yue.
\newblock Problem definitions and evaluation criteria for the {CEC} 2020
  special session on multimodal multiobjective optimization.
\newblock {\em Technical Report 201912, Computational Intelligence Laboratory,
  Zhengzhou University, Zhengzhou China And Technical Report, Nanyang
  Technological University, Singapore}, 2019.

\end{thebibliography}

\end{document}